\documentclass[10pt]{article}
\usepackage[preprint]{tmlr}
\usepackage{amsmath,amssymb}
\usepackage{graphicx}
\usepackage{booktabs,tabularx,array,longtable}
\usepackage{microtype}
\usepackage{xcolor}
\usepackage[most]{tcolorbox}
\newif\ifprovisional
\provisionalfalse 

\definecolor{provred}{RGB}{190,0,0}

\tcbuselibrary{breakable} \newtcolorbox{outcomebox}[1][]{breakable,enhanced jigsaw,colback=provred!3,colframe=provred!55,boxrule=0.5pt,arc=1mm, left=4pt,right=4pt,top=2pt,bottom=2pt,fonttitle=\footnotesize\bfseries,coltitle=white,colbacktitle=provred!70, title={Pre-written readings: the one that matches the result is kept, the others are deleted},#1} 

\newcommand{\provbanner}{}

\ifprovisional
  \usepackage{draftwatermark}
\SetWatermarkText{PROVISIONAL} \SetWatermarkScale{0.55} \SetWatermarkColor[gray]{0.94} \fi

\newcommand{\stem}{\mbox{[stem]}}

\newcommand{\rev}[1]{#1}
\newcommand{\revcolor}{}

\newcommand{\vio}[1]{#1}
\newcommand{\grn}[1]{#1}
\newcommand{\exdiff}[1]{\underline{#1}}

\newcommand{\NcandTotal}{282}
\newcommand{\NratedTotal}{225}
\newcommand{\NratedRoundOne}{225}
\newcommand{\NratedRoundTwo}{\prov{[n]}}
\newcommand{\NratedMoral}{89}
\newcommand{\NratedNorm}{96}
\newcommand{\NratedNormKind}{24}
\newcommand{\NratedDescr}{40}
\newcommand{\NreserveTotal}{57}

\newcommand{\NraterCount}{\prov{four}}       
\newcommand{\NraterCountCap}{\prov{Four}}
\newcommand{\TwoThirds}{\prov{three of four}} 
\newcommand{\TwoThirdsN}{\prov{three}}
\newcommand{\HalfRule}{\prov{at most one}}   
\newcommand{\NaccTotal}{\prov{[n]}}
\newcommand{\NaccMoral}{\prov{[n]}}
\newcommand{\NaccNorm}{\prov{[n]}}
\newcommand{\NaccNormP}{\prov{[n]}}
\newcommand{\NaccNormE}{\prov{[n]}}
\newcommand{\NaccNormA}{\prov{[n]}}
\newcommand{\NaccNormK}{\prov{[n]}}
\newcommand{\NconfNormP}{\prov{[n]}}
\newcommand{\NconfNormE}{\prov{[n]}}
\newcommand{\NconfNormA}{\prov{[n]}}
\newcommand{\NconfNormK}{\prov{[n]}}
\newcommand{\NaccDescr}{\prov{[n]}}
\newcommand{\NrejTotal}{\prov{[n]}}
\newcommand{\NrejMoral}{\prov{[n]}}
\newcommand{\NrejNorm}{\prov{[n]}}

\newcommand{\MinimaMet}{\prov{met}}

\newcommand{\ExStem}{\prov{[stem]}}
\newcommand{\ExMoralMarked}{\ExMoral}
\newcommand{\ExNormPMarked}{\ExNormP}
\newcommand{\ExNormEMarked}{\ExNormE}
\newcommand{\ExNormAMarked}{\ExNormA}
\newcommand{\ExNormKMarked}{\ExNormK}
\newcommand{\ExDescrMarked}{\ExDescr}

\newcommand{\AlphaQone}{\TBD{$\alpha$ Q1}}
\newcommand{\AlphaQtwo}{\TBD{$\alpha$ Q2}}
\newcommand{\AlphaQthree}{\TBD{$\alpha$ Q3, six-way}}
\newcommand{\AlphaQthreeClass}{\TBD{$\alpha$ Q3, three classes}}

\newcommand{\AlphaQfour}{\TBD{$\alpha$ Q4}}
\newcommand{\NstrictTotal}{\TBD{strict subset size}}
\newcommand{\NcleanTotal}{\TBD{clean-contrast subset size}}
\newcommand{\AlphaGradedOthers}{\TBD{$\alpha$ of the four other ratings}}
\newcommand{\OSFref}{(\url{https://github.com/daviligade/normative-mechanistic-interpretability})}
\IfFileExists{filled/values_ratings.tex}{

\renewcommand{\NraterCount}{five}
\renewcommand{\NraterCountCap}{Five}
\renewcommand{\TwoThirds}{four of five}
\renewcommand{\TwoThirdsN}{four}
\renewcommand{\HalfRule}{at most two}
\renewcommand{\NaccTotal}{227}
\renewcommand{\NaccMoral}{80}
\renewcommand{\NaccNorm}{107}
\renewcommand{\NaccDescr}{40}
\renewcommand{\NaccNormP}{37}
\renewcommand{\NconfNormP}{15}
\renewcommand{\NaccNormE}{22}
\renewcommand{\NconfNormE}{18}
\renewcommand{\NaccNormA}{24}
\renewcommand{\NconfNormA}{22}
\renewcommand{\NaccNormK}{24}
\renewcommand{\NconfNormK}{24}
\renewcommand{\NrejTotal}{15}
\renewcommand{\NrejMoral}{9}
\renewcommand{\NrejNorm}{6}

\renewcommand{\MinimaMet}{met}
\renewcommand{\NratedTotal}{242}
\renewcommand{\NratedRoundTwo}{17}
\renewcommand{\NratedRoundOne}{225}

\renewcommand{\ExStem}{Comforting a crying child at the playground}
\renewcommand{\AlphaQone}{0.977}
\renewcommand{\AlphaQtwo}{0.335}
\renewcommand{\AlphaQthree}{0.738}
\renewcommand{\AlphaQthreeClass}{0.851}
\renewcommand{\AlphaQfour}{0.883}
\renewcommand{\AlphaGradedOthers}{P 0.577, E 0.676, A 0.766, K 0.714}
\renewcommand{\NstrictTotal}{170}
\renewcommand{\NcleanTotal}{210}
}{}   
\IfFileExists{filled/values_results.tex}{
\newcommand{\HOneRatio}{1.00}
\newcommand{\HOneCI}{[0.92, 1.07]}
\newcommand{\HOneBand}{2}
\newcommand{\HOneLevel}{97.5}
\newcommand{\HTwoRatio}{0.90}
\newcommand{\HTwoCI}{[0.84, 0.96]}
\newcommand{\HTwoBand}{2}

\newcommand{\HZeroRatio}{1.12}
\newcommand{\HZeroCI}{[1.05, 1.18]}

\newcommand{\HThreeMDAone}{1.19 [1.11, 1.26]}
\newcommand{\HThreeMDB}{1.09 [1.02, 1.16]}
\newcommand{\HThreeNDAone}{1.19 [1.13, 1.26]}
\newcommand{\HThreeNDB}{1.21 [1.14, 1.28]}
\newcommand{\HFourMet}{4 of 7}
\newcommand{\HSixGroupSize}{46}
\newcommand{\HFiveFrameDriven}{14}

\newcommand{\HSixMoreThanNorm}{12}
\newcommand{\HSixMoreThanNormB}{3}
\newcommand{\HSixConditions}{14}

\newcommand{\NecessityABeats}{8 of 14}
\newcommand{\NecessityBBeats}{12 of 14}
}{}   
\IfFileExists{filled/values_examples_marked.tex}{
\renewcommand{\ExMoralMarked}{\exdiff{Comforting} a crying child at the playground / \exdiff{Mocking} a crying child at the playground}
\renewcommand{\ExNormPMarked}{\exdiff{Inspecting} the tent before the first camping trip / \exdiff{Forgetting} the tent before the first camping trip}
\renewcommand{\ExNormEMarked}{\exdiff{Greeting} the host when you arrive at a party / \exdiff{Avoiding} the host when you arrive at a party}
\renewcommand{\ExNormAMarked}{\exdiff{Trimming} the hedges in front of a historic mansion / \exdiff{Hacking} the hedges in front of a historic mansion}
\renewcommand{\ExNormKMarked}{\exdiff{Verifying} a surprising statistic before citing it / \exdiff{Trusting} a surprising statistic before citing it}
\renewcommand{\ExDescrMarked}{The harvest after a season of \exdiff{steady rain} / The harvest after a season of \exdiff{severe drought}}
}{}   
\usepackage{hyperref}
\usepackage{url}
\usepackage{myacronyms}
\usepackage{mytodonotes}
\usepackage[nameinlink,noabbrev]{cleveref}
\crefname{appendix}{appendix}{appendices}
\Crefname{appendix}{Appendix}{Appendices}
\hypersetup{colorlinks=true,linkcolor=blue!45!black,citecolor=blue!45!black,urlcolor=blue!45!black,
pdftitle={Disambiguating Moral Circuits and Moral Features in Language Models: Moral Content, Normative Judgment or Cue Words?
Evidence from 14 Models},   pdfauthor={Davide Liga, Liuwen Yu, Matteo Magnini}}
\graphicspath{{figures/}}
\newcolumntype{Y}{>{\raggedright\arraybackslash}X}

\title{Disambiguating Moral Circuits and Moral Features in Language Models:\\
\large Moral content, normative judgment or cue words?
Evidence from 14 models}
\author{\name Davide Liga \email davide.liga@uni.lu \\
      \addr University of Luxembourg
      \AND
      \name Liuwen Yu \email liuwen.yu@list.lu \\
      \addr Luxembourg Institute of Science and Technology (LIST)
      \AND
      \name Matteo Magnini \email matteo.magnini@uni.lu \\
      \addr University of Luxembourg}

\begin{document}
\maketitle
\provbanner

\begin{abstract}
\vio{Interpretability methods can locate the attention heads and \ac{SAE} features that change the answer that a language model gives to a moral prompt, for example whether ``Stealing medicine for a dying child is morally'' continues with ``justified'' or ``wrong''.
Such a component could respond to moral content, to the judgment of conduct by any norm, or to the word ``morally'' in the prompt.
We built \NaccTotal{} minimal pairs of short phrases, validated by \NraterCount{} annotators, whose two versions differ in a moral standard, in another kind of norm (prudence, etiquette, aesthetics or evidence) or in a descriptive quality, and showed them as text to complete to 14 open-weight models (Llama, Gemma and Qwen, 1B to 14B parameters, base and \ac{IT}, the latter without their chat format), with and without the word ``morally''.
In every model, the heads that change the moral answer belong to an evaluation circuit that the models use for every kind of normative judgment that we tested: these heads carry the same share of the effect for moral as for other normative pairs (pooled ratio \HOneRatio{}, \HOneLevel\,\% \ac{CI} \HOneCI, with the word; \HTwoRatio{}, \HTwoCI, without it), and the models need these heads as much for moral as for other normative pairs.
The models need these heads also when nothing in the prompt asks for a judgment, and on 400 scenarios of the ETHICS benchmark, where removing these heads takes away 37 to 39\,\% of the difference between acceptable and unacceptable scenarios, against 1 to 3\,\% for ten heads chosen at random.
The \ac{SAE} features that we located respond mostly to the word ``morally'' rather than to what the phrase is about.
The models do represent the category of a pair (meaning, the standard that the annotators assigned to it): this category can be read from their hidden states, and, in the eight models that we tested, removing the direction that carries the category weakens the moral verdict by a few per cent\grn{, and more than the other verdicts in seven of them}.
We release the pairs, the ratings, the code, the trained \ac{SAE} dictionaries and the result files, so that others can test the components they find on the same contrasts.}
\end{abstract}
\acresetall 

\section{Introduction}
\label{s:intro}

Ask a language model to complete ``Stealing medicine for a dying child is morally'' and it may prefer ``justified'' to ``wrong''.
Internal components of the model (attention heads, for example) carry that preference, and interpretability methods can find them: they copy the activity of a component from one prompt into a slightly different prompt, remove it, or add to it, and see whether the answer moves.
Finding these components is only the first step.
The second is to ask what they do.
Does the model have machinery of its own for moral judgment, or does it judge a theft (a moral matter) with the same machinery as a rude remark (a matter of etiquette) or a risky shortcut (a matter of prudence)?
Moral psychology has asked the same question of people for decades, and has answered it both ways: moral judgment as a domain of its own \citep{turiel1983development}, or as one application of a general capacity for evaluating conduct \citep{haidt2001emotional,sripada2007framework}.
A mechanistic study can put this question to a language model, whose components can be located, removed and measured.
To answer it, the study needs the right controls.

\begin{figure}[t]
\centering
\includegraphics[width=0.9\linewidth]{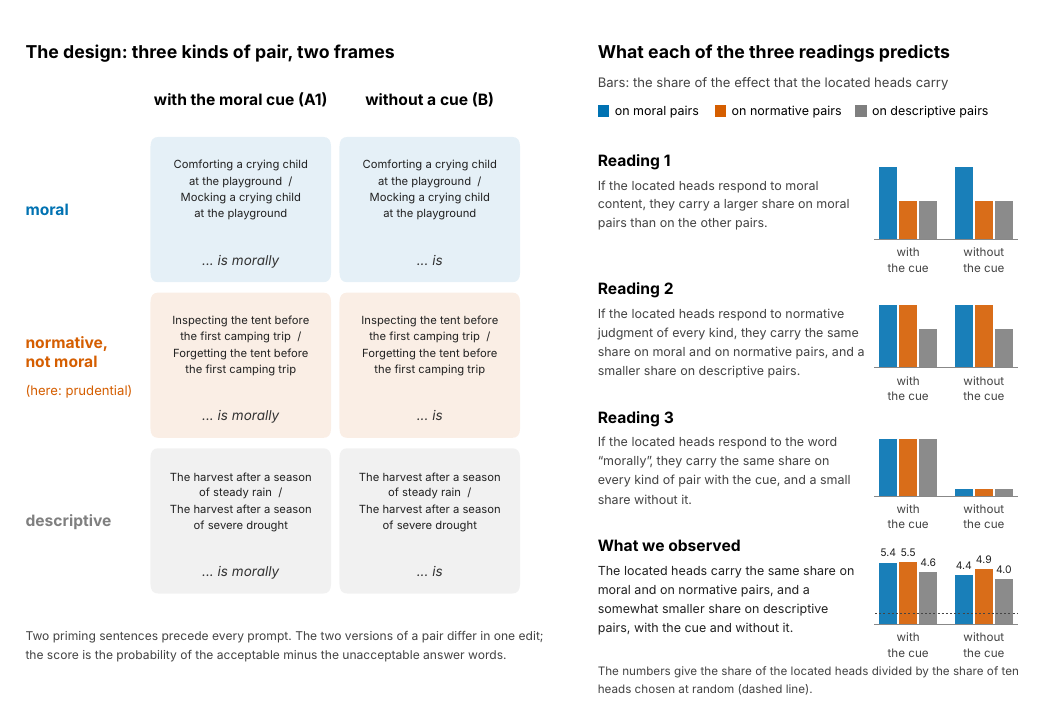}
\caption{\textbf{The design and the three readings.}
Left: three kinds of pair (the examples are retained pairs of this study), each shown in the frame with the moral cue (A1) and in the frame without a cue (B).
Right: \rev{what each of the three readings predicts for the share of the effect that the located heads carry on each kind of pair (schematic bars), and what we observed (last row: geometric means over the models; \Cref{s:heads}).
\vio{The observed shares fit Reading~2 better than Readings~1 and~3, but they do not single out normativity: without the cue, phrases in which something good or bad happens to a person, with nobody's conduct judged, receive the share of normative pairs; with the cue, normative pairs keep a somewhat larger share (\Cref{s:heads}).}}}
\label{fig:design}
\end{figure}

The difficulty is that a prompt such as ``Stealing medicine for a dying child is morally'' has three properties at once.
It is about morality; it asks for a judgment of someone's conduct, as prompts about prudence (skipping the safety check is unwise), manners (interrupting the speaker is rude), taste (the mural is ugly) and evidence (the conclusion is unjustified) also do; and it contains the word ``morally''.
A component that moves the answer could respond to any of the three: to moral content, to normative judgment of every kind, or to that word.
Each reading describes a different organisation of the model and predicts a different pattern on prompts that vary one property at a time (\Cref{fig:design}).
Prompts of one kind cannot tell the readings apart.

This paper builds the controls and uses them to describe how moral judgment is organised in 14 open-weight models (Llama-3.2-1B and 3B, Llama-3.1-8B, Gemma-2B and Qwen2.5-3B, 7B and 14B, each in base and \ac{IT} form).
We first located the components on 73 moral minimal pairs ending in ``is morally'', with 32 descriptive control pairs: every model has attention heads whose output, copied from one version of a pair into the other, moves the moral answer, and each of 14 \ac{SAE} dictionaries \rev{(trained on seven of the 14 models, so not one per model)}, which \rev{approximate} the model's \rev{hidden} state as a \rev{weighted} sum of a few features out of a large learned set, holds features that separate moral from neutral prompts (\Cref{s:study1}).
We call the ten heads that this search identified as moral in each model the \emph{located heads}\rev{, and the features of each dictionary that separate moral from neutral prompts the \emph{located features}}.
We then wrote \NcandTotal{} new minimal pairs of three kinds: moral pairs; normative pairs of another kind, which differ in a prudential, etiquette, aesthetic or epistemic standard and have the same form as the moral pairs; and descriptive pairs in which nobody's conduct is judged.
\rev{Moral pairs are normative too; for short, we call the pairs of the second kind the normative pairs.}
\NraterCountCap{} annotators, who did not know which category a pair had been written for, rated \NratedTotal{} of these pairs, and their ratings decided which pairs count: \NaccTotal{} were retained\rev{.
The annotators found the retained normative pairs hardly moral: on a scale from 0 to 3, the mean moral rating is 0.2 for the normative pairs and 2.8 for the moral pairs}.
Every pair is shown in two frames (a frame is the fixed text around a pair's phrase, the \emph{stem}): one ends in a moral cue, the word ``morally'' (``\stem{} is morally''), and the other has no cue (``\stem{} is''), so that the three readings above make different predictions (\Cref{fig:design}).
The hypotheses, the analyses and the reading of each possible outcome were \grn{written down} before any model saw a new pair, so that the analysis could not be fitted to the results (\Cref{s:design}).

\paragraph{\rev{Terms.}}
\rev{At each token of a prompt (a word or a piece of a word) and after each layer, a model holds a vector, the \emph{hidden state} (also called the residual stream); its answer is read from the hidden state of the last token.
Each layer changes the hidden state through \emph{attention heads}, which bring in information from earlier tokens, and through \emph{neurons}, the units of its feed-forward (\ac{MLP}) block.
A \emph{direction} is a vector in the space of hidden states; it can be added to a hidden state or removed from it.
A \ac{SAE} is trained on the vectors that one layer of one model produces and approximates each of them as a weighted sum of a few vectors out of a large learned set; we call a trained \ac{SAE} a \emph{dictionary} and its vectors \emph{features}.
A \emph{circuit} is a set of components (here, attention heads) with what they read, what they write and what feeds them; the \emph{pathway} is the larger set of components through which the model forms its verdict on a pair (heads, neurons and a direction), of which a circuit is one part.
One version of a pair is the better one by the standard of the pair (comforting, not mocking, a crying child); the model's \emph{preference} for the better version is how much more its answer favours that version than the worse one.
The \emph{effect} of a head on a pair is how far the answer to one version moves when the head's output is copied from the other version, and the \emph{share} of a set of heads is its part of the sum of the effects of all heads.}

\paragraph{What we found.}
\rev{The} models judge a theft, a risky shortcut and a rude remark with the same machinery.
In every one of the 14 models, the components that move a moral answer are those of a general evaluation pathway.
Part of this pathway is a circuit of attention heads, the located heads of each model.
The located heads carry the same share of \rev{the effect} for moral and for normative pairs, with the cue (\rev{their share on moral pairs divided by their share on normative pairs, pooled over the 14 models, is} \HOneRatio{}, \HOneLevel\,\% \ac{CI} \HOneCI) and without it (\HTwoRatio{}, \HTwoCI)\rev{, so the located heads treat a moral pair like a pair of any of the four other standards (prudential, etiquette, aesthetic or epistemic).
The located heads carry about a fifth more of the effect for normative than for descriptive pairs.
\rev{At least half of this difference does not depend on a norm, and separates phrases about a person from phrases about a thing: on 58 further descriptive pairs about a person, in which something good or bad happens to someone and nobody's conduct is judged, the located heads carry the same share as on normative pairs without the cue, and a share halfway between normative pairs and pairs about a thing with the cue (\Cref{s:heads}).}}
The models need the located heads: removing their output takes away a fifth to a quarter of the preference, as much on normative as on moral pairs.
\rev{The models need the located heads also when nothing in the prompt asks for a verdict, when a plain question asks for one, and for scenarios of the ETHICS benchmark, which we did not write (\Cref{s:heads}).}
The located heads read the cue and the words that differ between the two versions of a pair, in the same way \rev{on moral and on normative pairs.
The layer at which the preference for the better version of a pair takes shape is about the same for moral and for normative pairs, and later with the word ``morally'' in the prompt than without: that layer depends on the cue and does not depend on what the pair is about.}
\rev{At the level of single neurons, the} same neurons shift with the judgment \rev{for moral and for normative pairs.
One direction, added to the hidden state at the last token of the prompt,} moves moral, normative and descriptive judgments alike in all 14 models.
\rev{Where these components make a difference, it is between the moral and normative pairs on one side and the descriptive pairs \vio{about a thing} on the other,} not between moral and other norms\rev{.
This means that the same parts of the model produce a moral verdict and any other verdict on someone's conduct\vio{, and also a verdict on what is good or bad for a person when nobody's conduct is judged}: the second of the three readings\vio{, in a wider form}} (\Cref{s:heads}; \Cref{fig:across-models}).

\rev{Yet} the models can tell which standard is at stake: shown ``\stem{} is a matter of'' after five example sentences, every model gives the highest probability, among the words for the five standards, to the annotators' standard 81 to 92\,\% of the time\rev{.
In eight models (those up to 3B parameters), a Jacobian lens, which \rev{estimates at each layer the model's next-word probabilities \vio{through a linear approximation of the remaining layers} (loosely, what the model is about to say)}, tells moral from the other normative pairs} even when the prompt asks only whether something is good or bad.
\rev{This information about the standard feeds the moral judgment a little: removing from the hidden state of one layer the direction that the lens associates with the moral words, or exchanging it for the direction of another standard, lowers the model's preference for the better version of a moral pair by 2 to 11\,\% in seven of the eight models, and its preference for the better version of the other pairs by less or not at all.
Removing the dictionary features or the few hundred neurons that separate moral from normative pairs has a similar effect in some models (5 to 10\,\%).
In their answers, the models} often weigh a harm to others more than the same harm to oneself (\Cref{s:beyond}).
The \rev{located features} are \vio{mostly} detectors of the word at the end of the prompt, ``morally'' or ``ethically'', \vio{and respond much less to} what the pair is about\rev{.
With the cue in the prompt,} more features tell moral from normative pairs than tell one kind of norm from another \rev{in 12 of 14 dictionaries (in 3 without a cue, and in at most 2 with a non-moral cue, ``socially'' or ``logically''; \Cref{s:features})}.

\paragraph{Contributions.}
\rev{The paper makes \vio{five} contributions.}
\begin{list}{(\arabic{enumi})}{\usecounter{enumi}\setlength{\leftmargin}{1.7em}\setlength{\labelwidth}{1.3em}\setlength{\labelsep}{0.4em}\setlength{\itemsep}{0pt}\setlength{\parsep}{0pt}\setlength{\topsep}{2pt}\setlength{\partopsep}{0pt}}
\item \rev{\vio{The located heads form a general evaluation circuit, which the models need for moral and for other normative pairs alike, also when the prompt does not ask for a verdict and for scenarios of the ETHICS benchmark, which we did not write (\Cref{s:heads}).
In all 14 models (from 1B to 14B parameters, in three families, before and after instruction tuning), the located heads, the neurons that shift with the judgment and the shared direction behave in the same way for moral and for normative pairs.
Information about the category of a pair (meaning, the standard that the annotators assigned to it) can be read from the hidden state at the last position, and, in the eight models that we read through a lens, removing from the hidden state the direction that carries this information weakens the moral verdict by a few per cent (\Cref{s:beyond}).}}
\item \rev{We present a design that makes the question ``is this component moral'' testable: normative control pairs of the same form as the moral pairs, shown in the same frame with and without the moral cue, and validated by people rather than by language models (\Cref{s:design}).}
\item \rev{What sets moral pairs apart \vio{does not show in the share of the located heads; it shows} (\Cref{s:beyond}) in the information about \vio{the category of a pair}, which every model has (asked what a pair is a matter of, the models give the highest probability to the annotators' standard for 81 to 92\,\% of the prompts), \vio{in features that, with a moral cue in the prompt, tell moral from the other normative pairs more than they tell one kind of norm from another (in 12 of 14 dictionaries, and in 3 without a cue),} and in the models' answers, which often weigh a harm to others more than the same harm to oneself.
\vio{The located features mostly detect the word at the end of the prompt rather than what the pair is about (\Cref{s:features}).}}
\item \rev{We report three methodological results (\Cref{s:study1}).
Comparing the raw effects of heads overstates how specific a head is, because almost every head has a larger effect on moral prompts; comparing shares of the effect removes this.
Control prompts in a different frame overstate specificity as well, because the difference between the frames is counted as moral content; controls in the same frame remove this.
And components tested on the pairs on which they were selected look more specific than they are; selecting them on separate pairs removes this.}
\item \rev{We release the pairs, the ratings, the code, the trained \ac{SAE} dictionaries and the result files \OSFref.}
\end{list}

\section{Related work}
\label{s:related}

Developmental psychology distinguishes moral from conventional transgressions by their perceived seriousness, generality and independence from authority \citep{turiel1983development}, while intuitionist accounts and accounts of the psychology of norms treat moral judgment as one form of a broader capacity for normative evaluation \citep{haidt2001emotional,sripada2007framework}.
Our categories follow this contrast; we take no position in the debate and use its distinctions to build controls that ask for a meaningful judgment without asking for a moral one.

Behavioural benchmarks measure what a model answers to ethical questions \citep{hendrycks2021aligning,emelin2021moral,scherrer2023evaluating}, not which components produce the answer.
Mechanistic studies have begun to ask that, and they label components as moral in several ways: \citet{schacht2025mapping} call neurons moral when their activation differs between moral and immoral statements, with neutral statements as the control, and ablate them; \rev{\citet{liga2025neurons} find the \ac{MLP} neurons whose activation at the last token differs most between the two versions of a moral minimal pair (two prompts that end in ``is morally'' and call for opposite judgments) and steer the judgment with them, with neutral prompts as the control (our pairs follow this construction);} \citet{su2025value} find ``value neurons'' from activation differences between opposite descriptions of a value and edit them; \citet{li2026indifference} call \ac{SAE} features moral when they correlate with moral dimensions of a labelled corpus, and steer with them; \citet{yu2026tracing} trace moral foundations across 14 models with directions, \ac{SAE} features and steering, and contrast them with social norms; \citet{reblitzrichardson2026organize} reports a common component shared by representations of moral concepts; \citet{dasdan2026frame} label neurons from their automatically generated descriptions.
In most of these\rev{,} a component counts as moral because it differs between moral and immoral or neutral inputs, or because intervening on it changes a moral output; of those cited, \rev{only} \citet{yu2026tracing} \rev{compare} moral content with another kind of norm.
Our question is narrower and causal: whether the heads whose replacement moves a moral answer carry a larger share of the effect for moral than for normative contrasts of the same form, and whether that depends on a moral word in the prompt.

Activation patching is a standard tool for locating the components that carry a behaviour \citep{vig2020investigating,meng2022locating,wang2023interpretability,conmy2023towards,goldowskydill2023path}; its conclusions depend on how inputs are paired and how the output is scored \citep{zhang2024patching}, and it can produce convincing but misleading localisations \citep{makelov2024subspace}.
Selecting components on the data used to test them inflates effects \citep{kriegeskorte2009circular}.
\acp{SAE} approximate activations with a sparse combination of learned features \citep{bricken2023monosemanticity,templeton2024scaling,gao2024scaling,he2024llamascope,marks2024sparse}; a feature is a coordinate of the approximation, not an established concept, and whatever the \ac{SAE} fails to reconstruct is invisible to it, which is why we report, for every \rev{dictionary}, how much of the difference between the two versions of a pair it retains.

\section{Models, pairs, frames and plan}
\label{s:design}

The design follows from the three readings \rev{of \Cref{fig:design}}.
To tell moral content from normative content in general, the moral pairs need a comparison set that asks for a judgment of the same kind, in the same words, about something other than morality\rev{.
The normative pairs are that set: each is judged by a normative standard that is not specifically moral}.
To make sure that any difference is one of content and not of form, the normative pairs have the same length as the moral pairs, an edit of the same size in the same position and the same sentence templates, and every pair is shown in the same frame.
To tell content from the cue word, every pair is also shown with the cue removed.
And to make sure that the pairs are what we think they are, people who do not know our labels judge them before any analysis.
Every pair is thus a small experiment that holds everything fixed except one thing: the content of the edit, or the presence of the cue word.

\paragraph{Models.}
Llama-3.2-1B, Llama-3.2-3B, Llama-3.1-8B, Gemma-2B and Qwen2.5-3B, 7B and 14B, each in base and \ac{IT} form; the \ac{IT} models receive raw text without a chat template.
The \rev{feature} analyses use the 14 \rev{dictionaries} on which the features were located: TopK dictionaries \rev{(a TopK dictionary keeps, for each vector that it encodes, the $k$ features with the largest values and sets the others to zero)} trained for this project on the \rev{hidden states} of Llama-3.2-1B and 3B (base and \ac{IT}, several layers and sparsities) and on the output of the first \ac{MLP} block of Llama-3.2-1B, the Llama Scope dictionary at layer 12 of Llama-3.1-8B \citep{he2024llamascope} \rev{(a public dictionary, the only one of the 14 that we did not train; its authors trained it as a TopK dictionary and released it in a converted form, in which a feature is active when its value exceeds a threshold)}, and TopK dictionaries at layer 0 of Qwen2.5-7B, base and \ac{IT} (\Cref{tab:sae-conditions}).
\rev{The dictionaries are not one per model, and that there are 14 of them, like the models, is a coincidence: they belong to seven of the models (five to Llama-3.2-1B and its \ac{IT} version, six to Llama-3.2-3B and its \ac{IT} version, one to Llama-3.1-8B and one to each of the two Qwen2.5-7B models).
Each dictionary is used only on the model on whose hidden states it was trained.
Training a dictionary takes considerable compute, so only some models have one: we trained dictionaries for six models, and for the base Llama-3.1-8B we used the public Llama Scope dictionary.}

\paragraph{Pairs.}
Each item is a minimal pair of short phrases (a \emph{stem}): a better (permissive) and a worse (restrictive) version that differ in one to three tokens (\Cref{tab:examples}).
In a \emph{moral} pair the two versions differ in harm, fairness, honesty, consent, property or loyalty.
In a \emph{normative} pair they differ in a non-moral standard: prudential (the agent's own interests), etiquette (manners and convention), aesthetic (\rev{how well something is designed, made or performed}) or epistemic (the quality of reasoning or evidence).
In a \emph{descriptive} pair they differ in how good a thing or a situation is, and nobody's conduct is judged.
\rev{In each of the three frames and for each of the 14 models, the better and the worse version of a pair have the same number of tokens, so that the two can be compared position by position (\Cref{s:study1}).
No stem contains a word that the score counts as an answer, none shares half of its content words with another pair, and moral and normative pairs use the same four sentence templates in similar proportions (what changes is the first verb, a function word, an object or modifier, or the purpose or reason at the end).
Of the \NcandTotal{} candidates that pass these checks, \NratedTotal{} were rated by the annotators (\Cref{sec:items}).}

\paragraph{Validation.}
\NraterCountCap{} annotators, \rev{all fluent in English, rated the \NratedTotal{} pairs (\NratedMoral{} written as moral, 113 as normative and \NratedDescr{} as descriptive).}
They were told neither the category a pair had been written for, nor that the pairs formed groups, nor what the study expected.
For each pair they said which version is better; how much worse the worse one is; whether anyone's action, choice or work is judged at all \rev{(if not, the pair counts as descriptive for that annotator); and, if so,} how much the difference is a matter of each of the five standards \rev{(moral, prudential, etiquette, aesthetic, epistemic)}, each rated separately from 0 to 3.
\rev{The retention rule, fixed before any rating was collected, was that} a pair stays if at least \TwoThirds{} annotators chose the intended better version and at least \TwoThirds{} saw the intended class of standard (moral; any non-moral standard; none), with a median moral rating of at least 2 for \rev{the \NratedMoral{} moral pairs and at most 1 for the 113 normative and the \NratedDescr{} descriptive pairs}.
Of the \NratedTotal{} rated pairs, \NaccTotal{} were retained: \NaccMoral{} \rev{of the \NratedMoral{}} moral, \NaccNorm{} \rev{of the 113} normative (\NaccNormP{} prudential, \NaccNormE{} etiquette, \NaccNormA{} aesthetic, \NaccNormK{} epistemic) and \rev{all} \NaccDescr{} descriptive, so every \rev{required} minimum was met.
Agreement was $\alpha=$~\AlphaQone{} for the better version and \AlphaQthree{} for the class of standard \vio{derived from each annotator's answers (\AlphaQthreeClass{} at the level of the three classes that the retention rule uses)}.
\rev{The \NaccMoral{} retained moral} pairs received a mean moral rating of 2.8 and at most 0.2 on any other standard; \rev{the \NaccNorm{} retained normative pairs received a mean moral rating of 0.2, and 61 of them received 0 from every annotator (so they can be considered \emph{mostly} non-moral);} the \NrejTotal{} rejected pairs were ones in which a second standard seemed as relevant as the intended one \rev{(\Cref{app:validation})}.
\rev{Excluding these \NrejTotal{} pairs ensures that no pair counted as moral was seen mainly as a matter of another standard, and that no pair counted as normative was seen mainly as a moral matter.}

\begin{table}[t]
\centering
\caption{\textbf{Pairs and frames.}
One retained new pair per category (better version / worse version), and the three frames in which every pair is shown.
The model's next-token probabilities at the final position are recorded.
\rev{Underlined: the words that differ between the two versions of a pair, and the words that follow the phrase in each frame.}}
\label{tab:examples}
\small
\begin{tabularx}{\linewidth}{@{}p{0.27\linewidth}Y@{}}
\toprule
Category & Pair \\
\midrule
Moral & \ExMoralMarked \\
Prudential & \ExNormPMarked \\
Etiquette & \ExNormEMarked \\
Aesthetic & \ExNormAMarked \\
Epistemic & \ExNormKMarked \\
Descriptive & \ExDescrMarked \\
\midrule
Frame & Prompt \\
\midrule
A1 (primary, cue present) & A sunny day at the beach is good.
A flat tire on the highway is bad.
\ExStem{} \exdiff{is morally} \\
B (primary, no cue) & A sunny day at the beach is good.
A flat tire on the highway is bad.
\ExStem{} \exdiff{is} \\
A0 (secondary, the original frame\rev{, as in \citealp{liga2025neurons}}) & \ExStem{} \exdiff{is morally} \\
\bottomrule
\end{tabularx}
\end{table}

\paragraph{Frames and score.}
Every category is shown in the same three frames (\Cref{tab:examples}).
In \textbf{A1} two fixed priming sentences precede ``\stem{} is morally''; in \textbf{B} the same sentences precede ``\stem{} is''; in \textbf{A0} ``\stem{} is morally'' stands alone, as on the original pairs.
So A1 and B differ only in the cue word, and A1 and A0 only in the priming.
\rev{A1 and B are the \emph{primary} frames: the two main tests of the paper (H1 and H2, \Cref{s:heads}) compare moral with normative pairs in them.
A0, the frame of the original pairs, is \emph{secondary}: it is analysed in the same way, as an additional check.}
The priming is there because a bare ``\stem{} is'' would rarely be completed with an evaluative word: on the original pairs with Llama-3.2-1B, a frame without ``morally'' put 2 to 16\,\% of the next-token probability on evaluative words without priming and 31 to 63\,\% with it.
The model's answer is read from the next-token probabilities at the final position: the score $S$ sums the probability of 14 positive answer words (acceptable, justified, right, good, ...) and subtracts that of 14 negative ones (unacceptable, wrong, bad, ...), all single tokens in the three tokenizers.
The \emph{behavioural gap} of a pair, $S(\text{permissive}) - S(\text{restrictive})$, is the model's judgment of the pair in one number, and every effect of a component in this paper is a change of this gap.
A pair enters a model's analysis in a frame only if its gap is at least 0.02, so that there is a preference for the components to carry.
\vio{Unless a passage says otherwise, every number is computed on $S$.
Two other measures appear in a few analyses, and each is named where it is used: \ac{LD}, the logarithm of the summed probability of the positive words minus the logarithm of that of the negative words, for prompts in which the answer words receive little probability (\Cref{s:heads}); and, when the model chooses between two options labelled A and B, the logit of ``A'' minus the logit of ``B''.}
Throughout, ``with the cue'' means frame A1 and ``without the cue'' frame B.

\paragraph{Plan.}
Before any model was run on the new pairs, we wrote down the hypotheses, the frames and answer words, the head and feature sets to be tested, the retention rule, the analyses and the interpretation bands.
We then ran every model on all candidate pairs, retained or not, so that the runs did not depend on the ratings; which pairs enter the analysis was decided by the annotators alone, who saw no model output, and no result was read before the retained list was fixed.
Later analyses were written down with their readings before they ran and are marked exploratory (\Cref{app:prereg}).

\section{Locating the components}
\label{s:study1}

Before \rev{any} controls can be applied, there must be components to apply them to.
They were found on the original pairs: 73 moral minimal pairs ending in ``is morally'' \rev{(the construction of \citealp{liga2025neurons})} and 32 descriptive control pairs ending in \rev{a neutral phrase such as} ``is typically'', later joined by 189 \rev{normative} pairs and 73 descriptive pairs for the features (\Cref{app:study1}).

\paragraph{Heads that move the moral answer, measured on shares.}
For every pair we run the \rev{\emph{worse}} version, but with \rev{the output of one attention head} copied from \rev{the run on} the \rev{\emph{better}} version\rev{, and we record by how much the score of the worse version changes (the score is the probability of the positive answer words minus that of the negative ones, as in \Cref{s:design}).
This change is the \emph{effect} of that head on that pair \rev{(this is activation patching; \citealp{vig2020investigating,meng2022locating,wang2023interpretability,zhang2024patching})}.
The effect is positive when the copied output brings the model's answer closer to its answer for the better version.
The copying is repeated for every head of the model, one head at a time, so each pair has one effect per head.
We call these effects \emph{raw} effects as long as nothing is done to them.}
Every model has heads whose effect is large on moral pairs.
The natural first test \rev{asks, for each head, whether its} mean absolute effect is larger on moral than on control pairs\rev{.
This test (a permutation test, corrected for the number of heads) finds a significant difference for about half of the heads or more in every base model (47 to 100\,\% of them; 442 of the 512 heads of Llama-3.2-1B, for example),} because raw effects are larger on moral pairs for almost every head: for 86 to 100\,\% of the heads of each base model.
The reason is not that most heads are moral.
\rev{The effects of all heads are larger on the moral pairs: averaged over all heads, the raw effect on moral pairs is 1.5 to 2.3 times the raw effect on control pairs in the base models (16 times in Qwen2.5-3B).
A comparison of raw effects mostly measures this difference between the two sets of pairs.
Almost every head shares it, and we call it the \emph{offset} (more details in \Cref{fig:raw-vs-share}, \Cref{app:study1}); a pilot described below shows that it is not a matter of moral content.
To remove the offset, we express the effects of each pair as \emph{shares}: the absolute effect of each head is divided by the sum of the absolute effects of all heads on that pair,} so that every pair distributes a total of one among the heads\vio{; a share compares heads on the same pair and is not a part of the total effect, since the effects of single heads, each measured with the other heads in place, need not add up to the effect of all heads together}\rev{.
Something survives this step.
In each model, the \emph{top head} is the head for which the first test finds the largest difference between moral and control pairs.
Measured on shares, the top head still has the largest difference of all heads, and the difference is still significant, in 11 of the 14 models (the exceptions are the base model of Gemma-2B and the two Qwen2.5-7B models).
In these 11 models the mean share of the top head on moral pairs is 1.9 to 6.7 times its mean share on control pairs.
In the five base models among them this ratio is 2.4 to 3.8, while the same ratio computed on raw effects is 4.6 to 7.1 (and 87 in Qwen2.5-3B), so the offset accounts for a part of the raw ratio and not for all of it (\Cref{tab:dev-heads} in \Cref{app:study1}).
The \emph{located heads} of a model are the ten heads for which the first test finds the largest differences, the top head being the first of them.
They are} the \rev{head set of the primary tests}.
\rev{Measured on shares, the ten located heads together carry 1.3 to 2.1 times the share on moral pairs that they carry on control pairs in 13 of the 14 models (0.7 times in the base model of Qwen2.5-3B, where the effects on control pairs are very small).}

\paragraph{\rev{Base and \ac{IT} models.}}
\rev{Instruction tuning changes little in how the effect is divided among the heads.
The mean shares of all heads of a model on the moral pairs form the \emph{profile} of that model, and the profile of each base model correlates with the profile of its \ac{IT} version at 0.87 to 0.97.
What instruction tuning changes is the size of the raw effects on moral pairs.
Summed over all heads, the raw effects are smaller in the \ac{IT} version by 48\,\% in Llama-3.2-1B, 46\,\% in Qwen2.5-3B, 33\,\% in Llama-3.2-3B and 13\,\% in Gemma-2B.
In the three larger models the size changes less and not always in the same direction (14\,\% smaller in Llama-3.1-8B, 13\,\% larger in Qwen2.5-7B, 20\,\% larger in Qwen2.5-14B), and the two profiles correlate at 0.94 to 0.97 (\Cref{tab:dev-base-it} in \Cref{app:study1}).}

\paragraph{Controls in the same frame, and a mirror image.}
\rev{The original controls are 32 pairs about how good a thing or a situation is: the weather in Paris in summer and in winter, the safety of a reinforced and of a condemned building.
Like the prompts that \citet{liga2025neurons} used as a control, they are about neutral matters; here they are written as minimal pairs.
A control pair ends in a neutral phrase (``is typically'', ``is generally'', ``is considered'' and the like), and each pair has two lists of answer words of its own, fixed when the pair was written: \emph{warm}, \emph{hot} and \emph{pleasant} against \emph{cold}, \emph{cool} and \emph{chilly} for the weather, \emph{excellent} and \emph{great} against \emph{poor} and \emph{terrible} for the building.
These lists belong to the original controls only; the new pairs of this paper all share the answer words of \Cref{s:design}.
The original controls therefore differ from the moral pairs in three things at once: the frame (a neutral phrase instead of ``is morally''), the answer words (two lists for each pair instead of the same verdict words for every pair) and the content (a thing or a situation instead of someone's conduct).
A pilot in Llama-3.2-1B removed the first two differences: it} compared the moral pairs with 81 normative pairs in the same ``is morally'' frame \rev{and with the same answer words.
In this comparison the offset defined above is absent: averaged over all heads, the raw effect on moral pairs is 1.03 times the raw effect on the normative pairs, against 2.25 times the raw effect on the original controls in the same model.
The offset is therefore not a matter of moral content.
It comes with what sets the original controls apart from the moral and from the normative pairs, which is their frame, their answer words or their descriptive content; the pilot cannot tell these three apart.}

\rev{The pilot showed a second thing, which matters for every comparison in the rest of the paper.
When the effects of each pair are expressed as shares, 210 of the 512 heads of Llama-3.2-1B carry a significantly larger share on the moral pairs than on the normative pairs.
The reason is that the moral pairs and the normative pairs do not divide the effect among the heads in exactly the same way (their mean shares across the heads correlate at 0.76), as is to be expected of any two sets of pairs.
This has a consequence for heads that are selected on one set.
The heads with the largest shares on a set of moral pairs can be expected to carry a somewhat larger share on other moral pairs than on normative pairs, simply because they were chosen where moral pairs put their effect, even if they do not respond to moral content at all.
From here on we measure a set of heads by a ratio: the share that the set carries on moral pairs divided by the share that it carries on normative pairs.
A ratio of 1 means the same share on both, and a ratio above 1 a larger share on moral pairs.
For heads that were selected on moral pairs, a ratio above 1 can thus come from moral content or from the selection alone.
To tell the two apart, we repeat the selection the other way round: heads are selected in the same way on normative pairs, and for them the share on normative pairs is divided by the share on moral pairs.
We call this second selection the \emph{mirror image} of the first.
If the heads selected on normative pairs favour normative pairs as much as the heads selected on moral pairs favour moral pairs, the excess comes from the selection; only an excess beyond that of the mirror image would point to moral content (\Cref{s:heads} reports both).}

\paragraph{Features that separate moral from other prompts, and what they follow.}
A \ac{SAE} \rev{approximates} the \rev{hidden state of} one layer as a \rev{weighted} sum of a few features out of a large dictionary \rev{(a feature is a direction in the space of hidden states, and its weight for a prompt is its activation; one of the 14 dictionaries is trained on the output of an \ac{MLP} block instead, \Cref{s:design}).
The search for features used the 73 original moral pairs, 221 normative pairs in the same ``is morally'' frame (the 32 original ones and the 189 added) and 105 neutral pairs (the 32 original controls and the 73 added descriptive pairs).
Each model keeps the pairs on which it prefers the better version: 54 to 71 moral, 172 to 210 normative and 64 to 83 neutral pairs.
The search compares the moral pairs with one comparison set at a time, first the neutral pairs and then the normative pairs.
For each feature of a dictionary it asks whether the activation of the feature changes more between the two versions of a moral pair than between the two versions of a pair of the comparison set, and it keeps the features for which this difference is significant (a permutation test, corrected for the number of features of the dictionary).
With the neutral pairs as the comparison set, the search keeps between 37 and 222 features in each of the 14 dictionaries; these are the located features of that dictionary.
With the normative pairs as the comparison set, it keeps between 16 and 124 features in each dictionary.
In every dictionary, then, more features tell moral pairs from neutral pairs than tell moral pairs from normative pairs: 3.9 to 18.4 times as many when both are counted at the same size of effect and with the same number of pairs in the two comparison sets.
One could read this as follows: most of the features that tell moral pairs from neutral pairs respond to something that moral and normative pairs have in common and neutral pairs lack, which could be the judgment of someone's conduct by a norm, so that \emph{the features mostly encode normativity}.
A test in Llama-3.2-1B shows that this reading is wrong.
The moral and the normative pairs have something else in common, the frame ``is morally'', and the test exchanged the frames: it replaced the last word of the neutral prompts by ``morally'' and the last word of the moral prompts by ``typically''.
The 91 located features of a dictionary of that model (at layer 7) follow the last word and not the content.
Their mean activation is 0.150 on the moral prompts and 0.108 on the normative prompts; on the neutral prompts it is 0.001 in their own frame and 0.096 once they end in ``morally''; on the moral prompts it falls to 0.007 once they end in ``typically''.
The difference between the two counts therefore measures the frame, not the content, and the hypothesis H5 of \Cref{s:features} tests on the new pairs whether the located features follow the word ``morally''.}

\rev{The dictionaries have a second limit.
They describe each prompt fairly well but capture little of what changes between the two versions of a pair: on the original pairs, the difference between the two reconstructions explains little of the difference between the two hidden states ($R^2$ between $-0.27$ and $0.38$ over the 14 dictionaries, with a median of 0.04; a value of 1 would mean that the dictionary captures the whole difference).
Two rules follow for the rest of the paper.
When we remove features (\Cref{s:features}), we change only the part of the hidden state that the dictionary reconstructs and leave the remainder, called the reconstruction error, as it is \rev{(as \citealp{marks2024sparse} do)}, so that the model keeps what the dictionary does not capture.
And the causal claims of the paper rest on head patching (the copying of head outputs described at the start of this section and used in \Cref{s:heads}), not on features.}

\paragraph{Selection on separate pairs, and labels checked by people.}
\rev{Heads that are selected on a set of pairs and then tested on the same pairs look better than they are, because a selection favours the heads whose values on these particular pairs are high partly by chance.
In this section, every estimate of how much the located heads or the located features prefer moral pairs comes from the pairs on which they were selected, and is therefore too large.
This is why the located heads (ten in each of the 14 models) and the located features are tested from \Cref{s:heads} on with the \NaccTotal{} new pairs.
None of the new pairs was used when the located heads and the located features were selected.
The 262 pairs that were added to the original sets (189 normative and 73 descriptive) had their labels checked by three language models and by an author.
The \NratedTotal{} new pairs that were rated were judged by \NraterCount{} annotators who did not know the intended labels (\Cref{s:design}; \Cref{app:audit}).}

\section{The heads: an evaluation circuit for every norm}
\label{s:heads}

\paragraph{Measurement.}
For every pair we run the \rev{\emph{worse}} version, but with one head's output copied from the \rev{\emph{better}} version; \rev{the resulting change of the score $S$ of the worse version (defined in \Cref{s:design})} is that head's effect on the pair \rev{(as in \Cref{s:study1}).
For pair $i$ and head $h$ we write this effect $e_{ih}$ (\Cref{eq:head-effect} in \Cref{app:study1})}.
Because the two versions have the same number of tokens, positions align exactly.
Each pair's absolute effects are divided by their sum over all heads, so that every pair distributes a total of one\rev{: the \emph{share} of head $h$ on pair $i$ is} $\mathrm{share}_{ih} = |e_{ih}| / \sum_{h'} |e_{ih'}|$\rev{, where the sum runs over all heads $h'$ of the model}; \rev{a pair on which the effects are large as a whole} does not make every head look large, and what counts is how \rev{the total} is divided among the heads.
The head set of the primary tests holds the located heads of each model (\rev{the ten heads defined in} \Cref{s:study1}).
The selectivity of a head set in a model and frame is the ratio of its mean share on moral pairs to its mean share on normative pairs, $\rho$; the pooled value $\exp\theta$ is the geometric mean of $\rho$ over the 14 models.
Both equal 1 when the heads carry the same share for moral and for normative pairs.
Intervals are percentile bootstrap intervals over items (10{,}000 resamples), per-model tests permute category labels with Benjamini-Hochberg correction across models, and the analysis keeps, per model and frame, the pairs with a behavioural gap of at least 0.02 (57 to 80 moral and 80 to 107 normative pairs per model and frame), so that all 14 models enter both primary tests (\Cref{app:head-method}, \Cref{tab:heldout-heads}).

\paragraph{H1 and H2.}
\textbf{H1}: with the cue, the located heads carry a larger share of the effect for moral than for normative contrasts; \textbf{H2}: the same without the cue.
Each is tested at $\alpha=0.025$ with a two-sided 97.5\,\% interval of $\exp\theta$, read against \rev{five bands}: an interval entirely above 1.5 is a strong moral preference, one entirely inside 0.8 to 1.25 is no meaningful content specificity (if it also excludes 1, this is said and the difference is described as within that margin), one entirely above 1 a weak to moderate preference, one entirely below 1 a smaller share for moral content, anything else inconclusive (\Cref{tab:bands}).
\vio{The margin of 0.8 to 1.25}\grn{, part of the analysis plan (\Cref{app:prereg}),} \vio{is symmetric on the logarithmic scale (a quarter more, or a fifth less)}\grn{; the same range is a common convention for the equivalence of a ratio of two means \citep{fda2001bioequivalence}.
Every ratio is reported with its interval}.

\begin{figure}[t]
\centering
\includegraphics[width=0.9\linewidth]{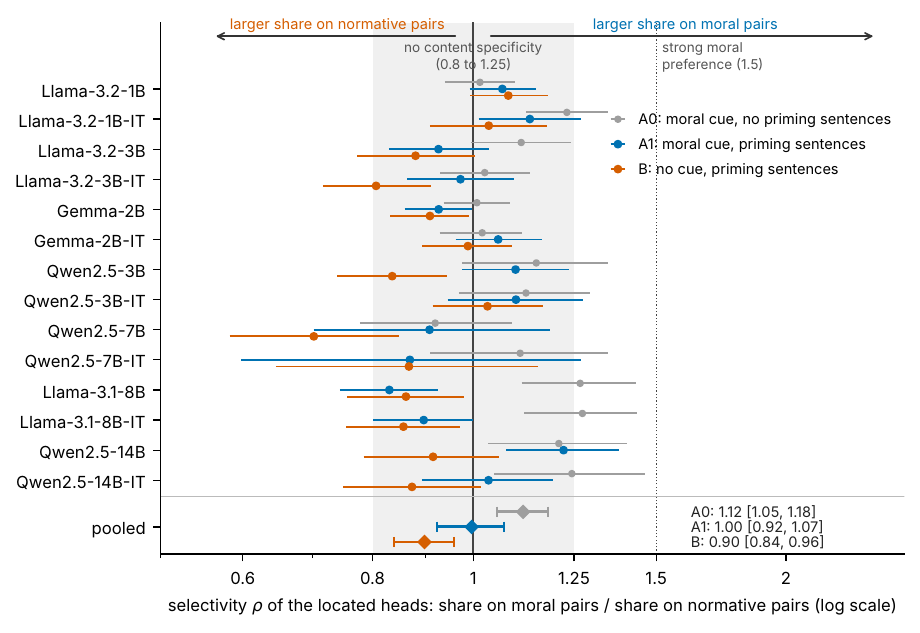}
\caption{\textbf{The located heads on the new pairs (H1, H2).}
Per-model selectivity $\rho$ of the located heads with 95\,\% intervals in frames A1 (\rev{moral cue after the priming sentences;} blue), B (\rev{priming sentences, no cue;} orange) and A0 (\rev{moral cue without the priming sentences;} grey), and the pooled ratios with their 97.5\,\% (A1, B) and 95\,\% (A0) intervals; \rev{a point to the right of 1 means that the located heads carry a larger share of the effect on moral than on normative pairs, a point to the left a smaller one.
The shaded band (0.8 to 1.25) marks differences that we treat as too small to call the heads specific to moral content: inside it, the share on moral pairs is at most a quarter larger, and at most a fifth smaller, than the share on normative pairs.
The dotted line at 1.5 marks the start of a strong moral preference (\Cref{tab:bands}).}
Every pooled interval lies inside the band; the per-model point estimates lie inside it too, except \vio{three: Qwen2.5-7B in frame B (0.70) and the two Llama-3.1-8B models in frame A0 (1.27 each)}, and no model approaches the line of strong preference at 1.5.}
\label{fig:h1h2}
\end{figure}

The located heads carry the same share for moral as for normative pairs, with and without the cue.
With the cue the pooled ratio is \HOneRatio{} (\HOneLevel\,\% \ac{CI} \HOneCI); without it \HTwoRatio{} (\HTwoCI); both intervals lie inside the margin of 0.8 to 1.25, and the second also excludes 1, so without the cue the located heads carry a slightly smaller share for moral pairs, a difference within that margin (\Cref{fig:h1h2}).
In frame A0, with the cue but without the priming, the ratio is \HZeroRatio{} (95\,\% \ac{CI} \HZeroCI)\rev{: inside the same margin and above 1, so with the cue alone the located heads carry a slightly larger share for moral pairs (the interval lies above 1 in 5 of the 14 models), a difference that is absent with the priming sentences \rev{and about two thirds as large when ``socially'' stands in the place of ``morally'' (1.08, 95\,\% \ac{CI} [1.02, 1.14], 14 models; Appendix~\ref{sec:further})}\grn{, and as large with the synonym ``ethically'' (1.14, [1.08, 1.21])}}.
No single model shows a strong moral preference either: the per-model ratios range from 0.83 to 1.22 with the cue and from 0.70 to 1.08 without it \rev{(0.92 to 1.27 in frame A0)}.
Of the three readings, the located heads fit the second\rev{.
They do not fit the third: without the word ``morally'' the located heads still carry four to five times the share of ten heads chosen at random (\Cref{fig:design}; on moral and on normative pairs, at least twice in every model), and the models still need them (below).}
The located heads do the same share of the work when a model judges a prudential, an etiquette, an aesthetic or an epistemic pair as when it judges a moral one.

\paragraph{\rev{A fifth more for pairs about a person than for pairs about a thing (H3).}}
A head set that treats moral and normative pairs alike could still be a set for evaluation of any kind, including judgments about things.
The same ratio for moral against descriptive pairs is \HThreeMDAone{} with the cue and \HThreeMDB{} without it, and for normative against descriptive pairs \HThreeNDAone{} and \HThreeNDB{}: the located heads carry about a fifth more of the effect for normative than for descriptive contrasts\rev{, and for moral contrasts a fifth more with the cue and a tenth more without it.
On descriptive pairs the located heads still carry four times the share of ten heads chosen at random (\Cref{fig:design}).
The difference of a fifth has two possible sources: all 40 descriptive pairs are noun phrases about a thing (``The taste of ripe strawberries'') and the other pairs are phrases about someone's action, so the located heads may carry more because conduct is judged or because of the form of the phrase.
\rev{To separate the two sources we wrote 66 further descriptive pairs with a person in them, in the sentence templates of the moral and normative pairs: something good or bad happens to the person, and nobody's conduct is judged (``Winning the raffle at the village fair'' against ``Losing the raffle at the village fair'').
Three language models answered the annotators' questions for these pairs and kept 58 of them (an exploratory analysis on the 14 models; \Cref{sec:further}).
Without the cue, the located heads carry the same share on the 58 pairs about a person as on the normative pairs (ratio of normative to person 1.00, 95\,\% \ac{CI} [0.92, 1.09]) and 1.22 times their share on the descriptive pairs about a thing ([1.11, 1.32]): the whole difference lies between phrases about a person and phrases about a thing, and it does not require that conduct is judged.
With the cue, the pairs about a person lie halfway between the two (normative against person 1.11, [1.04, 1.19]; person against thing 1.08, [1.01, 1.15]).}}

\paragraph{The same under every planned variation (H4 and sensitivity analyses).}
Instruction tuning preserves most of the pattern across heads on the new moral pairs (profile correlation 0.88 to 0.96 in six \vio{of the seven base models} and 0.78 in Qwen2.5-7B, against split-half reliabilities of 0.94 to 0.99), although the base model's top head stays in the instruct top ten in only \HFourMet{} \vio{base models.
The two largest models keep the pattern best: the correlation is 0.95 in Llama-3.1-8B and 0.96 in Qwen2.5-14B, and the located heads, searched separately in each model, have 8 and 6 of their ten heads in common between the base and the \ac{IT} versions of these two models, against 2 to 5 in the five smaller ones (seven base models; a description, which we did not test)}.
The sensitivity analyses (other patching settings and scores, four other head sets, moral against each kind of norm separately, comparisons at equal behavioural gap and at equal rated size, stricter rating subsets, another inclusion threshold, all rated items) keep the pooled share ratio inside the band of the primary result, 0.95 to 1.12 with the cue and 0.86 to 1.03 without it, and adjusting the share for the behavioural gap, the template and the token counts moves the moral-minus-normative difference slightly below zero (mean $-0.08$ with the cue, $-0.13$ without it; \Cref{tab:sensitivity}).
None of these changes makes the located heads look moral.

\paragraph{\vio{Heads chosen on the new pairs show no moral excess beyond the control.}}
Perhaps the located heads were not the right ones.
Ten heads chosen on a random half of the new moral pairs give a moral/normative ratio on the other half of 1.04 with the cue and 0.99 without it (geometric means over 200 splits and 14 models); the mirror image, ten heads chosen on half of the normative pairs, gives a normative/moral ratio of 1.01 and 1.12.
\vio{With the cue, neither selection is more selective for its own category than the other.
In the other two frames there is a small asymmetry, in opposite directions: without the cue, heads chosen on normative pairs keep an advantage on fresh pairs (1.12 against 0.99), and with the cue but without the priming sentences (frame A0), heads chosen on moral pairs do (1.18 against 0.93).
A} selection aimed directly at the excess of moral over normative share does no better once it is compared with the same selection run the other way round and with selections that separate two kinds of norm from the other two (\Cref{sec:heads-discovery})\vio{: the moral selectivity of single heads does not exceed what a split between two kinds of norm shows}.

\paragraph{The circuit.}
Four further analyses ask whether the located heads could be a moral circuit after all: a set that the model needs for moral pairs but not for normative ones, that reads something only moral pairs contain, or that is fed differently on moral pairs.
In every case the answer is the general one (\rev{exploratory analyses;} \Cref{sec:circuits}).
\emph{The model needs them.}
Replacing the output of the located heads by its mean at every position \rev{(mean-ablation; \citealp{wang2023interpretability})} removes a median of 23\,\% of the gap on moral pairs with the cue and 18\,\% without it, more than every one of 20 random sets of ten heads \vio{(drawn from all other heads, without regard to layer)} in 10 of 14 models in each frame\grn{, and more than every one of 20 random sets drawn from the same layers as the located heads in 10 and in 8 models};  on normative pairs the same ablation removes 28\,\% and 21\,\%, and the difference between moral and normative pairs is negative or indistinguishable from zero in every model with the cue and in 12 of 14 without it.
\emph{They read the cue and the edit.}
At the final position a median of 28\,\% of their attention falls on ``is morally'' (15\,\% for the average head) and 19\,\% on ``is'' without the cue (8\,\%); they attend more than the average head to the edited tokens and to the rest of the stem, and much less to the first token; these shares are the same on moral and on normative pairs within 0.02.
What they write directly to the answer \rev{(direct logit attribution; \citealp{elhage2021mathematical})} is small (a median of 2.2\,\% of the logit difference with the cue), so they act through later components.
\emph{\rev{The layer depends on the cue, not on the content.}}
Patching the \rev{hidden state} layer by layer, the layer at which the final position first recovers half of the gap is the same for moral and normative pairs in 7 of 14 models with the cue and within one layer in six of the other seven, and the same in 11 of 14 without it; what moves \rev{that layer is the cue}, from about half of the depth without the cue (0.53) to two thirds with it (0.67).
\emph{The same inputs feed them.}
Edge attribution patching into the located heads' query, key and value inputs finds the signal arriving in roughly equal measure at the edited tokens, the rest of the stem and the cue, with no difference between moral and normative pairs beyond a few points.
The located heads are better named a \rev{\emph{normative/evaluative}} circuit than a moral one.

\paragraph{\rev{Other prompts, and text that we did not write.}}
\rev{Three properties of the analyses above could limit what they show: the three frames ask the model for a good-or-bad word (through the cue, through the priming sentences, or through both); we wrote every pair; and the judgment is read from evaluative words.
Further analyses vary each property.
They cover the 14 models and are exploratory (\Cref{sec:further}, \Cref{tab:further}).
Because the answer words receive little probability in some of these prompts, \vio{the analyses of the three prompts below} measure the judgment as the log-odds of the 14 positive against the 14 negative answer words \vio{(\ac{LD}; \Cref{s:design}); the scenarios of ETHICS keep the score $S$, and the choice between two options is read from the logits of the labels ``A'' and ``B''}.
\emph{A prompt that asks nothing.}
When the prompt is the phrase followed by ``is'' and nothing else (``\stem{} is''), the answer words receive 2 to 3\,\% of the probability, and the models still lean to the positive words for the better version (in 94 to 97\,\% of the pairs).
In this prompt the located heads carry 5.0 times the share of ten random heads on moral pairs and 4.8 times on normative pairs (ratio of the two shares 1.05, 95\,\% \ac{CI} [0.99, 1.11]), and removing the located heads takes away 24\,\% of the gap on moral and 22\,\% on normative pairs (random sets of ten heads: 3\,\%), more than every random set in 13 and in 12 of the 14 models.
\emph{A plain question.}
With ``Question: Is the following good or bad?
\stem{} Answer: It is'', the located heads carry 4.5 and 5.3 times the random share, and removing the located heads takes away 30\,\% and 35\,\% of the gap.
\emph{Another cue.}
With ``socially'' in the place of ``morally'' (``\stem{} is socially''), the located heads carry 5.3 and 4.8 times the random share, and removing the located heads takes away 37\,\% and 35\,\% of the gap, so the located heads do not belong to the word ``morally''.
\emph{Scenarios that we did not write.}
We took 400 one-sentence scenarios from the ETHICS benchmark \citep{hendrycks2021aligning}, 200 labelled acceptable (``I trained my enormous dog to shake hands with people'') and 200 labelled unacceptable (``I smothered the crying baby in my arms'').
Every model scores the acceptable scenarios higher than the unacceptable ones (lowest \ac{AUC} 0.73; medians 0.89 to 0.93 in three prompts), and removing the located heads takes away a median of 37 to 39\,\% of the difference between the two classes (random sets: 1 to 3\,\%), more than every random set in 10 to 13 of the 14 models.
The models therefore need the located heads when nothing in the prompt asks for a verdict, and for sentences that we did not write.}

\rev{\emph{A choice between two options.}
One limit remains.
We showed the two versions of a pair as ``Option A'' and ``Option B'', followed by ``I choose Option'', and read the answer from the labels ``A'' and ``B'', so that neither the prompt nor the answer holds an evaluative word.
The models choose the better version (for a median of 90\,\% of the pairs).
Removing the located heads takes away a median of 24\,\% of the gap on moral and 21\,\% on normative pairs (random sets: 5\,\% and 7\,\%), but more than every random set in only 6 and 7 of the 14 models, and in two \ac{IT} models the gap grows.
In a control with the same options, which asks which option holds a given word, the located heads take away about what random sets take away (6 to 7\,\% against 5 to 6\,\%).
When the judgment is a choice between two options, some models need the located heads and others do not.}

\section{The features: what they follow}
\label{s:features}

\paragraph{Measurement and capture.}
A \ac{SAE} \rev{approximates} the \rev{hidden state of} one layer as a \rev{weighted} sum of a few ``features'' out of a large dictionary.
Heads are components of the computation; features are coordinates of a description of the \rev{hidden} state, and they are only as good as that description.
For each \rev{dictionary} we encode the \rev{hidden} state at the final position of each prompt and record, for every feature, the size of its change between the two versions of a pair.
Before reading any feature result we ask how much of the pair difference the dictionary sees at all: with $d$ the difference between the \rev{hidden} states of the two versions and $\hat d$ the same difference between the two reconstructions, the capture is $R^2 = 1 - \lVert d - \hat d\rVert^2 / \lVert d\rVert^2$.
It is low everywhere: with the cue the mean $R^2$ is below zero in 13 of the 14 \rev{dictionaries} and 0.38 in the layer-19 dictionary of Llama-3.2-3B-\ac{IT}; without the cue it is 0.10 to 0.49.
The dictionaries reconstruct each prompt fairly well but see little of what changes between the two versions of a pair, so the results below describe what the dictionaries capture, not what the models lack.
\vio{Every intervention keeps the reconstruction error in place (\Cref{app:feature-method}), so that removing features does not also remove the part of the hidden state that the dictionary fails to reconstruct; this does not make the features cover more of the pair difference.}

\paragraph{The located features follow the word (H5).}
If the \rev{located features} respond to the word ``morally'', they should fire on descriptive pairs almost as much as on moral pairs when the word is present and fall silent for every category when it is absent; if they respond to moral content, they should fire on moral pairs in both frames and little on descriptive pairs.
The endpoint is the ratio of their mean activation on descriptive pairs to that on moral pairs with the cue: above 0.5 the set follows the frame, below it the content.
The ratio is above 0.5 in all \HFiveFrameDriven{} \rev{dictionaries}, from 0.70 [0.68, 0.72] in Llama-3.2-1B to 0.99 in its \ac{MLP}-0 dictionary, and the same sets are nearly silent without the cue: on moral pairs their activation is 1 to 33\,\% of its value with the cue.
These features light up for ``The harvest after a season of steady rain is morally'' almost as much as for a moral pair, and go dark for a moral pair without the word ``morally''.
Removing them with the cue, with the reconstruction error kept, changes the gap for moral, normative and descriptive pairs alike.
The smaller sets chosen against normative pairs behave differently: their descriptive-to-moral ratio is 0.09 to 0.40 in 11 \rev{dictionaries}, so they prefer moral content, as they were selected to, in dictionaries that capture little of the difference.

\paragraph{A distinction that appears with the moral \rev{cue} (H6).}
The new pairs also allow a search for features that separate moral from normative content.
Such a search needs a yardstick, because any two groups of pairs differ in something: the yardstick is the normative pairs themselves.
If moral pairs are just one more kind of normative pair, telling moral from normative pairs should need about as many features as telling one pair of norm kinds from another (prudential and etiquette from aesthetic and epistemic, say).
For each feature active on at least five prompts we test the moral-versus-normative difference with a permutation test and Benjamini-Hochberg correction, and do the same for the three splits of the four kinds of norm into two against two, at equalised group sizes (\HSixGroupSize{} pairs, 20 subsamples).
With the cue, moral versus normative yields more features than every split among norms in \HSixMoreThanNorm{} of \HSixConditions{} dictionaries: 7 to 60 features at equalised size against at most 1 to 39 for the best split, in the ten dictionaries that detect features at all \vio{(the two Qwen2.5-7B dictionaries, read at the first layer, detect almost no features and count only nominally)}.
Without the cue it is the other way round: in only \HSixMoreThanNormB{} dictionaries, and in the other 11 one of the splits yields as many or more.
So the features at the final token distinguish moral from normative pairs more than they distinguish kinds of norm only when the prompt \rev{carries the moral cue}; without it, the distinction between moral and normative pairs is one normative distinction among others.

\paragraph{These features carry a little of the moral judgment.}
The H6 features are a mixture: features of moral vocabulary in the stems (stealing, torture, a stranger, a child), a few features for abstract moral ideas (justifying and excusing, injustice), and many generic features of function words and punctuation.
Since they were selected on the pairs on which their effect would be measured, we split the pairs in two halves, selected the features on one half and removed them on the other, in both directions.
Without the cue, in 4 of 10 dictionaries the held-out features cut the moral gap by 0.01 to 0.06 (5 to 10\,\% of it) and the normative gap by at most 0.02, with descriptive pairs unaffected (for example Llama-3.2-1B-\ac{IT}: $-0.054$ [$-0.073$, $-0.036$] on moral pairs against $-0.000$ on normative pairs); with the cue the same holds in 2 of 10.
So in the small models the \rev{hidden} state at the final token carries a component specific to moral pairs, with a small causal weight that shows on pairs on which it was not selected and does not need the word ``morally''.

\begin{figure}[t]
\centering
\includegraphics[width=\linewidth]{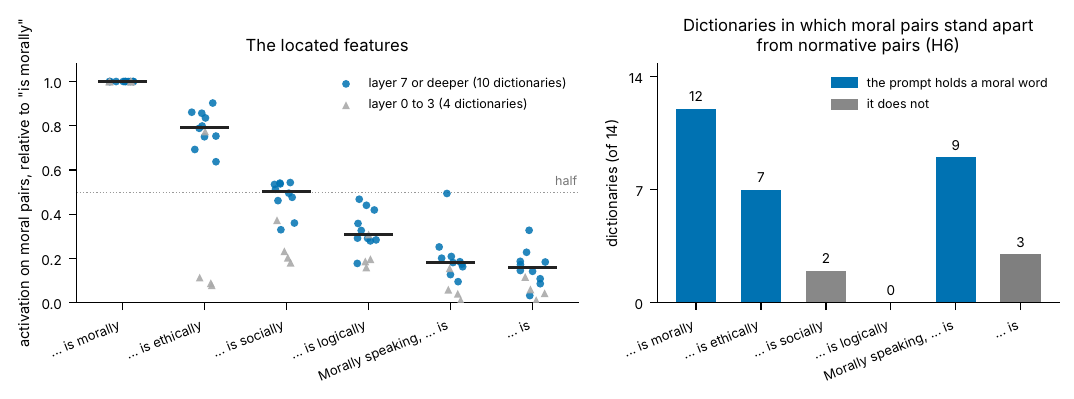}
\caption{\textbf{What the features follow, across six frames.}
Left: the activation of the \rev{located features} on moral pairs in each frame, relative to their activation with ``is morally'' (one point per \rev{dictionary}; black bar, the median of the ten \rev{dictionaries} at layer 7 or deeper\rev{; dotted line, half of the activation with ``is morally''}).
Right: the number of the 14 \rev{dictionaries} in which more features separate moral from normative pairs than separate kinds of norm (H6)\rev{; blue bars, the frames whose prompt holds a moral word}.
The \rev{located features} follow the word at the final position, most of all a moral one; the H6 distinction follows the moral \rev{cue}, wherever it stands.}
\label{fig:cue-frames}
\end{figure}

\paragraph{Swapping the cue word.}
``With the cue'' also means ``with a \rev{word} that suits the moral pairs best'': with ``is morally'' the models put more probability on the answer words, and separate the two versions of a pair more, for moral than for normative pairs and least for descriptive pairs (medians over the 14 models of the behavioural gap: 1.00, 0.69 and 0.37; without the cue 0.71, 0.68 and 0.64).
To see what the features respond to, we ran four further frames on the same pairs \rev{(exploratory)}: a synonym (``\stem{} is ethically''), the moral cue before the stem (``Morally speaking, \stem{} is'', so that the prompt ends in ``is'', as without the cue), and two cues that are not moral, ``\stem{} is logically'' (meant to suit the epistemic pairs) and ``\stem{} is socially'' (meant to suit the etiquette pairs).
\Cref{fig:cue-frames} shows both results at once.
The \rev{located features} respond to ``ethically'' at about four fifths of their level with ``morally'', to ``socially'' at about half, to ``logically'' at about a third, and to ``Morally speaking'' at the start of the sentence hardly more than to no cue at all: they encode the word at the position where they are read, most of all a moral one.
The H6 distinction \rev{instead follows the moral cue wherever it stands}: it appears in 9 of 14 dictionaries when ``Morally speaking'' opens the sentence, in 7 with ``ethically'', in 2 with ``socially'' and in none with ``logically'', against 12 with ``morally'' and 3 without a cue.
Whether a moral \rev{cue} does this because it suits the moral pairs or because it brings out something specific to moral content, these frames cannot settle: ``logically'' did not suit the epistemic pairs in the models' answers, and with ``socially'', which does suit the etiquette pairs, the planned comparison was not informative by its own rule.
Descriptively, ``socially'' sets the etiquette pairs apart from the other norms in 5 of 14 dictionaries, and the frames with ``morally'', with ``logically'' or without a cue in none, so a non-moral \rev{cue} that suits a kind of norm can do for that kind what ``morally'' does for the moral pairs, and part of the moral distinction may come from \rev{how well the cue fits the moral pairs} (\Cref{sec:features-cues}).

\section{\rev{Where moral pairs differ from the other pairs}}
\label{s:beyond}

\begin{figure}[t]
\centering
\includegraphics[width=\linewidth]{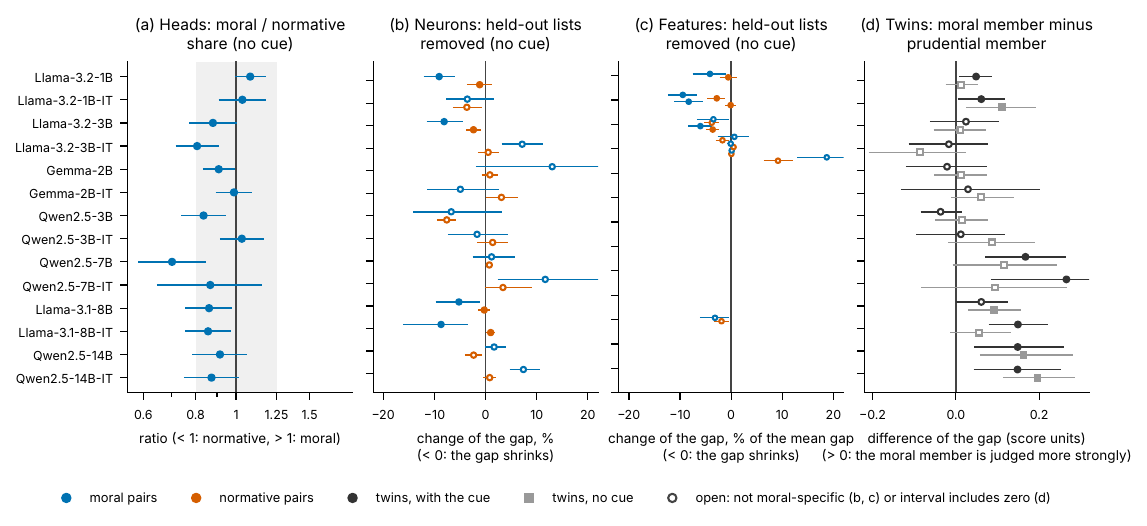}
\caption{\textbf{\rev{Where moral pairs differ from the other pairs, and by how much.}}
Every part shows frame B (no cue), per model.
(a) The located heads: \rev{their share on moral pairs divided by their share on normative pairs,} with its 95\,\% interval \rev{(right of 1: a larger share on moral pairs); the shaded band, as in \Cref{fig:h1h2}, marks no content specificity}.
(b) The moral-versus-normative neuron lists, chosen on half of the pairs and removed on the other half: change of the gap as a percentage of each pair's gap, on moral and on normative pairs, set minus random lists \rev{(left of 0: removing the list makes the preference for the better version smaller)}.
(c) The same for the feature lists of H6, one point per dictionary, as a percentage of the model's mean gap on that category (the layer-19 dictionary of Llama-3.2-3B-\ac{IT} raises the gap).
(d) The models' answers on the twins \rev{(two pairs that differ only in who bears the consequence: other people in the moral member, the agent alone in the prudential member)}: the gap of the moral member minus that of \rev{the prudential member}, with the cue (dark) and without it (light) \rev{(right of 0: the model judges the moral member more strongly)}.
Filled markers mark a fall specific to moral pairs \rev{(the interval of the moral change and that of the moral minus the normative change both lie below zero;} b, c) or an interval above zero (d).
The located heads show no preference; the neurons and features carry a few per cent in some Llama models; the answers tell the twins apart more often than any component does.}
\label{fig:specificity}
\end{figure}

The located heads treat every standard alike.
The analyses of this section \rev{(exploratory)} ask whether that is because the models cannot tell the standards apart, and where, if anywhere, moral content leaves a mark of its own.
\Cref{fig:specificity} puts the sizes side by side; the full account is in \Cref{sec:exploratory}.

\paragraph{The models can tell which standard is at stake.}
We asked it with a frame of its own, ``\stem{} is a matter of'', reading the next-token probabilities of fixed category words (morality or ethics; prudence or safety; manners or politeness; taste or beauty; evidence or logic; fact or chance for the descriptive pairs), with five fixed example sentences that give the label vocabulary.
All 14 models can: the most probable of the five standards is the annotators' standard for 81 to 92\,\% of the prompts (median 0.88; chance 0.20), from Llama-3.2-1B to the 14B models, and the confusions are the natural ones (etiquette with aesthetic, epistemic with moral, moral with prudential).
So every model holds the information that a pair is a moral rather than a prudential or an aesthetic matter, while the located heads treat all standards alike.

\paragraph{\rev{The models hold the standard even when the prompt does not ask for it, and this information moves the judgment a little.}}
A Jacobian lens \citep{gurnee2026workspace} carries the \rev{hidden} state of a layer to the output through the average Jacobian of the remaining layers, so it \rev{estimates, at each layer, the model's next-word probabilities (which is loosely interpretable as showing what the model is about to say at that layer)}.
We fitted it for the eight models up to 3B parameters and read the final position of every prompt of the frame without the cue, which names no standard.
The lens probability of the moral words separates moral from normative pairs with an \ac{AUC} of 0.95 to 0.99 at the best layer, which lies at 0.6 to 0.9 of the depth, and in the Llama and Qwen models the most probable of the five standards is the annotators' standard for 77 to 89\,\% of the pairs whose standard they confirmed, at the best layer.
The lens also gives directions: removing, reversing or exchanging the direction of the moral words at the final position lowers the moral gap by 2 to 8\,\% (in Gemma-2B-\ac{IT} by 9 to 11\,\%, and by 19\,\% when reversed), while the descriptive gap barely moves and the normative gap falls by about half as much or not at all: a fall specific to moral pairs in seven of the eight models\rev{.
The direction is built from the category words through the lens and not from our pairs, so the fall cannot come from having chosen the direction on the pairs on which it is tested}.
\vio{The located heads are not blind to the category of a pair: in all 14 models, the category of a pair can be read from what the located heads write into the hidden state.
Without the cue, a linear classifier trained on that output identifies which of the five standards a pair was written for in 83\,\% of cases (median over the models, with each standard weighted equally), where guessing would be right in one case out of five, and the same kind of classifier reaches the same accuracy (83\,\%) on what ten heads chosen at random write.
Yet the located heads carry the same share of the effect for every category (\Cref{sec:category-heads}).}

\paragraph{\rev{Neurons that separate moral from normative pairs.}}
Counting \ac{MLP} neurons that respond differently to moral and to normative pairs gives tens of thousands per model, but such counts cannot tell a difference of standard from a difference of topic: moral pairs about care differ from the other moral pairs in 13 to 1\,900 neurons, and the splits between kinds of norm do not consistently exceed that.
The causal test on held-out pairs does better.
The 25 neurons per layer with the strongest moral-versus-normative difference in share, chosen on half of the pairs and replaced by their mean on the other half, lower the moral gap and leave the normative and descriptive gaps alone in four models, all Llama, and only without the cue: Llama-3.2-1B (moral $-9$\,\%, normative $-1$\,\%, descriptive $+2$\,\%), Llama-3.2-3B ($-8$, $-2$, $+3$), Llama-3.1-8B ($-5$, $0$, $+1$) and Llama-3.1-8B-Instruct ($-9$, $+1$, $+1$).
The neurons that shift most consistently whenever a pair becomes unacceptable are instead shared: their mean shifts on moral and on normative pairs correlate at 0.70 to 0.94 in the late layers of every model.

\paragraph{One shared direction is enough to move every judgment.}
Adding to the \rev{hidden state}, at three quarters of the depth, the mean difference between the acceptable and the unacceptable versions of normative pairs \rev{(a steering direction built as in \citealp{zou2023representation,rimsky2024steering})} (computed on one half of the pairs and added to the other) raises the evaluation of both versions of a pair by 40 to 240\,\% of the pair's gap, for moral, normative and descriptive pairs alike, with and without the cue, in all 14 models.
The directions computed on moral pairs alone and on normative pairs alone point the same way (cosine 0.72 to 0.94).
The code for stance that the neurons share is not only needed for the judgment but enough to move it, for every kind of pair.
The direction of the moral stems minus the normative stems, by contrast, makes the models call a matter moral when asked (it raises the probability of the moral words in all 14 models) but, added in the frame without the cue, lowers every kind of evaluation in every model rather than singling out the moral pairs.

\paragraph{Where the residue comes from, and what the answers do.}
The feature and neuron residues leave open whether they respond to the edited words or to what moral pairs are about.
We wrote three further sets of pairs, annotated by three language models that first passed a test against our annotators on the 242 rated pairs (they are more lenient than the annotators, so a few of these pairs may be ones people would drop).
In 73 pairs whose two versions differ only in a word with no valence of its own (\emph{with} and \emph{without}, a number), the neuron residue survives in two Llama models, so it is not only a response to the edited words.
\rev{We also wrote \emph{twins}: two pairs that share the scenario, the verb and the edit and differ only in who bears the consequence, other people in the moral member of the twin and the agent alone in its prudential member} (``Driving off without fastening your child's seatbelt'' against ``... your own seatbelt'').
\rev{On 23 twins,} the same neuron lists lower the moral member and \rev{the prudential member} alike; compared twin by twin, the moral member falls more only in the two 8B Llama models without the cue, by one to two points.
The residue follows \rev{the situation that a pair describes (the same harm, whoever bears it)} more than the moral standard.
The models' answers, by contrast, do tell the twins apart: the gap is larger for the moral member than for \rev{the prudential member} in 11 of the 28 combinations of model and frame, and smaller in none.
The models often weigh a harm to others more than the same harm to oneself, a difference that the components we can name reflect at most weakly.
\grn{The models' hidden states tell the twins apart as well.
Within a twin the scenario and nearly all the words are the same, and only who bears the consequence, and with it the standard, differs.
Asked what each phrase is a matter of, the models give more probability to the moral words for the moral member than for its twin in 84\,\% of the twins (13 of the 14 models), and a classifier trained on the main pairs, applied without retraining, picks out the moral member in 90\,\% of the twins (14 of the 14 models; \Cref{sec:twins-standard}).
The information about the category follows the standard, not only the topic.}

\section{Discussion}
\label{s:discussion}

\Cref{tab:summary} puts the results side by side.

\begin{table}[t]
\centering
\caption{\textbf{The results side by side.}
For each part of the model or measurement, what we found and what it means\rev{, intuitively}.}
\label{tab:summary}
\footnotesize
\renewcommand{\arraystretch}{1.0}
\begin{tabularx}{\linewidth}{@{}>{\raggedright\arraybackslash}p{0.2\linewidth}YY@{}}
\toprule
What we examined & What we found & What it means \\
\midrule
The located heads & The same share of the effect for moral and for normative pairs (ratio
\HOneRatio{} with the cue, \HTwoRatio{} without), and about a fifth more for normative than for descriptive pairs \rev{about a thing (on descriptive pairs about a person: the share of normative pairs without the cue, a share halfway with it)} & \rev{They belong to a general evaluation circuit that every standard uses}, not to moral judgment in particular \\
\addlinespace
Removing the located heads & A fifth to a quarter of the preference disappears, at least as much on normative as on moral
pairs & \vio{The models need them for moral, other normative and descriptive pairs alike} \\
\addlinespace
\rev{The located heads in other prompts and on other text} & \rev{With ``\stem{} is'' alone, with a plain question and with ``socially'' as the cue, they carry 4 to 5 times the share of ten random heads on moral and normative pairs, and removing them takes away 22 to 37\,\% of the preference; on ETHICS scenarios, 37 to 39\,\% of the difference between acceptable and unacceptable; in a choice between two options they exceed every random set in 6 and 7 of 14 models} & \rev{The circuit does not depend on a prompt that asks for a verdict, on the word ``morally'' or on our pairs; a choice between two options needs it in some models only} \\
\addlinespace
Heads chosen on the new pairs & Heads chosen on moral pairs favour moral pairs no more than heads chosen on normative
pairs favour normative pairs & \vio{Their moral selectivity does not exceed what a split between two kinds of norm shows} \\
\addlinespace
\rev{The located features} & Active with ``morally'' (past the first layer also ``ethically'') ending the
prompt, \vio{much less dependent on the content (descriptive phrases reach 70 to 99\,\% of the activation of moral phrases)}; nearly silent without a cue or when ``Morally speaking'' opens the sentence, half as
active with ``socially'', a third with ``logically'' & \vio{They respond mostly to the word where they are read, most of all a moral one, and much less to moral content} \\
\addlinespace
Features found on the new pairs & More features separate moral from normative pairs than separate kinds of norm in 12
of 14 dictionaries with the cue, 9 with the \rev{cue} at the start of the sentence, 3 without a cue, 2 with ``socially''
and none with ``logically''; removed, they lower the moral preference by 5 to 10\,\% in some small models & \rev{With the moral cue, the dictionaries} mark moral pairs as a class of their own, possibly in part because that \rev{cue} suits them \\
\addlinespace
Neurons chosen for a moral difference & Without the cue they cut the moral preference by 5 to 9\,\% in four Llama
models; on twins, the prudential version about as much & \rev{They follow the situation that a pair describes (the same harm, whoever bears it) more than morality as such} \\
\addlinespace
Neurons and a direction shared by every judgment & The same neurons shift for moral and for normative pairs; adding
their common direction moves every kind of judgment in all 14 models & \rev{The same neurons and one direction serve every kind of judgment; the judgment needs them, and the direction is enough to move it} \\
\addlinespace
\rev{Information about} the standard & Completing ``is a matter of'', every model gives the highest probability to the annotators'
standard 81 to 92\,\% of the time; unasked, those up to 3B hold it, and removing \rev{or exchanging} \grn{the direction that carries it} cuts the moral preference by 2 to 11\,\% \rev{in seven of the eight} & \grn{The models can tell that a matter is moral; removing the direction that carries this information weakens the moral judgment a little} \\
\addlinespace
Behaviour on twins & The moral version is judged more strongly than \rev{the prudential member of its twin} in 11 of the 28 combinations of
model and frame, and less strongly in none & The answers often weigh a harm to others more than the same harm to
oneself \\
\bottomrule
\end{tabularx}
\end{table}

\paragraph{What the results add up to.}
The components we located form three levels.
At the bottom, the located heads are general evaluation machinery: the model needs them for moral, normative and descriptive pairs alike, they read the cue and the edited words, and \rev{the layer at which the preference takes shape depends on the cue, not on what the pair is about}.
One level up, they \rev{carry} about a fifth more of the effect for normative than for descriptive contrasts \rev{(this difference lies, wholly without the cue and by half with it, between phrases about a person and phrases about a thing, and not between judged conduct and the rest: on descriptive pairs about a person the located heads carry the share of normative pairs without the cue, and a share halfway with it)}.
At the top, they treat moral and other norms alike.
Of the three readings of a ``moral'' head, the located heads therefore fit the second, and not the third: \rev{without the word ``morally'' they still carry four to five times the share of ten heads chosen at random, and the models still need them\rev{; this is so also with ``socially'' in the place of ``morally'', with a plain question, and with the phrase followed by ``is'' and nothing else}}.
\grn{In the components that we measured, the models judge every kind of pair through a shared pathway.
What sets moral pairs apart in these components is small: removing the features, neurons or lens direction chosen for it lowers the moral gap by a few per cent, and on twins the fall caused by the features and neurons follows the situation that a pair describes (the same harm, whoever bears it) more than the moral standard.
The information about the standard is there all the same: on twins, a classifier trained on the main pairs tells the moral member from its prudential twin in the hidden state of all 14 models.}
\rev{The models use the same heads for sentences that we did not write: on scenarios of the ETHICS benchmark, removing the located heads takes away 37 to 39\,\% of the difference between acceptable and unacceptable scenarios.}

\begin{figure}[t!]
\centering
\includegraphics[width=\linewidth]{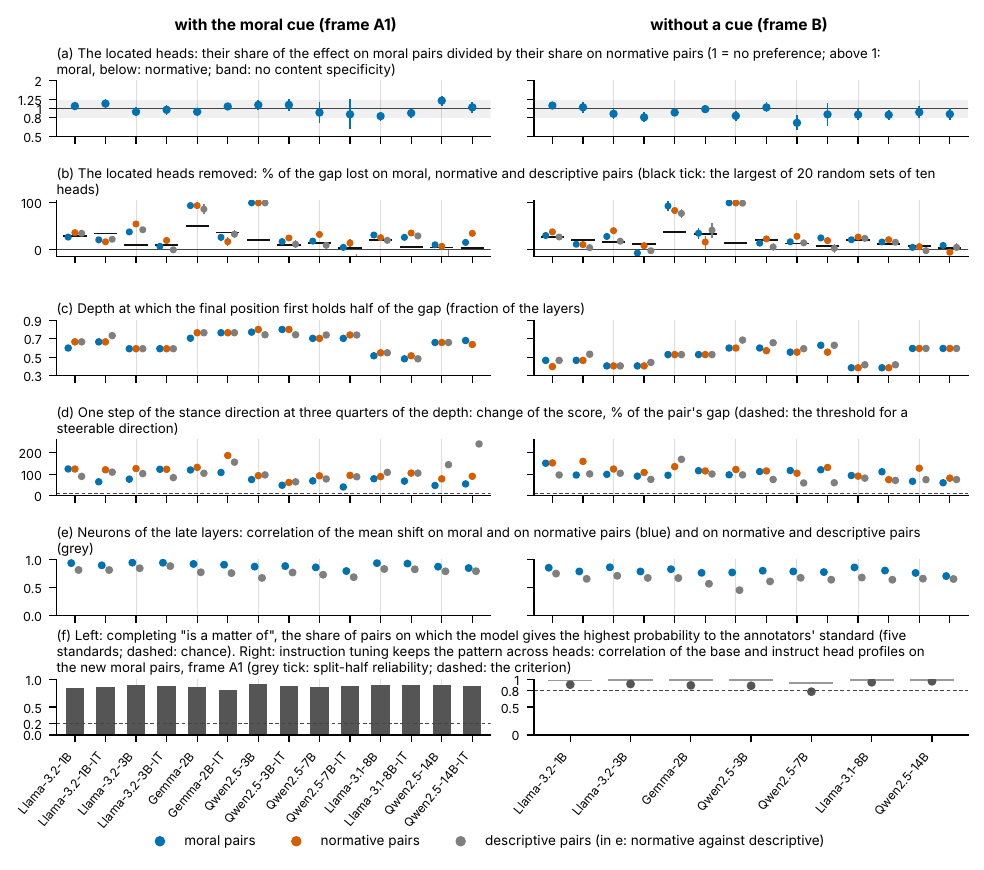}
\caption{\textbf{The same organisation in every model.}
\rev{Each column of points is one model.}
\grn{In no model do these measurements single out moral pairs: the blue points lie with the orange points in (b) to (e), and no interval in (a) lies above the band.}
\vio{In no model does the comparison of moral with normative pairs come out the other way; the interval of the moral/normative ratio lies inside the band of 0.8 to 1.25 in 8 of the 14 models with the cue and in 5 without, and above it in none.}
The shared pathway in each of the 14 models, with the moral cue (left) and without (right); blue, moral; orange, normative; grey, descriptive pairs.
(a) The located heads' share of the effect on moral pairs divided by their share on normative pairs, with its 95\,\% interval and the band of no content specificity; (b) the percentage of the gap lost when the located heads are removed, against the largest loss of 20 random sets of ten heads (\rev{black} tick); (c) the depth at which the final position first holds half the gap; (d) the change of score from one step of the stance direction at three quarters of depth; (e) over the neurons of the late layers, the correlation of the mean shift on moral and normative pairs (blue) and on normative and descriptive pairs (grey); (f) left, the share of pairs on which the model completing ``\stem{} is a matter of'' gives the highest probability to the annotators' standard; right, the correlation of base and instruct head profiles on the new moral pairs (ceiling: split-half reliability).
In every model the moral and normative values lie together in (a) to (e)\rev{, and the cue moves the depth in (c)}.}
\label{fig:across-models}
\end{figure}

\paragraph{The same organisation in every model.}
\Cref{fig:across-models} puts the measurements of the shared pathway side by side for the 14 models.
In every one of them, moral judgment on these pairs is organised in the same way: the located heads \grn{take a similar part in moral and in normative judgments} (per-model ratios 0.83 to 1.22 with the cue and 0.70 to 1.08 without\vio{; the interval of a single model lies wholly inside the margin of 0.8 to 1.25 in 8 of the 14 models with the cue and in 5 without, and wholly above it in none}\grn{; in the other models it crosses one end of the margin, so that model alone does not decide}); removing them costs about as much on normative as on moral pairs; the difference between the two versions of a pair reaches the final position at the same depth for both kinds, and the frame, not the content, moves that depth; the neurons that shift with the judgment are the same for both (correlation 0.70 to 0.94 in the late layers of every model); one direction moves moral, normative and descriptive judgments alike in all 14 models, in both frames; every model gives the highest probability to the annotators' standard when it completes ``is a matter of''; and instruction tuning keeps the pattern across heads in all seven \vio{base models}.
\rev{Where the located heads make a difference, it is between the normative pairs and the descriptive pairs, which are about a thing (the interval of that ratio lies above 1 in 9 of 13 models with the cue and in 10 of 14 without it),} not between moral and other norms.
The models differ in other respects (how fragile the pathway is to mean-ablation in Qwen2.5-3B and Gemma-2B, which head comes first in Qwen2.5-7B, \rev{at which layer} the Gemma models form the five-way distinction among standards; \Cref{sec:circuits}), and in no model does the comparison of moral with normative pairs come out the other way.
The organisation described above is thus the same from 1B to 14B parameters, in three model families, before and after instruction tuning: \vio{a shared} evaluation pathway that every standard uses, and \rev{information about} \vio{the category of a pair (meaning, the standard that the annotators assigned to it), which can be read from the hidden state \grn{and from what the located heads write}, leaves the share of the located heads unchanged and, when its direction is removed from the hidden state, lowers} the moral verdict by a few per cent.
Whether it belongs to the training recipe they share or to language models in general is a question for other families and larger models.

\paragraph{Relation to moral psychology and to prior work.}
\rev{Our results describe components of language models, not human judgment, and take no side between the view of moral judgment as a domain of its own \citep{turiel1983development} and the view of it as one application of a general capacity for evaluation \citep{haidt2001emotional,sripada2007framework}.
What they show is that, at the level of the located heads, a moral pair is treated like a pair judged by any other norm:} the located heads carry the same share of a prudential, an etiquette, an aesthetic or an epistemic answer, and the layer at which the decision arrives does not depend on which standard is at stake.
\rev{Among the components, moral pairs are told apart only at the final token: by more features under the moral cue, and, without the cue, by a few hundred neurons in four Llama models and by the information about the standard that the Jacobian lens reads, each by a few per cent.}
This is compatible with the findings of \citet{yu2026tracing} and \citet{reblitzrichardson2026organize}, who locate moral content in directions and features rather than in dedicated components, and it suggests that what a \rev{dictionary} finds as ``moral'' depends on whether the prompt \rev{already carries the moral cue}.

\subsection{Limitations}
\label{sec:limitations}
\grn{For the heads, the wording of the cue was varied with two further words, the synonym ``ethically'' and ``socially''; with both, the ratio of moral to normative share stays inside the margin (Section~\ref{s:heads}; Appendix~\ref{sec:further})}; for the features, four further wordings and positions were run.
\rev{The three frames of the main analysis ask the model for a good-or-bad word, through the moral cue, through two example sentences about things, or through both.
The located heads carry about as large a share of the effect, and the models need them, when the prompt makes no such request (\Cref{s:heads}).
What remains is that the judgment is read from evaluative words: when a model chooses between the two versions of a pair shown as two options, the located heads remove more than every random set of ten heads in 6 and 7 of the 14 models, so the results about the located heads hold for judgments expressed in evaluative words, \grn{and for a choice in half of the models}.}
A cue suits some pairs better than others: ``is morally'' suits the moral pairs best, and \rev{our tests do not fully separate} this fit from moral content wherever a result depends on the cue.
The items are short English phrases \rev{(and, in one further test, one-sentence scenarios of the ETHICS benchmark)}, and the score is one family of evaluative next-word probabilities; we do not measure free-form judgments, explanations or behaviour.
\vio{The \ac{IT} models receive raw text without a chat template\grn{, the same text as the base models}, so the results describe how they complete a sentence, not how they answer in a conversation.}
\rev{The descriptive pairs of the main analysis are noun phrases about a thing, with no agent and in one sentence template, so the larger share of the located heads on normative pairs is read with the help of the descriptive pairs about a person.}
\grn{The main pairs do not separate the standard from the topic and the wording of the phrases.
\grn{The twins hold the scenario and the edit fixed and change only who bears the consequence, and with it the standard} (we checked for moral against prudential pairs, and there the information follows the standard; \Cref{s:beyond}).
The descriptive pairs about a person and the twins were checked by language models rather than by the annotators: we used very recent models from Anthropic, OpenAI and Google (namely Claude Sonnet~5.5, \ac{GPT}-5.5 and Gemini~3.1~Pro), each of which first passed a test against our annotators.}
The human validation rests on \NraterCount{} lay annotators; their agreement with an intended category shows that people see the contrast we built, not that it fits a classification derived from theory.
The \rev{dictionaries} \vio{cover seven of the 14 models, at one to four layers each} \grn{(training a dictionary takes considerable compute)}, and capture little of the pair difference, so the results about features describe the dictionaries.
\grn{The Jacobian lens was fitted for the eight models up to 3B parameters, because fitting it for the larger models was beyond our computational budget; the classifier on the hidden state, which covers all 14 models, shows the same information (\Cref{s:beyond}).
The lens} describes its linear approximation of the remaining layers.
\grn{The ten located heads are a core of the circuit: removing them takes away a fifth to a quarter of the preference, against about 1\,\% for ten random heads, and the ten heads with the largest share take away more.}
\grn{Random sets drawn from the same layers as the located heads remove little (a median of 3\,\% of the preference with the cue and 2\,\% without), and the located heads remove more than every such set in 10 of the 14 models with the cue and in 8 without.}
\vio{In the control of the choice prompt, which needs no evaluation, the located heads remove more than every random set in only 3 of 14 models.}
Head patching shows whether a head's output can move the score on these contrasts\rev{, and we cannot currently tell what the head exactly computes}.

\section{Conclusion}
\label{sec:conclusion}
The located heads could have been moral, normative in general, or readers of the word ``morally''; on new pairs validated by independent annotators they carry the same share of the effect for moral as for normative contrasts of the same form, with and without the moral cue; they are needed for both; and they read the cue and the edited words, nothing particular to moral content.
\rev{The models need the located heads also when nothing in the prompt asks for a verdict and for scenarios that we did not write.}
The \rev{located features} are \vio{mostly} detectors of a moral word at the end of the prompt, \vio{and respond much less to} the content.
\rev{We located two parts of the models that take part in a moral verdict}: a shared evaluation pathway that \vio{the models used in our tests employ for every good-or-bad judgment}, and \rev{information about} the moral standard, held without being asked, \vio{whose direction, when removed, lowers the moral verdict a little}.
\vio{The pathway} is the same in all 14 models, from 1B to 14B parameters, in three families, before and after instruction tuning.
\rev{What is specific to moral judgment can be located too: \vio{the information about the category of a pair}, features that tell moral from normative pairs under the moral cue, and a few hundred neurons in four Llama models; removing any of them lowers the moral verdict by a few per cent in some of the models, and the other verdicts less or not at all.}

\clearpage
\bibliography{references}
\bibliographystyle{tmlr}

\clearpage
\appendix
\crefalias{section}{appendix}
\section{Locating the components, in detail}
\label{app:study1}
This appendix gives the account on which \Cref{s:study1} draws, in the words of the long version: the original pairs, the heads and the features, with the two measurements that the paper uses.
Every estimate here is inflated by selection, because the heads and features were selected on these pairs.

\subsection{The original pairs}
\label{sec:dev-data}
The components were located on 73 moral minimal pairs ending in ``is morally'' \rev{(the construction of \citealp{liga2025neurons})}, spanning 16 ethical domains and four construction tiers (one word, one token, a phrase or the context changed), and 32 descriptive control pairs ending in \rev{a neutral phrase (``is typically'' in 8 of them, ``is generally'' in 10, ``is considered'' in 8, a similar phrase in the other 6)}, with their own answer words (for example ``Lying to save innocent lives is morally'' against ``Lying to save your public face is morally'', with the control ``The weather in Paris during summer is typically'' against ``... during winter is typically'')\rev{: each control pair has two lists of answer words, mostly words of quality such as \emph{great} and \emph{excellent} against \emph{poor} and \emph{terrible}}.
For the features we added 189 non-moral norm pairs to an original 32 (56 prudential, 53 etiquette, 57 aesthetic and 55 epistemic, all ending in ``is morally'') and 73 descriptive pairs\rev{, which make 105 neutral pairs with the 32 controls (81 of the 105 end in ``is typically'' or ``are typically'')}.
These sets were built for a first search, before the controls of this paper existed, and they differ from the new pairs in form (\Cref{app:audit}).
The added norm pairs mostly contrast whole sentences: 81\,\% share no prefix, and their mean edit is 7.7 tokens against 4.2 for the moral pairs.
Only 15\,\% of the norm pairs and 64\,\% of the moral pairs have the same number of tokens in both versions.
The descriptive controls use another frame and other answer words.
The labels of the expanded sets were checked by three language models (Claude Haiku, Sonnet and Opus), with an author reviewing flagged items and a sample of 30 pairs.

Before the controls can be applied, there must be components to apply them to.
This section describes how they were found, on the original pairs of \Cref{sec:dev-data}: attention heads whose patched output moves the moral answer, and dictionary features that separate moral from other prompts.
It also introduces the two measurements that the rest of the paper uses, and what the original pairs taught about them.
The heads and features were selected on these pairs, so every estimate in this section is inflated by selection; the sections that follow test the components on pairs they have never seen.

\subsection{Heads that move the moral answer}
\label{sec:heads-dev}
\paragraph{Measurement.}
For every pair we run the \rev{\emph{worse}} (restrictive) version, but with one head's output copied from the \rev{\emph{better}} (permissive) version; \rev{the resulting change of the score of the worse version} is that head's effect on the pair \rev{(positive when the copied output brings the model's answer closer to its answer for the better version)}.
The effects of all heads of a model on one pair form a profile.
\rev{Because the effects of all heads together are larger on some pairs and on some sets of pairs than on others, whatever the single heads do (shown below for the original pairs)}, we do not compare raw effects across pairs; we divide each pair's profile by its sum, so that every pair distributes a total of one among the heads.
The share of a head set is then the slice of that total that falls on its heads, and the selectivity of the set for moral content is the ratio of its average slice on moral pairs to its average slice on norm pairs.
The point of the share is that \rev{a pair on which the effects are large as a whole} does not make every head look large; what counts is how \rev{the total} is divided among the heads.
\vio{A share compares heads on the same pair; it is not a part of the total effect, since the effects of single heads, each measured with the other heads in place, need not add up to the effect of all heads together.}

Formally, for a pair $i$ and an attention head $h$, we run the restrictive prompt while replacing the head's output at every position with its output on the permissive prompt (direction p2r\rev{; activation patching, \citealp{vig2020investigating,meng2022locating,wang2023interpretability,zhang2024patching}}), and record the change in the score $S$,
\begin{equation}
e_{ih}=S_i(\text{restrictive, output of } h \text{ copied from permissive}) - S_i(\text{restrictive}).
\label{eq:head-effect}
\end{equation}
Because the two versions of a new pair have the same number of tokens in every frame, positions align exactly.
Raw effects are not comparable across categories: \rev{on the original pairs, the effects of almost every head are larger on moral than on control pairs}, as shown below.
We therefore divide each pair's absolute effects by their sum over all heads, so that each pair's profile sums to one:
\begin{equation}
\mathrm{share}_{ih} = \frac{|e_{ih}|}{\sum_{h'} |e_{ih'}|}.
\label{eq:share}
\end{equation}
For a head set $H$, a model $m$ and a frame $f$, let $\bar s_c(m,f)$ be the mean, over the included pairs of category $c$, of $\sum_{h\in H}\mathrm{share}_{ih}$.
The selectivity of $H$ and its pooled value are
\begin{equation}
\rho(m,f)=\frac{\bar s_{\text{moral}}(m,f)}{\bar s_{\text{norm}}(m,f)},\qquad
\theta(f)=\frac{1}{|M_f|}\sum_{m\in M_f}\log\rho(m,f),
\label{eq:theta}
\end{equation}
where $M_f$ is the set of models with sufficient behavioural contrast in frame $f$: at least 30 included moral and 30 included norm pairs (contrasts with descriptive pairs need 20 included descriptive pairs, contrasts with one kind of norm 10 pairs of that kind).
$\rho$ is the selectivity of one model, $\exp\theta$ its pooled value over models; both equal 1 when the heads carry the same share for moral and for norm pairs.
The head set of the primary tests holds for each model the ten heads that the search described next identified as specific to morality in that model.
These are the located heads \rev{(\Cref{app:head-method} gives the names of the set in the released files)}.
The sets were fixed before any model was run on the new pairs.
Intervals are percentile bootstrap intervals with 10{,}000 resamples of items within category, and pooled estimates use the same resample for every model, so that their intervals reflect the sampling of items rather than the number of models \citep{efron1993bootstrap}.
Per-model tests permute category labels (20{,}000 permutations) with Benjamini-Hochberg correction across the 14 models within each frame and contrast \citep{benjamini1995controlling,phipson2010permutation}.

\begin{figure}[t]
\centering
\includegraphics[width=\linewidth]{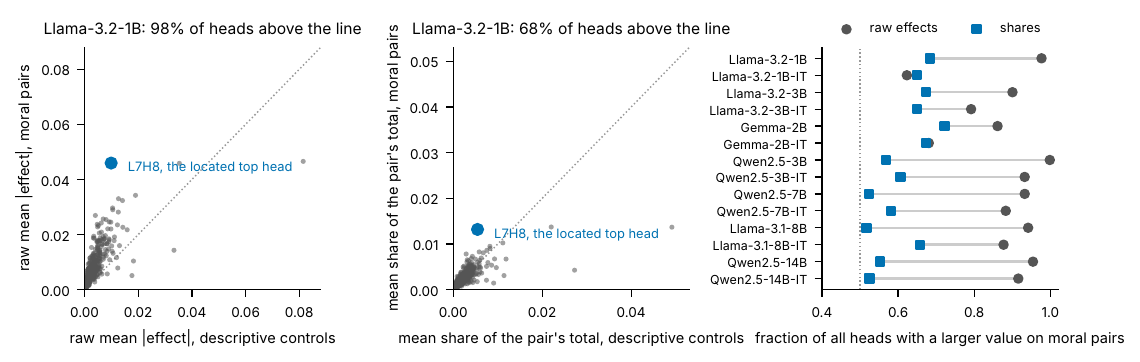}
\caption{\textbf{Why raw effects mislead.}
Left: in Llama-3.2-1B base, each dot is one of the 512 heads, placed by its mean absolute effect on the descriptive controls (horizontal) and on the moral pairs (vertical); almost every head lies above the diagonal\rev{: averaged over all heads, the effect on the moral pairs is 2.25 times that on the controls}.
Middle: the same heads with each pair's effects expressed as shares of the pair's total; the shared offset is gone and the top head (blue) stands out on its own.
Right: the fraction of heads with a larger value on moral pairs, for raw effects and for shares, in all 14 models.
The values are those of \Cref{tab:dev-heads}, computed from the released head effects on the original pairs.}
\label{fig:raw-vs-share}
\end{figure}

\paragraph{Every model has such heads, and shares show which ones.}
The natural first test of a head is whether its mean absolute effect is larger on moral than on control pairs.
Applied to the original pairs it \rev{finds a significant difference for about half of the heads or more in every base model (47 to 100\,\% of them; 442 of the 512 heads of Llama-3.2-1B, for example) and for 4 to 68\,\% of the heads of the \ac{IT} models}, because raw effects are larger on moral pairs for almost every head: for 86 to 100\,\% of the heads of each base model \rev{(62 to 93\,\% in the \ac{IT} models)}, with a ratio over all heads of 1.5 to 2.3 (16.1 in Qwen2.5-3B; \rev{0.9 to 1.7 in the \ac{IT} models;} \Cref{tab:dev-heads}).
The reason is not that most heads are moral.
\rev{The effects of all heads are larger on the moral pairs, and a raw comparison mostly measures this difference between the two sets of pairs, which almost every head shares (the \emph{offset}; \Cref{fig:raw-vs-share}).
The offset does not follow the size of the model's answer.
In Llama-3.2-1B the score differs on average by 0.39 between the two versions of a moral pair and by 0.25 between the two versions of a control pair, but also by 0.25 between the two versions of the normative pairs of the pilot described below, on which the effects of all heads together are as large as on the moral pairs (a mean sum of absolute effects of 3.4, against 3.5 on the moral pairs and 1.6 on the controls).}
Shares remove the offset, and something survives it.
After expressing each pair's effects as shares (\Cref{eq:share}), \rev{the head for which the first test finds the largest difference between moral and control pairs} (the \emph{top head} of each model) \rev{still has the largest difference of all heads, and a significant one} (permutation test, Benjamini-Hochberg over heads)\rev{,} in 11 of 14 models\rev{.
In these 11 models its mean share on moral pairs is 1.9 to 6.7 times its mean share on control pairs.
In the five base models among them the ratio is 2.4 to 3.8, while the same ratio on raw effects is 4.6 to 7.1} (87 in Qwen2.5-3B, where the control effects are near zero; \Cref{tab:dev-heads})\rev{; in the six \ac{IT} models it is 1.9 to 6.7, with raw ratios of 2.0 to 10.7}.
\rev{The top head loses the first place} in Gemma-2B base (L3H6: ratio 1.31, rank 7, $q=0.25$)\rev{, and its difference is not significant} in Qwen2.5-7B base and instruct ($q=0.10$ and $0.56$).
On the subset of pairs whose two versions have the same number of tokens, the top head's share ratio is significant in 12 of 14 models (all but the two Qwen2.5-7B models; Gemma-2B base reaches 1.80, $q=0.03$), and the top heads of Llama-3.2-1B and Llama-3.1-8B also stay first under the synonymous frame ``is ethically''.
\rev{The located heads of a model are the ten heads for which the first test finds the largest differences, the top head being the first of them.
Once the offset is removed, the ten located heads together carry 1.3 to 2.1 times the share on moral pairs that they carry on control pairs in 13 of the 14 models (0.7 times in Qwen2.5-3B base, where the effects on control pairs are very small), so they remain selective relative to the descriptive controls.}
But those controls differ from the moral pairs in frame, in answer words and in content, and the heads were chosen on the same moral pairs, so what the located heads respond to remained to be tested.

\begin{table}[t]
\centering
\caption{\textbf{The top heads on the original pairs, before and after share normalisation.}
Each row uses the model's own top head under the raw test (layer and head counted from zero).
``Heads M$>$C'': share of all heads with a larger raw mean effect on moral than on control pairs; ``All heads'': ratio of mean raw effects over all heads.
``Raw'': the top head's raw moral/control ratio.
``Share'': moral/control ratio of its share (\Cref{eq:share}), with 95\,\% bootstrap interval, Benjamini-Hochberg $q$ over heads and rank among all heads; the last two columns: the same on pairs whose two versions have the same number of tokens.
Selection inflates all values (the heads were chosen on these pairs).}
\label{tab:dev-heads}
\footnotesize
\setlength{\tabcolsep}{3pt}
\begin{tabular}{llrrrrrrrr}
\toprule
& & & & & \multicolumn{3}{c}{All pairs} & \multicolumn{2}{c}{Same token count} \\
\cmidrule(lr){6-8}\cmidrule(lr){9-10}
Model & Head & Heads M$>$C & All heads & Raw & Share [95\,\% \ac{CI}] & $q$ & Rank & Share & $q$ \\
\midrule
Llama-3.2-1B & L7H8 & 97.7\,\% & 2.25 & 4.64 & 2.41 [1.69, 3.65] & 0.003 & 1 & 2.63 & 0.006 \\
Llama-3.2-1B-\ac{IT} & L13H25 & 62.3\,\% & 0.92 & 6.63 & 6.71 [3.94, 12.29] & 0.003 & 1 & 5.86 & 0.005 \\
Llama-3.2-3B & L11H17 & 90.0\,\% & 1.60 & 5.12 & 3.43 [2.52, 4.73] & 0.004 & 1 & 4.32 & 0.013 \\
Llama-3.2-3B-\ac{IT} & L11H17 & 79.2\,\% & 1.17 & 3.44 & 3.57 [2.46, 5.37] & 0.003 & 1 & 3.83 & 0.003 \\
Gemma-2B & L3H6 & 86.1\,\% & 1.51 & 2.98 & 1.31 [0.81, 2.23] & 0.249 & 7 & 1.80 & 0.030 \\
Gemma-2B-\ac{IT} & L6H4 & 68.1\,\% & 0.95 & 1.97 & 1.88 [1.28, 2.90] & 0.024 & 1 & 2.09 & 0.023 \\
Qwen2.5-3B & L20H12 & 99.8\,\% & 16.10 & 87.42 & 3.83 [2.55, 5.90] & 0.001 & 1 & 3.26 & 0.013 \\
Qwen2.5-3B-\ac{IT} & L20H14 & 93.2\,\% & 1.72 & 10.70 & 4.61 [3.23, 6.93] & 0.002 & 1 & 3.99 & 0.012 \\
Qwen2.5-7B & L0H24 & 93.2\,\% & 1.72 & 5.11 & 2.63 [1.54, 4.64] & 0.099 & 1 & 4.01 & 0.270 \\
Qwen2.5-7B-\ac{IT} & L0H5 & 88.3\,\% & 1.47 & 2.55 & 1.47 [0.77, 2.94] & 0.561 & 1 & 2.09 & 0.474 \\
Llama-3.1-8B & L12H22 & 94.1\,\% & 1.87 & 5.50 & 3.11 [2.24, 4.45] & 0.003 & 1 & 3.20 & 0.007 \\
Llama-3.1-8B-\ac{IT} & L12H22 & 87.7\,\% & 1.48 & 4.80 & 3.54 [2.55, 5.09] & 0.005 & 1 & 3.40 & 0.005 \\
Qwen2.5-14B & L28H13 & 95.4\,\% & 1.88 & 7.10 & 3.41 [2.36, 5.06] & 0.003 & 1 & 3.72 & 0.009 \\
Qwen2.5-14B-\ac{IT} & L28H13 & 91.6\,\% & 1.71 & 5.18 & 3.56 [2.48, 5.32] & 0.003 & 1 & 3.99 & 0.006 \\
\bottomrule
\end{tabular}

\end{table}

\paragraph{A pilot with normative pairs in the same frame.}
Before the new pairs were written, a pilot on Llama-3.2-1B base compared the original moral pairs with 81 norm pairs in the same ``is morally'' frame.
The offset disappears (raw ratio over all heads 1.03), which shows that it \rev{is not a matter of moral content.
It comes with what sets the original controls apart from the moral and from the norm pairs (against the controls, the norm pairs show an offset of their own, 2.17): their frame, their answer words or their descriptive content, which the pilot cannot tell apart}.
The top head L7H8 keeps a preference for moral over norm content (share ratio 2.43, 95\,\% \ac{CI} 1.82 to 3.28; 2.16 on pairs of equal length, $q=0.002$), and its share on norm pairs equals its share on neutral pairs (ratio 0.99).
So does L8H24 (2.47), the head with the largest raw effects, which the raw test classes as domain-general because it also responds to the neutral controls.
The pilot also shows the limit of such ratios.
Under share normalisation, 210 of 512 heads carry a larger share on moral than on norm pairs at $q<0.05$ (shares sum to one, so others carry less), the mean profiles of the two categories correlate at only $r=0.76$, and their ten heads with the largest mean share have only two heads in common.
A ratio above 1 for heads selected on moral pairs can therefore reflect that the two profiles differ, not that the selected heads are special; for this reason the full design also runs the mirror image of such a selection\rev{: the same selection made on norm pairs, whose heads are then tested for a larger share on norm than on moral pairs} (\Cref{sec:heads-discovery}).
The norm pairs of the pilot also have the larger edits of the expanded set, and L7H8 was selected on the same moral pairs.

\paragraph{Base and \ac{IT} models.}
Measured on shares, which do not depend on the size of the effects, the pattern across heads changes little with instruction tuning.
The Pearson correlation between the base and the instruct profiles of mean shares over shared moral pairs is 0.87 to 0.97 and rises mildly with scale within each family (Llama 0.913, 0.967, 0.969; Qwen 0.868, 0.940, 0.969; Gemma 0.899; \Cref{tab:dev-base-it}).
The base model's top head stays rank 1 in Llama-3.2-3B, Llama-3.1-8B and Qwen2.5-14B and is demoted in Llama-3.2-1B (rank 3), Gemma-2B (3), Qwen2.5-3B (2) and Qwen2.5-7B (6).
\rev{What instruction tuning changes in the small models is the size of the raw effects on moral pairs, summed over all heads} ($-48$\,\% in Llama-3.2-1B, $-46$\,\% in Qwen2.5-3B, $-33$\,\% in Llama-3.2-3B\rev{, $-13$\,\% in Gemma-2B; in the three larger models it changes less and not always in the same direction, $-14$\,\% in Llama-3.1-8B, $+13$\,\% in Qwen2.5-7B and $+20$\,\% in Qwen2.5-14B}), \rev{and not how the effect is divided among the heads}.

\begin{table}[t]
\centering
\caption{\textbf{Base versus \ac{IT} models on shared original moral pairs.}
Profile correlation: Pearson and Spearman correlation between the base and \ac{IT} profiles of mean shares over all heads.
Share ratio: the base top head's moral/control share ratio in each model.
Rank in \ac{IT}: rank of the base top head in the \ac{IT} model.
Raw change: change of the summed raw moral effect from base to \ac{IT}.}
\label{tab:dev-base-it}
\small
\setlength{\tabcolsep}{3pt}
\begin{tabular}{llrrrcrr}
\toprule
\vio{Base model} & Base top head & Shared pairs & Pearson $r$ & Spearman & Share ratio base / \ac{IT} & Rank in \ac{IT} & Raw change \\
\midrule
Llama-3.2-1B & L7H8 & 57 & 0.913 & 0.954 & 2.38 / 1.78 & 3 & $-48$\,\% \\
Gemma-2B & L3H6 & 50 & 0.899 & 0.913 & 1.29 / 1.66 & 3 & $-13$\,\% \\
Qwen2.5-3B & L20H12 & 54 & 0.868 & 0.747 & 3.78 / 2.32 & 2 & $-46$\,\% \\
Llama-3.2-3B & L11H17 & 68 & 0.967 & 0.966 & 3.46 / 3.41 & 1 & $-33$\,\% \\
Qwen2.5-7B & L0H24 & 66 & 0.940 & 0.943 & 2.25 / 1.25 & 6 & $+13$\,\% \\
Llama-3.1-8B & L12H22 & 70 & 0.969 & 0.965 & 3.03 / 3.30 & 1 & $-14$\,\% \\
Qwen2.5-14B & L28H13 & 71 & 0.969 & 0.964 & 3.28 / 3.95 & 1 & $+20$\,\% \\
\bottomrule
\end{tabular}

\end{table}

\paragraph{A linear direction at the head's layer.}
The original pairs also give a direction for moral stance at the top head's layer.
Its effects exceed those of random directions, but they are modest.
Steering along it in Llama-3.2-1B moves the score over a range of 0.48 for $\alpha\in[-3,3]$, about four times the mean absolute range of 20 random unit directions (0.12; the largest is 0.35, and 3 of 20 reach half the moral effect).
The residual norm at that layer and position is only 5.8, so $\alpha=3$ is a perturbation of about half its size; step sizes should be scaled to the residual norm.
Projected zero-shot onto ETHICS commonsense scenarios \citep{hendrycks2021aligning}, the direction separates acceptable from unacceptable scenarios with \ac{ROC}-\ac{AUC} 0.67 (Llama-3.2-1B) to 0.71 (Llama-3.1-8B), although in Llama-3.2-1B the model's own score, without steering, barely separates them (mean $S$ $-0.190$ against $-0.198$) and steering moves both classes alike.
These are properties of a direction found on the original pairs; the new pairs do not test them, but \Cref{sec:steering} returns to directions with a design of its own.

\subsection{Features that separate moral from other prompts}
\label{sec:features-dev}
\paragraph{Measurement and capture.}
A \ac{SAE} \rev{approximates} the \rev{hidden state of} one layer as a \rev{weighted} sum of a few ``features'' out of a large dictionary, each meant to stand for one recurring pattern.
Heads are components of the computation; features are coordinates of a description of the \rev{hidden} state, and they are only as good as that description.
So the analysis of features asks two questions in order: how much of the difference between the two versions of a pair the dictionary can see at all, and, within what it sees, which features change between the versions and whether those features follow the content or the cue.

For each \rev{dictionary} (\Cref{tab:sae-conditions}) we encode the \rev{hidden} state at the final position of each prompt and compute, for feature $j$ and pair $i$, the size of its change between the permissive and the restrictive version,
\begin{equation}
a_{ij}=\left|z_j(x_i^{+})-z_j(x_i^{-})\right|.
\label{eq:feature-change}
\end{equation}
All feature analyses use the final token, and A1 is the primary frame.
Before interpreting any result about features we ask how much of the pair difference the dictionary sees at all.
With $d$ the difference between the \rev{hidden} states at the final position of the permissive and the restrictive version and $\hat d$ the same difference between the two \rev{reconstructions by the dictionary}, the capture is
\begin{equation}
R^2 = 1-\frac{\lVert d-\hat d\rVert^2}{\lVert d\rVert^2},
\label{eq:capture}
\end{equation}
reported with the cosine between $d$ and $\hat d$, per category and frame.
A \rev{dictionary} whose mean $R^2$ is below 0.5 in a frame is flagged for that frame: a null result about features there says little about the model.
The capture check is to the feature study what a resolution limit is to a microscope: it says what the dictionary can and cannot show.

\begin{table}[t]
\centering
\caption{\textbf{The 14 \rev{dictionaries}}, with the number of features whose change is larger on moral than on norm pairs in the same ``is morally'' frame on the original expanded sets ($n_M$ moral and $n_N$ norm pairs after behavioural filtering; one-sided permutation test, Benjamini-Hochberg at 0.05 within each \rev{dictionary}).
R$\ell$: \rev{hidden state (the residual stream)} after block $\ell$; \ac{MLP}0: output of the first \ac{MLP} block; $k$: TopK sparsity.
These counts show that features separating moral from norm pairs in the same frame exist; the larger edits of the original norm pairs make them hard to compare across \rev{dictionaries} or to read as a measure of how much processing is moral.}
\label{tab:sae-conditions}
\small
\begin{tabular}{llrrrr}
\toprule
Model & Site / $k$ & Width & $n_M$ & $n_N$ & Moral $>$ norm \\
\midrule
Llama-3.2-1B & R7 / 64 & 16{,}384 & 62 & 172 & 32 \\
Llama-3.2-1B & R7 / 128 & 16{,}384 & 62 & 172 & 66 \\
Llama-3.2-1B-\ac{IT} & R7 / 64 & 16{,}384 & 54 & 175 & 20 \\
Llama-3.2-1B-\ac{IT} & R7 / 128 & 16{,}384 & 54 & 175 & 52 \\
Llama-3.2-1B & \ac{MLP}0 / 64 & 16{,}384 & 62 & 172 & 16 \\
Llama-3.2-3B & R11 / 64 & 24{,}576 & 68 & 190 & 56 \\
Llama-3.2-3B & R11 / 128 & 24{,}576 & 68 & 190 & 124 \\
Llama-3.2-3B-\ac{IT} & R11 / 128 & 24{,}576 & 65 & 190 & 97 \\
Llama-3.2-3B-\ac{IT} & R3 / 128 & 24{,}576 & 65 & 190 & 30 \\
Llama-3.2-3B-\ac{IT} & R7 / 128 & 24{,}576 & 65 & 190 & 62 \\
Llama-3.2-3B-\ac{IT} & R19 / 128 & 24{,}576 & 65 & 190 & 49 \\
Llama-3.1-8B & R12 / JumpReLU & 32{,}768 & 71 & 192 & 18 \\
Qwen2.5-7B & R0 / 128 & 28{,}672 & 68 & 210 & 22 \\
Qwen2.5-7B-\ac{IT} & R0 / 128 & 28{,}672 & 70 & 204 & 21 \\
\bottomrule
\end{tabular}

\end{table}

\paragraph{Every dictionary holds such features, and most of them follow the frame.}
On the original pairs \rev{(73 moral, 221 norm and 105 neutral pairs, of which a model keeps 54 to 71, 172 to 210 and 64 to 83)} every one of the 14 dictionaries holds features that separate moral from neutral prompts \rev{(37 to 222 per dictionary; the located features)}, and, with the expanded norm set, \rev{between 16 and 124 features per dictionary} whose change is larger on moral than on norm pairs in the same frame, most at layer 11 of Llama-3.2-3B (\Cref{tab:sae-conditions}).
Because the original norm pairs have larger edits than the moral pairs, which tends to enlarge their feature changes, these counts are conservative as evidence that such features exist, but they cannot measure how much processing is moral; \Cref{sec:features-discovery} asks that question on pairs of the same form.
The two kinds of count differ a great deal: 3.9 to 18.4 times as many features separate moral from neutral prompts as separate moral from norm prompts at a fixed threshold on the effect size, which would suggest that \rev{\emph{the features mostly encode normativity}}.
A look at what differed between the prompts shows that they encode the frame instead: moral and norm pairs shared the frame ``is morally'' and its answer words, while \rev{most of} the neutral pairs ended in ``is typically'' \rev{(81 of 105; the others in a similar phrase)}.
Of the features that separated moral from neutral pairs, 63 to 95\,\% never fire on any neutral prompt, while their mean activation on norm prompts is close to that on moral prompts.
A test that swapped the frames in Llama-3.2-1B (with the \rev{dictionary} at layer 7) shows why: neutral content followed by ``is morally'' activates these features almost as much as norm content does (mean activation 0.096 against 0.108; moral content 0.150\rev{; neutral content in its own frame 0.001}), and moral content followed by ``is typically'' silences them (0.007).
Among features active on at least 10\,\% of both moral and neutral prompts, the asymmetry disappears (0 to 7 features with large effects against neutral pairs, 0 to 6 against norm pairs).
The ratio of the two counts measures the frame, not the content, and we do not use it; whether the moral-versus-neutral features follow the frame on new pairs of every category is the hypothesis H5 of \Cref{sec:features-heldout}.

\paragraph{What the dictionaries can and cannot see.}
At the final token the dictionaries capture little of the difference between the two versions of a pair: across the 14 \rev{dictionaries}, on the original pairs, $R^2$ ranges from $-0.27$ to $0.38$ (median 0.04; cosine 0.26 to 0.66).
In the \rev{dictionaries} examined in detail, the pair difference is 4 to 11\,\% of the squared activation norm, less than the reconstruction error of 16 to 29\,\%.
The dictionaries describe each prompt fairly well, then, but see little of what changes between the two versions of a pair.
Three rules follow, and the rest of the paper keeps them: null results about features are weak evidence, interventions on features must keep the reconstruction error in place (\Cref{app:feature-method}), and the main causal claims rest on head patching.
\Cref{app:audit} gives the checks on the original pairs behind these rules and behind the form of the new pairs.

\section{Checks on the original pairs, and what the design takes from them}
\label{app:audit}

Before the new pairs were written, the analyses on the original pairs were checked against their own data: the stored head effects and feature activations, a small replication on Llama-3.2-1B base (\ac{CPU}), and the reconstruction diagnostics run on all 14 \rev{dictionaries}.
The replication was validated against the stored results: behavioural validity counts agree exactly (62 moral, 28 control and 172 norm pairs valid on Llama-3.2-1B), per-layer \ac{MLP} effects agree to three decimals, and per-pair head effects correlate at 0.987 to 0.998 with the stored arrays.
The scripts (numbered below) and their outputs are released.
This appendix records what each natural statistic of such a search gives, what the check shows, and what the design of the new pairs takes from it.
The pattern is the same throughout: a statistic computed on prompts of one kind, against controls of another form, measures the form as much as the content.

\paragraph{Heads: raw effects measure a shared offset.}
The count of heads with a significantly larger raw effect on moral than on control pairs mostly measures the offset of \Cref{sec:heads-dev}: raw effects are larger on moral pairs for almost every head (Llama-3.2-1B: 98\,\% of heads, ratio over all heads 2.25).
With effects expressed as shares, 20 to 107 heads per model remain significant; the top head stays rank 1 and significant in 11 of 14 models, at 2.4 to 3.8 in base models, and fails in Gemma-2B base and both Qwen2.5-7B models; this holds on pairs of equal length (script 02).
A variant that compares raw effects on pairs of equal length with bootstrap intervals (moral minus control, ratios 2.81 to 9.67) is passed by 23 to 99.8\,\% of all heads; relative to the background the selected heads of the base models are at 1.9 to 4.3, about half of their raw ratios (09).
Every head comparison in this paper therefore uses shares (\Cref{eq:share}).
The same offset explains an apparent gradient with model size in the count of significant heads under instruction tuning (a drop of 32 to 95\,\% at 1B to 3B and none at 7B to 14B): the counts follow the change in the offset (Llama-3.2-1B: 2.25 to 0.92), driven by raw moral effects shrinking in small \ac{IT} models, while on profiles of shares the pattern is preserved everywhere ($r=0.87$ to $0.97$) and normalised counts rise under instruction tuning in 5 of 7 pairs (03).
H4 therefore compares profiles of shares.

\paragraph{Heads: token alignment.}
Of the 73 original moral pairs, 26 or 27 (depending on the tokenizer) are tokenised to different lengths in their two versions.
Patching aligned the two versions from the left, up to the shorter length, so positions after the edit were misaligned, and when the restrictive prompt was longer the final position was never patched.
An apparent split of upstream profiles from path patching into ``flat'' and ``layered'' ones (77 to 84\,\% flat, unchanged by instruction tuning) was a consequence: in 665 of 665 (model, pair) cases the profile is not flat if and only if the restrictive prompt has more tokens, for equal lengths every source layer gives the identical result, and base and instruct models share a tokenizer (01).
The new pairs have equal token counts in every frame.

\paragraph{Heads: controls for every intervention.}
The first \ac{MLP} block, which patching singles out as the largest single component, is equally dominant on non-moral pairs: mean absolute effect 0.202 (2.7 times the next \ac{MLP} layer) on neutral controls and 0.300 (1.9 times) on norm pairs, against 0.316 (1.8 times) on moral pairs (06); it carries the identity of the edited word, not a moral prior.
An overlap of top heads across the four construction tiers of the original pairs ($p<10^{-5}$ against a null of random head sets) is equalled by random four-way splits of the same moral pairs, and random splits of the control pairs share at least one top-5 head in 99 to 100\,\% of draws for Llama and Gemma (04); the null was uninformative.
On the original pairs, control pairs went through per-head patching only; \ac{MLP} patching, activation patching and the logit lens had no non-moral baseline.
Every analysis on the new pairs runs every category through every intervention.

\paragraph{Heads: what is kept.}
The same top head is recovered under ``is ethically'' in Llama-3.2-1B and Llama-3.1-8B (\Cref{sec:heads-dev}).
The direction for moral stance replicates in steering (range 0.48, about four times a typical random unit direction and above all 20 tested, although 3 of 20 random directions reach half the moral effect; $\alpha=3$ is a 52\,\% perturbation of the residual norm) and separates ETHICS scenarios at \ac{AUC} 0.67 to 0.71 (08c).
The late-layer dip of the emerging judgment is stable in Llama-3.2-1B (0.43, median 0.31), absent on the descriptive controls (0.10, median 0.00) and present on norm pairs in the same frame (0.22, median 0.18), at a comparable size relative to the gap; it is examined again on the new pairs in \Cref{app:dip} (08b).

\paragraph{Features: the frame.}
Of the features that separated moral from neutral pairs, 63 to 95\,\% never fire on any neutral prompt; neutral content with ``is morally'' activates them (0.096 against 0.108 for norm prompts) and moral content with ``is typically'' silences them (0.007); among features active in both frames the asymmetry disappears (05, 05b, 08a).
The ratio of moral-versus-neutral to moral-versus-norm features (3.9 to 18.4) is not used; every category of the new pairs is shown in the same frames, and H5 tests the frame dependence directly.
A first count of moral-versus-norm features, on the original 32 norm pairs, found none in 12 of 14 \rev{dictionaries}; with the expanded norm set every \rev{dictionary} has 16 to 124 such features, kept as an existence result (\Cref{tab:sae-conditions}).
A hypothesis that at most 64 features recover 90\,\% of the dense moral direction failed and is reported as a negative result.
Top-$K$ specificity ratios of 17.85 and 19.56 are inflated by selection (a version of the same analysis that accounts for the selection gives 1.92 and 1.49), and a share of features that survive instruction tuning (60 to 80\,\%) inherits the frame confound and has no noise ceiling; neither is used.

\paragraph{Features: reconstruction.}
A reconstruction gate computed over the whole sequence was passed by 2 of 14 \rev{dictionaries}, with a score of variance explained of $-1625.5$ for the Llama Scope dictionary of Llama-3.1-8B.
The gate averaged over all positions, including the beginning-of-sequence token, whose residual norm is 425 to 499 against 5 to 9 elsewhere.
Without that token the 8B score is 0.715 (the stored features are reproduced exactly, $r=1.00000$, so the loader was correct), the score of the Llama-3.2-1B dictionary drops from 0.9992 to 0.831, and across the 14 \rev{dictionaries} the score is 0.71 to 0.89, with 5 reaching 0.85.
The gate is replaced by the capture check at the final token (\Cref{sec:features-dev}), which is low in all 14 \rev{dictionaries} ($R^2$ $-0.27$ to $0.38$, median 0.04).

\paragraph{Features: the ablation.}
A first ablation design replaced the \rev{hidden} state with the \rev{dictionary's reconstruction of it} after zeroing the selected features, which mixes the removal with the reconstruction error.
In Llama-3.2-3B-\ac{IT}, zeroing 97 random features already changed the moral stance by $-0.092$; the 97 target features gave $-0.099$, a specific effect of 0.007 (larger in Llama-3.2-1B base: $-0.201$ against $-0.150$), and the bootstrap treated repeated random draws on the same pair as independent.
Every intervention on features in this paper keeps the reconstruction error in place (\Cref{app:feature-method}) and compares the set with random sets of similar activity.

\paragraph{Pairs.}
The labels of the expanded original sets \rev{(the 189 added norm pairs and the 73 added descriptive pairs)} were assigned with three language models (Claude Haiku, Sonnet and Opus) plus author review of flagged items and of a sample of 30 pairs; the new pairs were validated by annotators blind to the labels.
The expanded norm pairs mostly contrast whole sentences (\Cref{tab:stimulus-structure}): 81\,\% share no prefix, their mean edit is 7.7 tokens against 4.2 for the moral pairs, and 15\,\% have the same token count in both versions.
The new pairs differ in one to three tokens, with edits of the same size and position across categories.

\begin{table}[t]
\centering
\caption{Structure of the original stimulus sets (tokenizer counts only).}
\label{tab:stimulus-structure}
\small
\begin{tabular}{lrrr}
\toprule
Set & Mean edit (tokens) & Pairs with no shared prefix & Same token count \\
\midrule
Moral (73) & 4.2 & 1\,\% & 64\,\% \\
Norm, original 32 & 5.5 & 31\,\% & - \\
Norm, 189 added & 7.7 & 81\,\% & 15\,\% of all 221 \\
Neutral (105) & 3.4 & - & 38\,\% \\
\bottomrule
\end{tabular}
\end{table}

\section{A late-layer dip of the emerging judgment}
\label{app:dip}
\rev{This observation} is reported for the record and used in no claim.

On the original pairs we observed that, read through the logit lens \citep{nostalgebraist2020logitlens}, the emerging verdict on permissive prompts temporarily reverses in late layers before recovering at the output.
In Llama-3.2-1B the dip is stable across runs (mean 0.43, standard deviation 0.33, median 0.31 on moral pairs).
It is absent on the descriptive controls (mean 0.10, median 0.00) but present on norm pairs in the same frame (mean 0.22, median 0.18), \grn{and of a similar size relative to the behavioural gap: a mean dip of 0.43 against a mean gap of 0.39 on moral pairs, and of 0.22 against 0.25 on norm pairs}.
This points to the ``is morally'' construction rather than to moral content, but the controls also differ in content.
On the new pairs we compute the dip in frames A1 and B, with one forward pass per item.
If the dip belongs to the construction, it should be absent in frame B; if it belongs to moral content, it should be present in frame B for moral items.
It is not used in any claim of this paper.
With the cue, the dip is of similar size relative to the behavioural gap for moral and normative pairs (medians over models 0.25 and 0.22 of the gap) and small for descriptive pairs (0.07): as on the original pairs, with ``is morally'' in the prompt it belongs to the construction.
Without the cue the dip shrinks in the small Llama models (to 0.00, 0.07 and 0.16 of the gap in Llama-3.2-1B, 1B-\ac{IT} and 3B, against 0.37 to 0.64 with the cue) but stays for moral pairs in the larger models, at two to four times its size on normative pairs: 0.38 against 0.20 of the gap in Llama-3.1-8B (ratio 1.9, 95\,\% interval 1.1 to 3.6), 2.9 in its instruct version, 3.6 and 4.3 in the two Qwen2.5-7B models, 2.7 in Llama-3.2-3B, 2.1 and 2.6 in the two 14B models with intervals that include 1.
The dip therefore has a part tied to the construction and a part that leans towards moral pairs and survives the removal of the word ``morally''.
A lexical explanation competes for that second part: the permissive versions of moral pairs name harms (stealing medicine for a dying child), and their words, run alone in the frame without the cue, are more negative for the model than those of normative pairs.
Adjusting the difference between moral and norm pairs in the dip for the most negative word of each stem removes about half of it, and in most models what remains cannot be told from zero (without the cue: lexical in 4 models, undecided in 7, a clear moral remainder in 1, Qwen2.5-7B); but the minimum of the dip comes late, after the edited word has reached the final position (median depth 0.8 to 0.95 against 0.4 to 0.7), and within a category the size of the dip barely tracks the negativity of the words (rank correlations mostly between $-0.3$ and $0.1$), which a purely lexical account would not predict.
Read through the Jacobian lens of \Cref{sec:jlens}, which carries each layer's \rev{hidden} state to the output through the average Jacobian of the remaining layers instead of reading it with the output embedding, the dip remains (the eight models up to 3B parameters, the same measure).
With the cue it is as large as or larger than through the logit lens in every one of these models except Gemma-2B-\ac{IT} (Gemma-2B shows none through either lens), and as large for descriptive as for moral pairs.
Without the cue it is smaller and again larger for moral than for normative pairs, with the interval of the ratio above 1 in six of the seven models that show a dip, now including Llama-3.2-1B, where the logit lens saw none.
The lens also shows what most of the dip is: with the cue, the difference between the two versions of a pair dips by at most a third as much as the score of the permissive version does (for moral and normative pairs the logit lens shows the same), so both versions swing towards the negative answer words in late layers and recover together.
So the dip appears through both lenses, but most of it is a late shift of the whole score rather than a reversal of the judgment between the two versions.
We leave it as an open observation: a late shift tied to the evaluative construction, somewhat larger for moral content, whose cause these pairs cannot settle.

\section{Materials and measurement in detail}
\label{app:methods}
The models, the new pairs, the frames, the score and the plan, in the words of the long version; the two measurements are defined in \Cref{app:study1}, the human validation follows in \Cref{sec:validation} and the head and feature methods in \Cref{app:head-method,app:feature-method}.

\subsection{Models}
\label{sec:models}
We study 14 models: Llama-3.2-1B, Llama-3.2-3B, Llama-3.1-8B, Gemma-2B and Qwen2.5-3B, 7B and 14B, each in base and \ac{IT} form.
Instruction-tuned models receive raw text without a chat template throughout.
All runs use TransformerLens \citep{nanda2022transformerlens} in float32.
Every prompt starts with exactly one beginning-of-sequence token; the Qwen models, which have none, receive their end-of-sequence token instead, as in the head analysis on the original pairs.\footnote{The feature analysis on the original pairs tokenised Qwen prompts without a leading token (the default of TransformerLens 3.1.0 for Qwen2 models).
Its stored Qwen features are reproduced exactly without the leading token, and at $r=0.9925$ and $0.9951$ with it.
The Qwen feature sets were therefore selected under the other convention.
Llama and Gemma are not affected.}
The \rev{feature} analyses use the 14 \rev{dictionaries} on which the features were located (\Cref{tab:sae-conditions}): TopK \rev{dictionaries (a TopK dictionary keeps, for each vector that it encodes, the $k$ features with the largest values and sets the others to zero)} trained for this project on the \rev{hidden states} of Llama-3.2-1B and 3B (base and \ac{IT}, several layers and sparsities) and on the output of the first \ac{MLP} block of Llama-3.2-1B, the Llama Scope dictionary at layer 12 of Llama-3.1-8B \citep{he2024llamascope} \rev{(a public dictionary, the only one of the 14 that we did not train; its authors trained it as a TopK dictionary and released it in a converted form, in which a feature is active when its value exceeds a threshold)}, and TopK \rev{dictionaries} at layer 0 of Qwen2.5-7B, base and \ac{IT}.
\rev{The dictionaries are not one per model, and that there are 14 of them, like the models, is a coincidence: they belong to seven of the models (five to Llama-3.2-1B and its \ac{IT} version, six to Llama-3.2-3B and its \ac{IT} version, one to Llama-3.1-8B and one to each of the two Qwen2.5-7B models).
Each dictionary is used only on the model on whose hidden states it was trained.
Training a dictionary takes considerable compute, so only some models have one: we trained dictionaries for six models, and for the base Llama-3.1-8B we used the public Llama Scope dictionary.}

\subsection{The new pairs}
\label{sec:items}
Each new item is a minimal pair of short phrases (a \emph{stem}): a better (permissive) and a worse (restrictive) version that differ in one to three tokens.
There are three content categories.
In a \emph{moral} pair the two versions differ in harm, fairness, honesty, consent, property or loyalty.
In a \emph{norm} pair they differ in a non-moral standard: prudential (the agent's own interests), etiquette (manners and convention), aesthetic (\rev{how well something is designed, made or performed}) or epistemic (the quality of reasoning or evidence).
In a \emph{descriptive} pair they differ in how good a thing or a situation is, and nobody's conduct is judged.

All pairs satisfy mechanical constraints that were checked with tokenizers only.
\rev{In each of the three frames and for each of the 14 models, the better and the worse version of a pair have the same number of tokens (the models of a family share one tokenizer, so three tokenizers were checked: Llama-3, Qwen2.5 and Gemma).
The two versions can therefore be compared position by position when a head's output is copied from the run on one version into the run on the other.
No stem contains a word that the score counts as an answer (\Cref{sec:frames}), and no stem is a near duplicate of another item or of any original pair (a Jaccard similarity of the content words of at least 0.5).
Moral and norm pairs use the same four sentence templates in similar proportions (what changes is the first verb, a function word, an object or modifier, or the purpose or reason at the end): on the retained pairs each template makes up 19 to 31\,\% of either group.}
\rev{From the \NcandTotal{} candidates that pass, a rating set of \NratedRoundOne{} pairs was drawn by a rule that uses no model output and leaves no choice to us: all \NratedMoral{} moral candidates; \NratedNormKind{} norm pairs of each kind (\NratedNorm{} in all), drawn at random within fixed numbers per sentence template; and \NratedDescr{} descriptive pairs drawn at random.
The random draw of the norm pairs was repeated 5,000 times, and we kept the draw whose norm pairs were closest to the moral pairs in mean token count, edit size and edit position.
The random number generator started from a fixed seed, so that the script gives the same set whenever it is run.
The remaining \NreserveTotal{} candidates formed a reserve for a second rating round, to be used only if a category fell short of its minimum; \NratedRoundTwo{} of them were rated in that round (\Cref{app:validation}).}
The pairs were written before any pretrained model was run on them.

\subsection{Frames and score}
\label{sec:frames}
Every category is shown in the same three frames (\Cref{tab:frames}).
In \textbf{A1} (primary, cue present), two fixed priming sentences precede ``\stem{} is morally''.
In \textbf{B} (primary, cue absent), the same priming sentences precede ``\stem{} is''.
In \textbf{A0} (secondary), ``\stem{} is morally'' stands alone, as on the original pairs.
So A1 and B differ only in the cue word, and A1 and A0 differ only in the priming.
\rev{A1 and B are the \emph{primary} frames: the two main tests of the paper (H1 and H2, \Cref{sec:heads-heldout}) compare moral with normative pairs in them.
A0, the frame of the original pairs, is \emph{secondary}: it is analysed in the same way, as an additional check.
In the same sense, the analysis that the plan names for H1 and H2 is the primary analysis, and every variation of it is secondary.}
The priming sentences are there because a bare ``\stem{} is'' would rarely be completed with an evaluative word: on the original pairs with Llama-3.2-1B, a frame without ``morally'' put 2 to 16\,\% of the next-token probability on evaluative words without priming and 31 to 63\,\% with it.
Frames and answer words were chosen on the original pairs.
Throughout the paper, ``with the cue'' means frame A1 and ``without the cue'' means frame B.

\begin{table}[t]
\centering
\caption{The three frames, identical for every category.
The stem is a new pair; the model's next-token probabilities at the final position are recorded.
\rev{Underlined: the words that follow the stem in each frame.}}
\label{tab:frames}
\small
\begin{tabularx}{\linewidth}{@{}llY@{}}
\toprule
Frame & Role & Prompt \\
\midrule
A1 & primary, cue present & A sunny day at the beach is good.
A flat tire on the highway is bad.
\ExStem{} \exdiff{is morally} \\
B & primary, no cue & A sunny day at the beach is good.
A flat tire on the highway is bad.
\ExStem{} \exdiff{is} \\
A0 & secondary, the original frame\rev{, as in \citealp{liga2025neurons}} & \ExStem{} \exdiff{is morally} \\
\bottomrule
\end{tabularx}
\end{table}

The model's answer is read from its next-token probabilities at the final position: the score $S = P(\text{positive words}) - P(\text{negative words})$ sums the probability of 14 positive answer words (acceptable, justified, right, good, ...) and subtracts that of 14 negative ones (unacceptable, wrong, bad, ...), all single tokens in the three tokenizers (\Cref{app:head-method}).
Secondary scores are its log-odds version (\ac{LD}\vio{, the logarithm of the summed probability of the positive words minus the logarithm of that of the negative words}) and the score used when the components were located\vio{; a choice between two options labelled A and B is read from the logit of ``A'' minus the logit of ``B'' (\Cref{sec:further}).
Unless a passage says otherwise, every number is computed on $S$}.
The \emph{behavioural gap} of a pair is $S(\text{permissive}) - S(\text{restrictive})$: how much more the model favours the positive words for the better version than for the worse one.
The gap is the model's judgment of the pair in one number, and every effect of a component in this paper is measured as a change of this gap.
A pair enters a model's head analysis in a frame only if this gap is at least 0.02, so that there is a preference for the heads to carry.
A rehearsal on 36 original pairs found the frames viable in every model that could be loaded (\Cref{app:software}).

\subsection{Analysis plan}
\label{sec:prereg}
Before any model was run on the new pairs, we wrote down the hypotheses, the frames and answer words, the head and feature sets to be tested, the acceptance rule for the rated pairs, the analyses and the interpretation bands.
We then ran every model on all candidate pairs, retained or not (the two 14B models in two passes, the \NratedRoundOne{} pairs of the first rating round and then the \NreserveTotal{} spare pairs), so that the model runs did not have to wait for the ratings and could not depend on them.
Which pairs enter the analysis was decided by the annotators alone, who never saw any model output, and no result was examined before the list of retained pairs was fixed.
Further analyses were added \grn{after the plan was written} and before any result was seen: the analyses of the circuit in \Cref{sec:circuits} and the two of \Cref{sec:exploratory}; all are reported as exploratory, and every addition to the plan is listed in \Cref{app:prereg}.
\rev{Analyses that were designed after the main results were known are marked as exploratory where they are reported (among them the further tests of the located heads in \Cref{sec:further}).}


\subsection{Stimuli and human validation}
\label{sec:validation}

Each candidate item is a pair of short phrases that differ in one edit.
We built every pair with an intended better option, a broad category (moral, non-moral normative, or descriptive) and, for normative pairs, an intended subtype (prudential, etiquette or social convention, aesthetic, or epistemic).
These assignments were treated as hypotheses to be checked by people, not as gold labels.
\NraterCountCap{} annotators (\rev{all fluent in English}) rated every pair.
They saw the five standards with a one-line definition each, but were not told which category or subtype any pair had been built for, that the pairs formed groups, what the study expected to find, or any model result; they were asked to work alone, not to discuss the pairs with anyone and not to use an \ac{AI} assistant.
For each pair they answered which option is better (or that there is no clear difference); how much worse the worse option is (1 a little, 2 clearly, 3 very much); whether anyone's action, choice or work is being judged at all (no, when the phrase only describes a thing, condition or situation); and, if so, how much the difference between the two options is a matter of each of the five standards, rated separately on a scale from 0 (not at all) to 3 (very much): moral (effects on other people), prudential (the doer's own interests), etiquette or social convention, aesthetic (\rev{how well something is designed, made or performed}) and epistemic (the quality of reasoning or evidence).
Several standards could be rated high for the same pair; no choice between them was required.
They could also flag a pair as unclear.

\rev{The retention rule, fixed before any annotation was collected, was the following.}
From each annotator's ratings we read off the class of standard they saw in a pair: the standard with the highest rating when one stands out; none when they answered that nobody's action, choice or work was judged (or rated every standard 0); and no class at all when the moral standard was tied with a non-moral one.
A pair was retained only if at least two thirds of the annotators (\TwoThirds) chose the intended better option and at least two thirds gave the intended class: moral for moral pairs, any of the four non-moral standards for normative pairs, and none for descriptive pairs.
Retention further required a median moral rating of at least 2 for moral pairs and at most 1 for normative and descriptive pairs,\footnote{With \NraterCount{} annotators the median is the mean of the two middle ratings, so a split such as 1, 1, 2, 2 (median 1.5) fails both thresholds.}
an answer to the deciding questions from every annotator, and fewer than half of the annotators (\HalfRule) flagging the pair as unclear.
All retained pairs entered the primary analyses; the ratings were used for no further selection.
(The rule that includes a pair in a model's analysis, which depends only on that model's own answer probabilities, is described in \Cref{sec:frames}.)
Analyses of a single normative subtype used only the retained normative pairs whose intended subtype was rated highest, and at least 2, by at least \TwoThirdsN{} annotators.
As sensitivity analyses we also used the stricter subset on which all \NraterCount{} annotators agreed on the better option and on the class of standard, and a clean-contrast subset of pairs that the annotators judged to be about one kind of standard only (median rating at least 2 for the intended standard and at most 1 for every other).
The ratings of how much worse the worse option is did not affect retention; they were used only in a sensitivity analysis that compares moral and normative pairs of the same rated size.
Inter-annotator agreement is reported descriptively as Krippendorff's $\alpha$ for each question \citep{hayes2007answering} (nominal for the better option and the derived class, ordinal for the rating scales) and was not used as a threshold.

The plan required at least 60 retained moral pairs, 60 normative pairs with at least 10 pairs per subtype whose subtype was confirmed, and 24 descriptive pairs; had a minimum been missed, a reserve set of pairs was to be rated by the same annotators before the retained set was fixed; this happened once, for the prudential subtype, as described below.
Retention depended only on the annotations and on this rule.

\paragraph{Outcome of the rating.}
Of the \NratedTotal{} rated pairs, \NaccTotal{} were retained (\Cref{tab:acceptance}): \NaccMoral{} of \NratedMoral{} moral pairs, \NaccNorm{} of 113 normative pairs (prudential \NaccNormP, etiquette \NaccNormE, aesthetic \NaccNormA, epistemic \NaccNormK; subtype confirmed for \NconfNormP, \NconfNormE, \NconfNormA{} and \NconfNormK{} of them) and \NaccDescr{} of \NratedDescr{} descriptive pairs, so every minimum was \MinimaMet.
Agreement was $\alpha=$~\AlphaQone{} for the better option, \AlphaQthree{} for the derived class of standard (\AlphaQthreeClass{} at the level of the three classes the rule uses), \AlphaQtwo{} for how much worse, and \AlphaQfour{} for the moral rating (the four other ratings: \AlphaGradedOthers).
The strict subset holds \NstrictTotal{} pairs and the clean-contrast subset \NcleanTotal{} pairs.
Of the \NrejTotal{} rejected pairs, \NrejMoral{} were moral pairs for which fewer than \TwoThirdsN{} annotators derived a moral standard (the ratings put another standard as high as the moral one), and \NrejNorm{} were normative pairs for which fewer than \TwoThirdsN{} derived a non-moral standard; one of the latter also had a median moral rating above 1.
No pair was rejected for its direction, for an unclear flag or for a missing answer.
The ratings describe how mixed the retained pairs are: moral pairs received a mean moral rating of 2.8 and at most 0.2 on any other standard; etiquette, aesthetic and epistemic pairs received 2.7 to 2.8 on their own standard and at most 0.5 on the others; prudential pairs received 2.1 on the prudential standard and 0.9 on the epistemic one, since the annotators read many pairs of the form ``choosing X based on Y'' as a matter of evidence as well as of self-interest, which is why that subtype needed the second round.
\rev{The retained normative pairs received a mean moral rating of 0.2 (aesthetic 0.0, prudential 0.1, etiquette 0.3, epistemic 0.5), and 61 of the \NaccNorm{} received 0 from every annotator, so they can be considered \emph{mostly} non-moral.}
The annotators' background answers and the per-pair details are in \Cref{app:validation}.
So the annotators saw the contrast we had built in nearly every pair; the pairs they dropped were ones in which a second standard seemed as relevant as the intended one.
\rev{Excluding these \NrejTotal{} pairs ensures that no pair counted as moral was seen mainly as a matter of another standard, and that no pair counted as normative was seen mainly as a moral matter.}

\subsection{Head methods}
\label{app:head-method}

\subsubsection{On the original pairs} The score used to locate the components added the probabilities of ten permissive words (\emph{acceptable, justified, right, permissible, okay, correct, good, fine, moral, ethical}) and subtracted those of ten restrictive words (\emph{wrong, unacceptable, immoral, unjustified, bad, incorrect, impermissible, unethical, unjust, forbidden}), each taken as the first token of the word with a leading space; duplicates were not removed.
Descriptive controls used word lists specific to each pair.
A pair was kept for a model if its permissive version scored strictly higher than its restrictive version after rounding to four decimals.
The intervention replaced a head's output\rev{, taken before the output projection,} at positions 0 to $\min(T_+,T_-)-1$ of the restrictive run with the permissive values, in one direction only.
The top head of each model maximised the mean absolute moral effect minus the mean absolute control effect; the located heads are the ten heads \rev{with the largest values of this difference in} each model.
The share analysis in \Cref{sec:heads-dev} uses the stored per-pair effects: it divides each pair's absolute effects by their sum over heads, tests each head with a permutation test on the shares (Benjamini-Hochberg over the heads of a model), and computes bootstrap intervals over pairs (50{,}000 permutations and 5{,}000 resamples).

\subsubsection{On the new pairs} \paragraph{Score.}
$S$ sums the probabilities of 14 positive and subtracts those of 14 negative answer words, each a single token in the Llama-3, Qwen2.5 and Gemma tokenizers: positive: acceptable, justified, right, permissible, okay, correct, good, fine, moral, ethical, appropriate, proper, great, sensible; negative: unacceptable, unjust, wrong, forbidden, terrible, incorrect, bad, poor, immoral, unethical, inappropriate, improper, awful, foolish.
\ac{LD} is the log-odds version\vio{, $\log P(\text{positive words}) - \log P(\text{negative words})$}; the original score is the one with 10 + 10 words described above.
\paragraph{Intervention.}
Patching of a head's output over the whole sequence, from the permissive into the restrictive run (p2r), at every position, which align exactly because the two versions have equal token counts; secondary: the reverse direction (r2p) and patching at the final position only.
Primed frames reuse a cached forward pass of the shared prefix; the largest difference between the cached and the plain computation was 0.000024 (tolerance 0.0001), 0.2 to 1.5\,\% of the median absolute head effect of each model, and heads with an effect below 0.00001 carry less than 0.01\,\% of an item's total effect.
\paragraph{Runs.}
Frames A1, B and A0, each with all four conditions (p2r and r2p, all positions and final position), cover all \NcandTotal{} candidate pairs for every model; the two 14B models ran the 225 pairs of the first rating round first and the 57 spare pairs after the second round.
All runs were made on one Apple computer with an M2 Max chip and 96\,\ac{GB} of unified memory (\ac{MPS} backend, float32); the head runs took between 6 minutes and 2.4 hours per model for the 12 models up to 8B and about 5 hours per 14B model for the rated pairs, with the spare pairs added later; the stages of the circuit analysis took a few per cent (attention and direct effects, edges), a third (layers) and half (mean-ablation) of a head run per model.
\paragraph{Head sets} (all taken from the search on the original pairs): the located heads (primary)\rev{, the first five of them, the top head alone, and two sets chosen by size of share: the ten heads with the largest mean share of the moral effect on the original pairs, and the 1\,\% of heads with the largest mean share (1 to 19 heads, depending on the size of the model).}
\paragraph{Inclusion.}
A pair enters a model's analysis in a frame if $S(\text{permissive})-S(\text{restrictive})\geq0.02$ (defined on $S$ whatever score the effects use).
A model and frame with fewer than 30 included moral or 30 included norm pairs is reported as having insufficient behavioural contrast and left out of every pooled estimate.
\paragraph{Statistics.}
The item is the unit of analysis everywhere.
Percentile bootstrap with 10{,}000 resamples of items within category; pooled estimates use the same resample for all models.
Per-model tests permute category labels (20{,}000 permutations), with Benjamini-Hochberg correction across the 14 models within each frame and contrast.
No result is called significant on the basis of a per-head test without \ac{FDR} control.

\subsection{Feature methods}
\label{app:feature-method}

\paragraph{Dictionaries.}
Thirteen \rev{dictionaries} \rev{are} TopK \rev{\acp{SAE}} trained for this project on SlimPajama text (context length 512, first four positions excluded from training), \rev{on the hidden state} after the listed block or \rev{on} the output of the first \ac{MLP} block; \rev{the fourteenth is} the Llama Scope dictionary at layer 12 of Llama-3.1-8B \citep{he2024llamascope}\rev{, which we did not train: its authors trained it as a TopK \ac{SAE} and converted it after training to the JumpReLU form that they released and that we use, in which a feature is active when its value exceeds a fixed threshold}.
Widths and sparsities are in \Cref{tab:sae-conditions}; checkpoints and configuration files are released.
\grn{The 13 dictionaries were trained with SAELens \citep{bloom2024saetrainingcodebase} on 500 million tokens of SlimPajama \citep{soboleva2023slimpajama}, with a TopK activation \citep{gao2024scaling}, eight times as many features as the hidden state has dimensions, the encoder initialised as the transpose of the decoder, decoder rows of unit norm, the auxiliary loss for inactive features of \citet{gao2024scaling} with weight 1/32, Adam with a learning rate of $10^{-4}$ (linear warm-up over 5,000 steps, no decay) and batches of 4,096 tokens.}
The loader reproduces the features stored when the components were located exactly for the bespoke Llama-3.2-1B dictionary and for the 8B Llama Scope dictionary ($r=1.00000$ for the 12 Llama \rev{dictionaries}; Qwen as in \Cref{sec:models}).

\paragraph{Statistics on the original pairs.}
The tests on the original pairs compared the mean change $a_{ij}$ (\Cref{eq:feature-change}) between moral pairs and a comparison group with one-sided permutation tests (labels of whole pairs permuted, $(r+1)/(B+1)$ \citep{phipson2010permutation}) and Benjamini-Hochberg correction over the features of one \rev{dictionary} and comparison \citep{benjamini1995controlling}.
The test that swapped the frames encoded the \rev{hidden} states at the final position of Llama-3.2-1B base at layer 7 for moral, norm and neutral content, each in its original frame and with the frames exchanged, and compared the mean activation of the moral-versus-neutral feature sets (validated against the stored features: largest absolute difference 0.000003, identical active sets).

\paragraph{Capture check.}
\Cref{eq:capture}, computed per pair and averaged by category and frame, with the cosine between $d$ and $\hat d$ and the projection of $\hat d$ on $d$.
A \rev{dictionary} with a mean $R^2$ below 0.5 in a frame is flagged for that frame.
The reconstruction gate first used on the original pairs, computed over the whole sequence, is replaced by these statistics at the final token (\Cref{app:audit}).

\paragraph{H5 and H6.}
As specified in \Cref{sec:features-heldout,sec:features-discovery}.
\rev{The feature set of H5 (the located features, found against the neutral pairs) and the two sets of the ablation (the located features, and the features found against the normative pairs)} are the features significant after \ac{FDR} correction in the expanded analyses of the original pairs.
For the Qwen \rev{dictionaries} they were selected under the original tokenisation without a leading token (\Cref{sec:models}).

\paragraph{Ablation that keeps the reconstruction error.}
For a \rev{hidden} state $h$, encoded features $z=E(h)$, decoder $D$ and a feature set $F$, let $z^{\setminus F}$ be $z$ with the coordinates in $F$ set to zero.
The intervened \rev{hidden} state is
\begin{equation}
h^{\setminus F}=h+D(z^{\setminus F})-D(z),
\label{eq:error-preserving}
\end{equation}
which subtracts only the decoded contribution of $F$ and returns $h$ unchanged when $F$ is empty, however poorly the \rev{dictionary} reconstructs \rev{(the error term is kept, as in \citealp{marks2024sparse})}.
The intervention acts at the final position.
Each set is compared with 20 random sets of the same size: for each feature of the set, one of the 20 nearest unused features by mean activation within the frame, over all items that were run.
The random sets cannot reach the same activity when the set holds the most active features (as \rev{the located features do} in frame A1), so the norm of the removed vector is reported for the set and the random sets next to every effect.
Items are the unit of analysis.

\paragraph{Further cue frames.}
Frames A1e (``\stem{} is ethically''), A1l (``\stem{} is logically'') and A1p (``Morally speaking, \stem{} is'', with the first letter of the stem in lower case) keep the priming sentences of A1 and B.
They were run on all candidate pairs with the 14 \rev{dictionaries} of \Cref{tab:sae-conditions}, encoding only the \rev{hidden} state at the final position (no ablation), and analysed on the retained pairs.
In the Llama-3 and Qwen2.5 tokenizers ``ethically'' is two tokens (``eth'' and ``ically'') and ``logically'' one; in A1p the lower-case first word changes the length of one version in 15 of the \NcandTotal{} pairs, which does not matter at the final position.
Activations are compared with those in A1 on the same pairs (paired bootstrap, 10{,}000 resamples, 95\,\% interval, read against 0.5), and the distinction between moral and normative pairs is counted as in H6.
A check on the answers decided whether a frame could be read: ``Morally speaking'' had to raise the gap of the moral pairs, relative to that of the normative pairs, above its value without a cue in at least 5 of the 7 models (it did in 5), and ``logically'' the gap of the epistemic pairs relative to the other normative pairs (it did in none).
The comparison planned for ``logically'' set the moral pairs and the epistemic pairs, each against a mix of prudential, etiquette and aesthetic pairs, at the size of the smallest kind (22 pairs), and asked whether, under ``logically'', the epistemic pairs stand apart in more dictionaries than with ``morally'' or without a cue, and the moral pairs in fewer than with ``morally''; it was not read.
Frame A1s (``\stem{} is socially'') was run afterwards in the same way; its prompts are those of A1 with the last token replaced (``socially'' is one token in both tokenizers).
Its check on the answers required the gap of the etiquette pairs, relative to that of the prudential, aesthetic and epistemic pairs, to be larger than without a cue in at least 5 of the 7 models (it was in all 7).
Its planned comparison set the moral and the etiquette pairs, each against the prudential, aesthetic and epistemic pairs, with each of these three kinds against the other two as calibration, at 22 pairs per group and 20 subsamples, in A1, B and A1s; it was to be read only if the moral pairs stood apart in at least four more dictionaries with ``morally'' than without a cue, and they did in three more (3 against 0).
The secondary measure is $\log\frac{n+1}{c+1}$, with $n$ the count of a group and $c$ the largest calibration count, compared between frames in the 12 dictionaries whose counts add up to at least 5.

\section{The heads on the new pairs, in detail}
\label{app:heads}

\subsection{The located heads, tested on pairs they have never seen}
\label{sec:heads-heldout}
This section uses the new pairs retained by the annotators and asks the question of the introduction directly: do the located heads carry a larger share of the effect for moral than for normative pairs of the same form, with the cue (H1) and without it (H2)?
The primary analysis is the one \grn{set out} in the plan: the located heads, direction p2r, all positions, score $S$, inclusion threshold 0.02, all retained pairs; every other setting is a sensitivity analysis.

\paragraph{Hypotheses and reading bands.}
\textbf{H1}: $\theta(\mathrm{A1})>0$; the located heads carry a larger share of the effect for moral than for normative contrasts of the same form \rev{with the moral cue}.
\textbf{H2}: $\theta(\mathrm{B})>0$; the same holds without the moral cue.
Each is tested at $\alpha=0.025$ with a two-sided 97.5\,\% bootstrap interval.
The interval of the pooled ratio $\exp\theta$ is read against five bands, from top to bottom, and the first band that fits is used (\Cref{tab:bands}).
\vio{The margin of 0.8 to 1.25}\grn{, part of the analysis plan (\Cref{app:prereg}),} \vio{is symmetric on the logarithmic scale (a quarter more, or a fifth less)}\grn{; the same range is a common convention for the equivalence of a ratio of two means \citep{fda2001bioequivalence}.
Every ratio is reported with its interval}.

\begin{table}[t]
\centering
\caption{\textbf{Reading bands for H1 and H2.}
Applied to the 97.5\,\% interval of the pooled ratio $\exp\theta$; rows are read from top to bottom and the first that fits is used.}
\label{tab:bands}
\small
\begin{tabularx}{\linewidth}{@{}clY@{}}
\toprule
& Interval of $\exp\theta$ & Reading \\
\midrule
1 & entirely above 1.5 & strong moral preference of the located heads \\
2 & entirely inside 0.8 to 1.25 & no meaningful content specificity: the located heads serve normative evaluation in general (if the
interval also excludes 1, this is said, and the difference is described as within the margin) \\
3 & entirely above 1.0 & weak to moderate moral preference \\
4 & entirely below 1.0 & the located heads carry a smaller share for moral than for norm content \\
5 & anything else & inconclusive \\
\bottomrule
\end{tabularx}

\end{table}

\paragraph{Behavioural inclusion.}
A pair can only show what the heads carry if the model has a preference on it, so the analysis keeps, per model and frame, the pairs with a behavioural gap of at least 0.02.
Of the \NaccTotal{} retained pairs, between 57 and 80 moral pairs and between 80 and 107 norm pairs per model and frame pass the inclusion threshold (\Cref{tab:heldout-heads}), so all 14 models enter both primary tests.
Descriptive pairs are included less often in \rev{the frames with the cue}, as the original pairs had led us to expect: Qwen2.5-3B includes only 3 of 40 in frame A0 and Qwen2.5-14B-\ac{IT} only 18 in frame A1, below the minimum of 20, so these two cells (one model in one frame each) are reported as having insufficient behavioural contrast with descriptive pairs and are left out of the corresponding pooled estimates.
In frame B every model includes at least 34 descriptive pairs.
The mean behavioural gap is larger for moral than for norm pairs in every frame, by less in frame B (for example 0.76 against 0.43 in Llama-3.1-8B in frame A1 and 0.61 against 0.60 in frame B)\rev{.
The analysis works with shares and not with raw effects, so that a kind of pair on which the effects of all heads are larger does not make the located heads look larger on it}.

\paragraph{Results for H1 and H2.}
\emph{The located heads carry the same share for moral as for normative pairs, with and without the cue.}
In frame A1 the pooled ratio is $\exp\hat\theta(\mathrm{A1})=$~\HOneRatio{} (\HOneLevel\,\% \ac{CI} \HOneCI), which falls in band~\HOneBand.
In frame B it is $\exp\hat\theta(\mathrm{B})=$~\HTwoRatio{} (\HTwoCI), band~\HTwoBand{} (\Cref{fig:h1h2}).
The per-model ratios, with tests corrected across the 14 models within each frame, are in \Cref{tab:heldout-heads}.
Both intervals lie inside the margin of 0.8 to 1.25; the interval of frame B also excludes 1, so the located heads carry a slightly smaller share for moral than for normative pairs without the cue, a difference that is within that margin.

In frame A0, \rev{with the cue but without the priming}, the ratio is \HZeroRatio{} (95\,\% \ac{CI} \HZeroCI)\rev{: inside the same margin and above 1, so with the cue alone the located heads carry a slightly larger share for moral pairs, a difference that is absent with the priming sentences (frames A0 and A1 differ only in those sentences) \rev{and about two thirds as large when ``socially'' stands in the place of ``morally'' (1.08, 95\,\% \ac{CI} [1.02, 1.14], 14 models; \Cref{sec:further}).}
As a multiple of the share of ten heads chosen at random, the share of the located heads on moral pairs is the same in the two frames (5.4), and the priming sentences raise their share on normative pairs (from 4.9 to 5.5) and on descriptive pairs (from 4.3 to 4.6)}.
No single model shows a strong moral preference either: the per-model ratios range from 0.83 to 1.22 in frame A1 and from 0.70 to 1.08 in frame B\rev{, and from 0.92 to 1.27 in frame A0, where four models lie above 1 at $q<0.05$ (Llama-3.2-1B-\ac{IT}, the two Llama-3.1-8B models and Qwen2.5-14B-\ac{IT})}; two models differ from 1 at $q<0.05$ in A1, in opposite directions (Llama-3.1-8B at 0.83, Qwen2.5-14B at 1.22), and three fall below 1 in B (Llama-3.2-3B-\ac{IT} 0.81, Qwen2.5-3B 0.84, Qwen2.5-7B 0.70).
\rev{The third reading does not fit either: without the word ``morally'' the located heads still carry four to five times the share of ten heads chosen at random (\Cref{fig:design}; on moral and on normative pairs, at least twice in every model).}
The located heads thus do the same share of the work when a model judges a prudential, an etiquette, an aesthetic or an epistemic pair as when it judges a moral one.

\begin{table}[t]
\centering
\caption{\textbf{The located heads, per model.}
Included pairs per frame (moral/norm/descriptive) and selectivity $\rho(m,f)$ of the located heads with 95\,\% interval and Benjamini-Hochberg $q$ across the 14 models; the last row is the pooled ratio with its 97.5\,\% (A1, B) or 95\,\% (A0) interval.}
\label{tab:heldout-heads}
\footnotesize
\setlength{\tabcolsep}{2.5pt}
\begin{tabular}{lcccccc}
\toprule
& \multicolumn{3}{c}{Included pairs (moral/norm/descr.)} & \multicolumn{3}{c}{$\rho(m,f)$ [95\,\% \ac{CI}]; $q$} \\
\cmidrule(lr){2-4}\cmidrule(lr){5-7}
Model & A1 & B & A0 & A1 & B & A0 \\
\midrule
Llama-3.2-1B & 78/97/39 & 73/91/35 & 74/82/27 & 1.07 [0.99, 1.15]; 0.216 & 1.08 [0.99, 1.18]; 0.130 & 1.01 [0.94, 1.10]; 0.768 \\
Llama-3.2-1B-\ac{IT} & 69/92/38 & 72/95/37 & 72/86/24 & 1.13 [1.01, 1.27]; 0.135 & 1.03 [0.91, 1.18]; 0.680 & 1.23 [1.12, 1.35]; 0.001 \\
Llama-3.2-3B & 79/104/40 & 78/100/40 & 78/90/25 & 0.93 [0.83, 1.04]; 0.304 & 0.88 [0.77, 1.00]; 0.123 & 1.11 [0.99, 1.24]; 0.154 \\
Llama-3.2-3B-\ac{IT} & 72/103/39 & 69/103/39 & 77/95/30 & 0.97 [0.86, 1.09]; 0.657 & 0.81 [0.72, 0.91]; 0.009 & 1.03 [0.93, 1.13]; 0.768 \\
Gemma-2B & 74/99/39 & 72/95/40 & 75/93/34 & 0.93 [0.86, 1.00]; 0.135 & 0.91 [0.83, 0.99]; 0.063 & 1.01 [0.94, 1.09]; 0.828 \\
Gemma-2B-\ac{IT} & 73/98/37 & 67/88/34 & 73/92/21 & 1.06 [0.96, 1.16]; 0.350 & 0.99 [0.89, 1.09]; 0.806 & 1.02 [0.93, 1.11]; 0.768 \\
Qwen2.5-3B & 78/98/39 & 75/101/40 & 57/80/3 & 1.10 [0.98, 1.24]; 0.255 & 0.84 [0.74, 0.94]; 0.020 & 1.15 [0.98, 1.35]; 0.178 \\
Qwen2.5-3B-\ac{IT} & 76/103/40 & 76/104/40 & 77/101/36 & 1.10 [0.94, 1.27]; 0.328 & 1.03 [0.91, 1.17]; 0.680 & 1.12 [0.97, 1.29]; 0.198 \\
Qwen2.5-7B & 79/104/40 & 77/107/40 & 78/99/28 & 0.91 [0.70, 1.18]; 0.545 & 0.70 [0.58, 0.85]; 0.006 & 0.92 [0.78, 1.09]; 0.516 \\
Qwen2.5-7B-\ac{IT} & 75/106/40 & 77/104/40 & 77/92/28 & 0.87 [0.60, 1.27]; 0.545 & 0.87 [0.65, 1.15]; 0.421 & 1.11 [0.91, 1.35]; 0.483 \\
Llama-3.1-8B & 78/104/38 & 77/104/40 & 78/98/31 & 0.83 [0.74, 0.93]; 0.016 & 0.86 [0.76, 0.98]; 0.063 & 1.27 [1.11, 1.43]; 0.001 \\
Llama-3.1-8B-\ac{IT} & 78/107/37 & 78/103/40 & 77/104/30 & 0.90 [0.80, 1.00]; 0.135 & 0.86 [0.75, 0.97]; 0.056 & 1.27 [1.12, 1.44]; 0.001 \\
Qwen2.5-14B & 79/99/36 & 79/106/40 & 80/101/32 & 1.22 [1.08, 1.38]; 0.017 & 0.91 [0.78, 1.06]; 0.343 & 1.21 [1.03, 1.41]; 0.053 \\
Qwen2.5-14B-\ac{IT} & 79/99/18 & 78/104/40 & 80/95/32 & 1.03 [0.89, 1.19]; 0.657 & 0.87 [0.75, 1.02]; 0.130 & 1.24 [1.05, 1.46]; 0.048 \\
\midrule
Pooled $\exp\hat\theta$ [\ac{CI}] & & & & 1.00 [0.92, 1.07] & 0.90 [0.84, 0.96] & 1.12 [1.05, 1.18] \\
\bottomrule
\end{tabular}

\end{table}

\paragraph{H3: normative or merely evaluative?}
A head set that treats moral and normative pairs alike could still be a set for evaluation of any kind, including judgments about things rather than conduct.
The descriptive pairs \rev{were written to decide this: by the reading of the plan, a larger share of the located heads for normative than for descriptive contrasts means that the located heads are normative rather than merely evaluative}.
We compute the same ratio for norm against descriptive pairs and for moral against descriptive pairs, in each frame: \HThreeMDAone{} for moral against descriptive pairs and \HThreeNDAone{} for norm against descriptive pairs in frame A1 (13 models), \HThreeMDB{} and \HThreeNDB{} in frame B (14 models), and 1.27 [1.19, 1.35] and 1.14 [1.08, 1.21] in frame A0 (13 models).
The located heads therefore carry a modestly larger share of the effect for normative than for descriptive contrasts, by about a fifth\rev{; for moral against descriptive contrasts the difference is a fifth with the cue and a tenth without it.
\rev{This would point} to a pathway that is normative rather than merely evaluative, and not moral in particular.
Two facts stand against that reading.
On descriptive pairs the located heads still carry four times the share of ten heads chosen at random (\Cref{fig:design}).
And the descriptive pairs differ from the other pairs in form as well as in content: all of them are noun phrases about a thing (``The taste of ripe strawberries''), with no agent and in one sentence template, while the moral and the normative pairs are phrases about someone's action in four templates.
The difference of a fifth may therefore come from the judgment of conduct or from the form of the phrase.
\rev{Descriptive pairs about a person separate the two (\Cref{sec:further}; 58 pairs, 14 models): without the cue the located heads carry on the pairs about a person the share that they carry on normative pairs (ratio of normative to person 1.00, 95\,\% \ac{CI} [0.92, 1.09]) and 1.22 times the share that they carry on the descriptive pairs about a thing ([1.11, 1.32]); with the cue the pairs about a person lie halfway.
Without the cue the whole difference of a fifth, and with the cue about half of it, is therefore a difference between phrases about a person and phrases about a thing, and the located heads serve evaluation in general, with a smaller share when the phrase is about a thing.}}
Contrasts with descriptive pairs need at least 20 included descriptive pairs per model and frame.
On the original pairs, descriptive pairs drew less probability onto the answer words in frame A1 (0.21 to 0.56 per model, against 0.53 to 0.92 for moral pairs), so some models may be reported as insufficient there.

\paragraph{H4: base to instruct.}
H4 asks whether instruction tuning keeps the same heads in place.
For each of the seven \vio{base models and its \ac{IT} version} we compute the correlation between the base and the instruct profiles of mean shares on moral pairs in frame A1 (criterion $r\geq0.8$), ask whether the top head of the base model ranks in the instruct top ten, and report the split-half reliability of each profile as the ceiling such a correlation can reach: \Cref{tab:h4}.
The profile correlation is 0.88 to 0.96 in six \vio{of the seven base models} and 0.78 in Qwen2.5-7B, against split-half reliabilities of 0.94 to 0.99, so instruction tuning preserves most of the pattern across heads on new moral pairs, as it did on the original ones.
The top head of the base model ranks in the instruct top ten in \HFourMet{} \vio{base models} (Llama-3.2-3B, Llama-3.1-8B, Qwen2.5-3B, Qwen2.5-14B) and at rank 12, 19 and 74 in Llama-3.2-1B, Gemma-2B and Qwen2.5-7B.
The prediction as stated, both criteria in every \vio{base model}, is not met; what is preserved is the profile, not the identity of the first head.
\vio{The two largest models keep the pattern best: the profile correlation is 0.95 in Llama-3.1-8B and 0.96 in Qwen2.5-14B, and the base model's top head ranks 4th and 1st after tuning.
On the original pairs the correlation also rises with size within each family (\Cref{sec:heads-dev}), and the located heads, searched separately in each model, have 8 and 6 of their ten heads in common between the base and the \ac{IT} versions of these two models, against 2 to 5 in the five smaller ones.
With seven base models this is a description, which we did not test.}

\begin{table}[t]
\centering
\caption{\textbf{H4 on the new pairs, frame A1.}
Correlation of the base and instruct profiles of mean shares over heads, whether the top head of the base model is among the ten largest shares of the instruct model, and the split-half reliability of each profile.}
\label{tab:h4}
\small
\begin{tabular}{lcccc}
\toprule
& Profile $r$ & Top head of the base model & \multicolumn{2}{c}{Split-half $r$} \\
\cmidrule(lr){4-5}
\vio{Base model} & base vs \ac{IT} & in the \ac{IT} top 10 (its rank) & base & \ac{IT} \\
\midrule
Llama-3.2-1B & 0.90 & no (12) & 0.99 & 0.98 \\
Llama-3.2-3B & 0.91 & yes (4) & 0.99 & 0.99 \\
Llama-3.1-8B & 0.95 & yes (4) & 0.98 & 0.99 \\
Gemma-2B & 0.89 & no (19) & 0.98 & 0.98 \\
Qwen2.5-3B & 0.88 & yes (1) & 0.99 & 0.98 \\
Qwen2.5-7B & 0.78 & no (74) & 0.97 & 0.94 \\
Qwen2.5-14B & 0.96 & yes (1) & 0.99 & 0.98 \\
\bottomrule
\end{tabular}

\end{table}

\paragraph{Sensitivity analyses.}
We also checked whether the answer changes under other reasonable choices of measurement, head set, item set or comparison.
\rev{The following} are reported with 95\,\% intervals (\Cref{tab:sensitivity}): frame A0 next to A1 (if the two fall in different bands, the priming changed the circuit, and we say so before interpreting the comparison of A1 with B); direction r2p; patching at the final position only; score \ac{LD}; \rev{other head sets (the top head alone, the first five located heads, the ten heads with the largest share and the 1\,\% of heads with the largest share; \Cref{app:head-method})}; moral against each subtype of norm, on the pairs whose subtype the annotators confirmed; raw effects with the same ratio over all heads shown next to them; a regression of the logit of the share on category, with the log behavioural gap, template type, token count and tokens after the edit as covariates; moral and norm items compared within quintiles of the behavioural gap, and within levels of the annotators' rating of how much worse the worse option is; the strict rating subset; inclusion threshold 0.10; all rated items, whatever the annotators said; and the clean-contrast subset of the ratings (pairs in which the intended standard was clearly present and every other standard at most slight).

Every point estimate stays inside the band of the primary result, 0.95 to 1.12 in frame A1 and 0.86 to 1.03 in frame B, and so does every interval but one: the smallest head set, \rev{the top head alone}, is the least precise (A1 1.12 [0.98, 1.26]), and patching at the final position only gives the lowest value, in frame B (0.86 [0.80, 0.91]).
Against each subtype of norm separately the ratios are 0.95 to 1.09 in A1 and 0.91 to 0.94 in B, with intervals that include 1 or end just below it.
The raw effects of the located heads are smaller on moral than on norm pairs by about the same factor as the raw effects of all heads (0.76 against 0.73 in A1, 0.74 against 0.79 in B), so nothing in the raw numbers singles the set out either.
Adjusting the logit of the share for the behavioural gap, the template type and the token counts moves the moral-minus-norm difference slightly below zero (mean $-0.08$ in A1 and $-0.13$ in B, with intervals below zero in 5 and 9 of the 14 models and above zero in 1 and 0), and comparing pairs with the same behavioural gap or the same rated size of difference leaves the ratios where they are (0.97 and 0.99 in A1, 0.88 and 0.90 in B).
None of these changes of measure, head set or pairs makes the located heads look moral.

\begin{table}[t]
\centering
\caption{\textbf{Sensitivity analyses for the located heads}: pooled moral/norm ratio $\exp\hat\theta$ with its 95\,\% interval (97.5\,\% for the primary analysis).
The four rows without an interval are summaries over the 14 models: geometric means of the per-model ratios (raw effects, and the comparisons at equal gap and at equal rated size) or the mean per-model difference (regression), with the range across models in parentheses.}
\label{tab:sensitivity}
\small
\setlength{\tabcolsep}{4pt}
\begin{tabularx}{\textwidth}{>{\raggedright\arraybackslash}Xcc}
\toprule
Analysis & A1 & B \\
\midrule
Primary analysis: located heads, p2r, all positions, $S$, threshold 0.02, retained pairs & 1.00 [0.92, 1.07] & 0.90 [0.84, 0.96] \\
Frame A0 in place of A1 & 1.12 [1.05, 1.18] & - \\
Direction r2p & 1.05 [1.00, 1.12] & 1.03 [0.95, 1.11] \\
Patching at the final position only & 1.04 [0.98, 1.10] & 0.86 [0.80, 0.91] \\
Score \ac{LD} & 1.02 [0.97, 1.06] & 0.98 [0.93, 1.03] \\
Original score (10 + 10 words) & 1.00 [0.93, 1.06] & 0.91 [0.86, 0.96] \\
\rev{Head set: the top head alone} & 1.12 [0.98, 1.26] & 0.95 [0.85, 1.06] \\
\rev{Head set: the first five located heads} & 1.02 [0.94, 1.10] & 0.91 [0.85, 0.97] \\
\rev{Head set: the ten heads with the largest share} & 1.04 [0.97, 1.11] & 0.90 [0.85, 0.97] \\
\rev{Head set: the 1\,\% of heads with the largest share} & 1.02 [0.94, 1.10] & 0.88 [0.81, 0.94] \\
Moral against one subtype of norm (pairs whose subtype was confirmed): prudential & 1.05 [0.94, 1.18] & 0.94 [0.85, 1.06] \\
\quad etiquette & 0.95 [0.86, 1.05] & 0.94 [0.86, 1.04] \\
\quad aesthetic & 1.09 [0.99, 1.20] & 0.91 [0.84, 0.99] \\
\quad epistemic & 0.99 [0.89, 1.10] & 0.91 [0.83, 0.99] \\
Raw effects of the located heads, moral/norm (same ratio over all heads) & 0.76 (0.28 to 1.26); all heads 0.73 & 0.74 (0.53 to 1.30); all heads 0.79 \\
Regression on the logit of the share (gap, template, tokens): adjusted moral $-$ norm & $-0.08$ (unadjusted $-0.02$) & $-0.13$ (unadjusted $-0.12$) \\
Within quintiles of the behavioural gap & 0.97 (0.83 to 1.13) & 0.88 (0.69 to 1.03) \\
Within levels of the rated size of the difference & 0.99 (0.86 to 1.24) & 0.90 (0.78 to 1.04) \\
Strict rating subset & 0.96 [0.90, 1.02] & 0.88 [0.83, 0.94] \\
Clean-contrast subset & 1.00 [0.93, 1.07] & 0.90 [0.85, 0.95] \\
Inclusion threshold 0.10 & 0.99 [0.93, 1.05] & 0.89 [0.83, 0.94] \\
All rated pairs & 1.00 [0.94, 1.06] & 0.90 [0.85, 0.96] \\
\bottomrule
\end{tabularx}

\end{table}

\subsection{Heads selected on the new pairs}
\label{sec:heads-discovery}
Perhaps the located heads were simply not the right ones, and heads chosen on the new moral pairs would be selective.
The new pairs allow this check without any use of the original pairs: ten heads are chosen on a random half of the new moral pairs, and the moral/norm ratio is computed on the other half, over 200 splits.
Its mirror image runs alongside: ten heads chosen on half of the norm pairs, and the norm/moral ratio on the other half.
The mirror image is the control: any two categories have somewhat different head profiles, so heads picked on one of them will favour it a little on fresh pairs.
If both selections come out above 1 by a similar factor, a ratio above 1 shows only that the two profiles differ, not that the heads chosen on moral pairs are special.
The same caution applies to the located heads, for which the original pairs offer no counterpart selected on norm pairs.

Heads chosen on moral pairs: geometric mean over models of 1.04 in frame A1 (range 0.87 to 1.22; the 2.5th percentile over splits is above 1 in 4 of 14 models), 0.99 in frame B (0.83 to 1.12; 1 model) and 1.18 in frame A0 (0.99 to 1.36; 11 models).
Heads chosen on norm pairs: 1.01 in A1 (0.84 to 1.25; above 1 in 3 models), 1.12 in B (0.93 to 1.45; 8 models) and 0.93 in A0 (0.79 to 1.14; 2 models).
With the moral cue, neither selection is more selective for its own category than for the other.
The two other frames show a moderate asymmetry that runs in opposite directions: with the cue but without the priming (frame A0) the heads chosen on moral pairs keep a small advantage on fresh pairs, with the priming but without the cue (frame B) the heads chosen on norm pairs do, and with both neither does.
\vio{So the heads chosen on the new moral pairs show no moral excess beyond the control either.}

A selection aimed directly at specificity does no better (an \rev{exploratory} analysis).
Choosing, on half of the moral and half of the normative pairs, the ten heads whose moral share most exceeds their normative share, the held-out ratio of moral to normative share is 1.1 to 1.7 across models and frames; but heads chosen the other way round reach 1.0 to 1.6 on fresh pairs, and heads chosen to separate two kinds of norm from the other two reach 0.9 to 1.7.
By the rule \rev{of this analysis}, which asks the lower end of the interval over 200 splits, for heads chosen on moral pairs, to lie above 1 and above the upper end of every split between kinds of norm, no model shows heads specific to moral pairs in any frame.
Counting the heads whose share differs between moral and normative pairs gives more than the splits between kinds in several of the larger models (with the cue, 284 heads against at most 7 in Qwen2.5-14B-Instruct); as with the neurons of \Cref{sec:neurons}, \vio{on fresh pairs such a difference is no larger than the difference that the same selection finds between two kinds of norm (held-out ratios up to 1.6 for heads chosen the other way round and up to 1.7 for splits between kinds of norm).}

\section{From heads to circuits: the analyses of the circuit}
\label{sec:circuits}
The analyses so far test what the located heads respond to.
They do not say whether the models need the located heads, what these heads read, which components feed them, or where in the network the difference between a moral and a normative contrast arises.
Each of these questions has a moral and a general answer: a set of heads that the model needs for moral pairs but not for normative ones, that reads something only moral pairs contain, or that is fed differently on moral pairs, would be a moral circuit after all.
The analyses in this section ask these questions on the retained new pairs, in frames A1 and B, for all 14 models; in every case the answer is the general one.
We report them as exploratory \rev{(\Cref{app:prereg})}.
Throughout, means are taken over the included pairs of a category, and ``fraction of the gap'' refers to the behavioural gap $S(\text{permissive})-S(\text{restrictive})$.

\paragraph{Does the model need the located heads?}
It does, but not for moral pairs in particular.
Head patching measures what a head carries from one prompt to the other; it does not show that the model depends on it.
We therefore replace the output of every head in a set by its mean over all prompts of the frame (mean-ablation\rev{, as in \citealp{wang2023interpretability}}), either at every position except the first or at the final position only, and measure how much of the gap disappears, as the ratio $1-\overline{\text{gap}}_{\text{ablated}}/\overline{\text{gap}}_{\text{clean}}$.
The reference is 20 random sets of ten heads drawn from outside the located sets \vio{without regard to layer}, so that the located heads are judged by their rank among them (a rank of 0 means that the located heads remove more of the gap than every random set).
The same is done for \rev{the ten heads with the largest share} and for \rev{two smaller sets, the first five located heads and the top head alone,} against random sets of their size, and separately for moral and for norm pairs: a set that the model needs for moral contrasts but not for normative ones would remove the moral gap and leave the normative gap.

\begin{figure}[t]
\centering
\includegraphics[width=\linewidth]{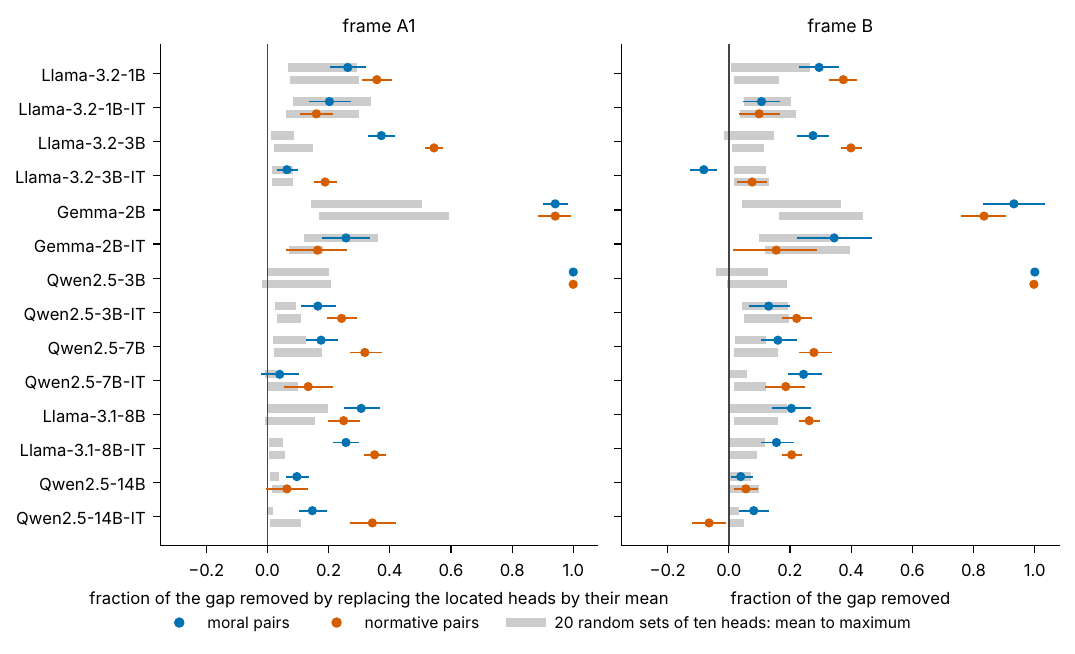}
\caption{\textbf{Necessity of the located heads, all positions.}
Fraction of the behavioural gap removed by replacing the output of the located heads by its mean at every position, on moral (blue) and \rev{normative} (orange) pairs with 95\,\% intervals, in frames A1 and B; the grey bar runs from the mean to the maximum of 20 random sets of ten heads.
The located heads are needed in most models, and at least as much for normative as for moral pairs.}
\label{fig:necessity}
\end{figure}

Replacing the output of the located heads by its mean at every position removes a median of 23\,\% of the gap on moral pairs in frame A1 and 18\,\% in frame B, more than every one of the 20 random sets in 10 of 14 models in each frame (\Cref{fig:necessity}; the random sets remove 1\,\% on average and at most 11 to 14\,\%)\grn{; random sets that take one head from the layer of each located head (exploratory) remove a median of 3\,\% and 2\,\%, and the located heads remove more than every one of them in 10 and in 8 of the 14 models}.
The located heads are needed, but not for moral pairs in particular: on norm pairs the same ablation removes a median of 28\,\% and 21\,\%, and the difference between moral and norm pairs is negative or indistinguishable from zero in every model in frame A1 (interval below zero in 7 models, above zero in none) and in 12 of 14 in frame B (below zero in 5, above in 2).
Descriptive pairs lose a similar fraction (medians 25\,\% and 14\,\%).

Ablating at the final position only removes less, a median of 7\,\% on moral and 11 to 13\,\% on norm pairs, with the located heads above every random set on moral pairs in \NecessityABeats{} models in A1 and \NecessityBBeats{} in B (\Cref{tab:necessity}).
\rev{The ten heads with the largest share remove} more than the located heads (median 41\,\% in A1), \rev{the first five located heads} about 10\,\% and \rev{the top head alone} about 2\,\%.
In two small models, Qwen2.5-3B and Gemma-2B, replacing any ten heads by their mean at every position destroys the gap entirely (fractions near 1 for every category and for some random sets), which says more about the fragility of these models under mean-ablation than about the located heads.

\begin{table}[t]
\centering
\caption{\textbf{Necessity of the located heads.}
Fraction of the clean gap removed by replacing the output of the located heads by its mean \vio{at the final position only (\Cref{fig:necessity} and the text give the removal at every position)}, with its rank among 20 random sets of ten heads, for moral and for \vio{normative} pairs in frames A1 and B.}
\label{tab:necessity}
\small
\setlength{\tabcolsep}{4pt}
\begin{tabular}{lcccccc}
\toprule
& \multicolumn{3}{c}{A1} & \multicolumn{3}{c}{B} \\
\cmidrule(lr){2-4}\cmidrule(lr){5-7}
Model & moral & norm & rank/20 (moral, norm) & moral & norm & rank/20 (moral, norm) \\
\midrule
Llama-3.2-1B & 0.14 & 0.17 & 0, 0 & 0.29 & 0.30 & 0, 0 \\
Llama-3.2-1B-\ac{IT} & -0.03 & -0.01 & 18, 15 & 0.07 & 0.10 & 0, 0 \\
Llama-3.2-3B & 0.31 & 0.44 & 0, 0 & 0.22 & 0.30 & 0, 0 \\
Llama-3.2-3B-\ac{IT} & 0.08 & 0.16 & 0, 0 & 0.06 & 0.10 & 0, 0 \\
Gemma-2B & 0.04 & 0.03 & 6, 9 & 0.28 & 0.22 & 0, 0 \\
Gemma-2B-\ac{IT} & 0.02 & 0.00 & 6, 14 & 0.05 & 0.03 & 3, 8 \\
Qwen2.5-3B & -0.02 & 0.00 & 16, 10 & 0.06 & 0.08 & 0, 0 \\
Qwen2.5-3B-\ac{IT} & 0.10 & 0.13 & 0, 0 & 0.11 & 0.19 & 0, 0 \\
Qwen2.5-7B & 0.03 & 0.07 & 1, 2 & 0.08 & 0.12 & 0, 0 \\
Qwen2.5-7B-\ac{IT} & 0.00 & 0.03 & 10, 0 & 0.02 & 0.05 & 1, 0 \\
Llama-3.1-8B & 0.15 & 0.17 & 0, 0 & 0.14 & 0.13 & 0, 0 \\
Llama-3.1-8B-\ac{IT} & 0.12 & 0.13 & 0, 0 & 0.16 & 0.13 & 0, 0 \\
Qwen2.5-14B & 0.08 & 0.14 & 0, 0 & 0.06 & 0.08 & 0, 0 \\
Qwen2.5-14B-\ac{IT} & 0.07 & 0.15 & 0, 0 & 0.03 & 0.07 & 0, 0 \\
\bottomrule
\end{tabular}

\end{table}

\paragraph{What the located heads read, and what they write.}
The located heads read the cue and the edited words, the same way on moral and on normative pairs, and they write little to the answer directly.
For every head we record how much of its attention at the final position goes to the edited token(s), to the cue ``is morally'' or ``is'', to the priming sentences and to the rest of the stem, and its direct contribution to the logit difference between the positive and the negative answer words (an exact decomposition of the final logits over the embedding, every head and every \ac{MLP} block\rev{; \citealp{elhage2021mathematical}}).
The located heads are compared with the average head: a head that reads the edited token and writes the answer behaves differently from one that reads the word ``morally'', and the comparison between frames A1 and B shows which of the two a head is.

\begin{figure}[t]
\centering
\includegraphics[width=\linewidth]{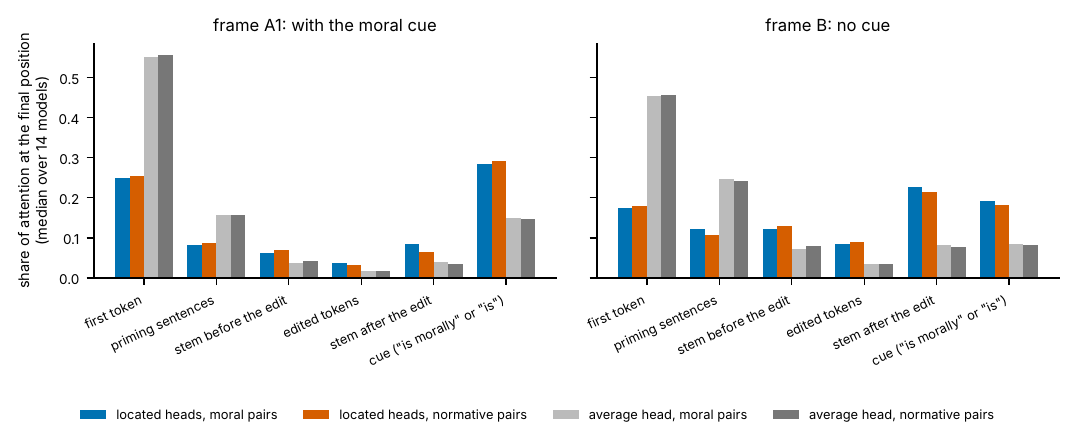}
\caption{\textbf{What the located heads read.}
Share of the attention of the located heads and of the average head at the final position, by part of the prompt, on moral and on normative pairs, with the cue (left) and without it (right); medians over the 14 models.
The located heads look at the cue and at the stem, the average head mostly at the first token; the two categories get the same treatment.}
\label{fig:attention}
\end{figure}

At the final position the located heads look at the cue (\Cref{fig:attention}): a median of 28\,\% of their attention falls on ``is morally'' in frame A1, against 15\,\% for the average head, and 19\,\% on ``is'' in frame B against 8\,\%; they attend more than the average head to the edited tokens (4 against 2\,\% in A1, 9 against 4\,\% in B) and to the rest of the stem after the edit (23 against 8\,\% in B), and much less to the first token (25 against 55\,\%), where the average head parks half of its attention.
These shares are the same on moral and on norm pairs (differences of 0.02 or less in every group): the located heads do not read moral pairs differently.

What they write is small: the direct contribution of the located heads to the logit difference is a median of 2.2\,\% of the total in A1 and 1.4\,\% in B (\rev{the ten heads with the largest share}: 5.4 and 1.9\,\%), while all heads together carry 33 and 27\,\% and the \ac{MLP} blocks 73 and 74\,\%.
Of the ten heads with the largest direct effect on the new moral pairs, a median of one per model is a located head in frame A1 and none in frame B.
The located heads therefore act through later components rather than on the answer directly, and they are heads of the construction that carries the evaluation, not of its moral content.

\begin{figure}[t]
\centering
\includegraphics[width=0.88\linewidth]{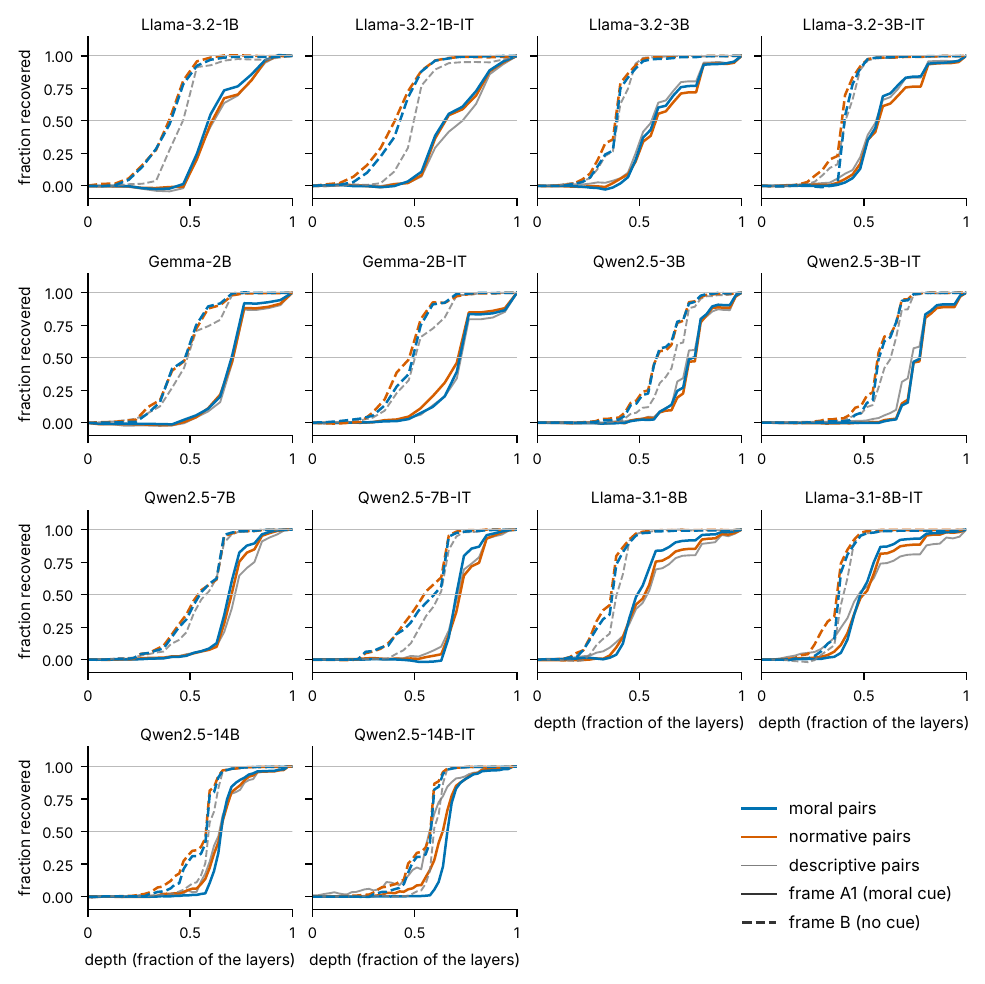}
\caption{\textbf{Layer profiles.}
Fraction of the behavioural gap recovered when the \rev{hidden state} at the final position is patched from the permissive into the restrictive run, layer by layer (depth normalised), for moral (blue), \rev{normative} (orange) and descriptive (grey) pairs, in frame A1 (solid) and frame B (dashed).
Moral and \rev{normative} pairs recover the gap at the same layers in every model; the moral cue of frame A1 delays the arrival at the final position from about half to about two thirds of the depth.}
\label{fig:layers}
\end{figure}

\paragraph{Where the difference arises.}
The difference arises at the same layers for moral and for normative pairs; what changes the picture is the frame.
Patching one layer at a time, we record how much of the gap is recovered by copying the \ac{MLP} output, the attention output, the \rev{hidden state} at the final position, or the \rev{hidden state} at the edited token(s) from the permissive into the restrictive run.
The resulting profiles show in which layers the edit is processed, when the information moves from the edited token to the final position, and whether the moral cue of frame A1 changes that picture relative to frame B (\Cref{fig:layers}).
\vio{A moral circuit could have made the moral difference arrive at a depth of its own; it does not.
This fits a shared pathway, although this test cannot exclude a moral computation that runs at the same layers as the shared one.}

The profiles of moral and norm pairs lie on top of each other in every model: the layer at which the final position first recovers half of the gap is the same for the two categories in 7 of 14 models in frame A1 and within one layer in six of the other seven, and the same in 11 of 14 in frame B.
What moves the profile is the frame, not the content: with the moral cue the final position reaches half of the gap at a median of 0.67 of the depth (0.48 to 0.80 across models), without it at 0.53 (0.39 to 0.63).
The edited tokens carry half of the gap until 0.21 to 0.27 of the depth for moral and norm pairs and until 0.41 to 0.47 for descriptive pairs, whose difference stays local for longer.
Patching the first \ac{MLP} block at every position recovers 70 to 80\,\% of the gap in every model and category, because it carries the identity of the edited word; the attention contribution peaks at about 0.4 of the depth.

\paragraph{What feeds the located heads.}
The located heads are fed by the same components, from the same positions, for both categories.
For each located head we estimate, for every upstream head, \ac{MLP} block and the embedding, the effect of patching that one connection into the head's query, key or value input (edge attribution patching \citep{syed2023attribution}), split by the position at which the signal arrives: the edited tokens, the rest of the stem or the cue.
The estimate is linear; on the original pairs it tracked exact edge patching at $r=0.93$ to $0.997$ in eight models and 0.55 to 0.84 in six (\Cref{app:software}).

The signal that reaches a located head arrives at three places in roughly equal measure: a median of 30 to 40\,\% of its incoming score at the edited tokens, 23 to 28\,\% at the rest of the stem after the edit and 21 to 30\,\% at the cue, with no difference between moral and norm pairs beyond a few points; the strongest single source is an \ac{MLP} block in about 60\,\% of the receiver inputs.
The first-order node scores rank the heads in about the same order as the exact effects of head patching in \Cref{sec:heads-heldout} (Spearman 0.69 and 0.68 on item means in frames A1 and B, ranges 0.44 to 0.90; a median of 8 and 7 of the ten strongest heads in common).
Exact patching of the five strongest estimated edges per receiver input was run for four of the six models with the roughest estimate on the original pairs; it agrees with the estimate at $r=0.96$ to $0.999$ over edge means in Qwen2.5-14B, 0.82 to 0.99 in Qwen2.5-7B-\ac{IT}, 0.74 to 0.93 in Gemma-2B-\ac{IT} and 0.46 to 0.58 in Qwen2.5-3B, where the estimate is not to be trusted.
In all four the strongest exact edges are the direct paths from the last \ac{MLP} blocks to the logits and, in the Qwen models, the embedding into the value inputs of heads in layer 0; the individual edges into the located heads move the score by 0.05 or less on average.
Nothing in the upstream picture distinguishes moral from normative pairs.

\paragraph{What the features respond to elsewhere.}
For the located feature sets and for the features found on the new pairs (H6), we list the strongest activating contexts in a general text corpus (WikiText-103) and in the original pairs, with each feature's activation frequency.
In the original pairs and in WikiText-103 the located moral-versus-norm sets mix a few features that fire on ``morally'' after a morally charged stem (6 of 18 in Llama-3.1-8B, 2 of 32 in Llama-3.2-1B), some evaluative features, and many features of ordinary text, one active on 10\,\% of all tokens in Llama-3.1-8B and one on 26\,\% in Llama-3.2-1B; the $k=128$ and \ac{MLP}-0 sets are mostly frequent features (median activity 6\,\% of tokens for the \ac{MLP}-0 moral-versus-norm set).
The contexts of the features found on the new pairs (H6) are listed in the released analysis output.

\subsection{\rev{Further tests of the located heads: other prompts, other pairs, other text}}
\label{sec:further}
\rev{Three properties of the main analysis could limit what it shows.
The frames A1, B and A0 all ask the model for a good-or-bad word: through the moral cue, through the two priming sentences, or through both.
Every pair was written for this study.
And the judgment is read from evaluative words.
The analyses of this subsection vary each of the three properties.
We report them as exploratory.
They cover the 14 models and use the head sets of the analyses above: the ten located heads of each model, and the 20 random sets of ten heads of the removal test.}

\paragraph{\rev{Measurement.}}
\rev{When nothing in the prompt asks for a judgment, the 28 answer words receive only 2 to 3\,\% of the probability, and the score $S$, a difference of probabilities, is then close to zero for both versions of a pair.
To compare prompts we therefore use the log-odds version of the score, \ac{LD} (\Cref{app:head-method}): the logarithm of the summed probability of the 14 positive answer words minus the logarithm of the summed probability of the 14 negative ones.
\ac{LD} does not depend on how much probability the answer words receive in total.
The gap of a pair is \ac{LD} of the better version minus \ac{LD} of the worse version; a pair is included when this gap is at least 0.1; the effect of a head and the share of the gap that a removal takes away are computed on \ac{LD} as they are computed on $S$ in the main analysis.
\vio{Two analyses of this subsection keep the score $S$ of the main analysis: the descriptive pairs about a person, shown in frames A1 and B, and the scenarios of the ETHICS benchmark.
The choice between two options uses neither score: its answer is read from the logits of the labels ``A'' and ``B''.}
\Cref{tab:further} also gives the frames A1, B and A0 on this measure, so that its rows can be compared with each other.
For this reason the numbers of \Cref{tab:further} differ somewhat from those of the main analysis, which uses $S$.}

\begin{table}[t]
\revcolor
\centering
\caption{\textbf{The located heads in other prompts (exploratory; 14 models; \vio{score \ac{LD}, except the choice and its control}).}
For moral (M), normative (N) and descriptive (D) pairs: the share of the effect that the located heads carry, as a multiple of the share of ten heads chosen at random (geometric mean over the models); the share of the gap that disappears when the located heads are removed (median over the models, in per cent; in parentheses the same for the random sets of ten heads, as the median over the models of the mean of the 20 sets); and the number of models in which the located heads remove more than every one of the 20 random sets.
The first three rows are the frames of the main analysis (priming: the two priming sentences stand before the phrase); the removal was not run in frame A0.
In the choice and in its control the answer is read from the labels A and B, head patching was not run, and 12 of the 14 models have enough included descriptive pairs in the choice.}
\label{tab:further}
\footnotesize
\setlength{\tabcolsep}{3pt}
\begin{tabularx}{\linewidth}{@{}Y>{\centering\arraybackslash}p{0.95cm}>{\centering\arraybackslash}p{0.95cm}>{\centering\arraybackslash}p{0.95cm}>{\centering\arraybackslash}p{1.3cm}>{\centering\arraybackslash}p{1.3cm}>{\centering\arraybackslash}p{1.3cm}>{\centering\arraybackslash}p{0.8cm}>{\centering\arraybackslash}p{0.8cm}>{\centering\arraybackslash}p{0.8cm}@{}}
\toprule
 & \multicolumn{3}{>{\centering\arraybackslash}p{3.2cm}}{Multiple of the random share} & \multicolumn{3}{>{\centering\arraybackslash}p{4.2cm}}{Gap removed, \% (random sets)} & \multicolumn{3}{>{\centering\arraybackslash}p{2.8cm}}{Above every random set} \\
\cmidrule(lr){2-4}\cmidrule(lr){5-7}\cmidrule(lr){8-10}
Prompt & M & N & D & M & N & D & M & N & D \\
\midrule
Priming, ``\stem{} is morally'' (A1) & 6.00 & 5.92 & 4.86 & 28 (2) & 34 (3) & 28 (2) & 11 & 11 & 8 \\
Priming, ``\stem{} is'' (B) & 5.21 & 5.31 & 4.39 & 26 (2) & 30 (2) & 23 (1) & 9 & 11 & 9 \\
``\stem{} is morally'' (A0) & 5.60 & 5.02 & 4.32 & \multicolumn{3}{c}{not run} & \multicolumn{3}{c}{not run} \\
\addlinespace
``\stem{} is socially'' & 5.25 & 4.75 & 4.24 & 37 (4) & 35 (3) & 22 (3) & 14 & 12 & 9 \\
``\stem{} is'' & 5.01 & 4.78 & 4.03 & 24 (3) & 22 (3) & 14 (2) & 13 & 12 & 8 \\
Question, then ``Answer: It is'' & 4.54 & 5.27 & 4.20 & 30 (3) & 35 (4) & 30 (4) & 11 & 12 & 8 \\
\addlinespace
Choice between the two versions & \multicolumn{3}{c}{not run} & 24 (5) & 21 (7) & 13 (5) & 6 & 7 & 3 \\
Control: which option holds a word & \multicolumn{3}{c}{not run} & 7 (5) & 6 (6) & 7 (6) & 3 & 3 & 3 \\
\bottomrule
\end{tabularx}
\end{table}

\paragraph{\rev{A prompt that asks nothing.}}
\rev{The prompt is the phrase followed by ``is'' and nothing else (``\stem{} is''): no cue, no priming sentences, no question.
The models then rarely continue with an answer word (the answer words receive a median over the models of 3\,\% of the probability on moral, 2\,\% on normative and 2\,\% on descriptive pairs), and the models still lean to the positive words for the better version: \ac{LD} is higher for the better version in a median of 97\,\%, 94\,\% and 97\,\% of the pairs, and 58 to 80 moral, 81 to 105 normative and 32 to 40 descriptive pairs per model are included.
In this prompt the located heads carry 5.01 times the share of ten random heads on moral pairs, 4.78 times on normative pairs and 4.03 times on descriptive pairs (in every model at least 2.6 times on moral and 2.5 times on normative pairs); with the two priming sentences added (frame B) the multiples are 5.21, 5.31 and 4.39.
The ratio of the share on moral pairs to the share on normative pairs is 1.05 (95\,\% \ac{CI} [0.99, 1.11]); its interval lies above 1 in two models and below 1 in none.
Removing the located heads takes away a median of 24\,\% of the gap on moral pairs and 22\,\% on normative pairs, against 3\,\% for the random sets, and more than every random set in 13 and in 12 of the 14 models (the exceptions are Llama-3.2-3B-\ac{IT} on moral pairs, and Llama-3.2-1B-\ac{IT} and Llama-3.2-3B-\ac{IT} on normative pairs).
The priming sentences, which contain the words ``good'' and ``bad'', therefore do not create the part that the located heads play: without the priming sentences, without a cue and without a question, the located heads carry nearly the same multiple of the random share and the models need them.}

\paragraph{\rev{A plain question.}}
\rev{The prompt is ``Question: Is the following good or bad?
\stem{} Answer: It is'', with the phrase on a line of its own: a request for a judgment without the moral cue and without the priming sentences.
The models answer it: the answer words receive 41 to 52\,\% of the probability and the better version is preferred in 97 to 100\,\% of the pairs (medians over the models, by kind of pair).
The located heads carry 4.54, 5.27 and 4.20 times the random share on moral, normative and descriptive pairs, and removing them takes away 30\,\%, 35\,\% and 30\,\% of the gap (random sets: 3 to 4\,\%), more than every random set in 11, 12 and 8 of the 14 models.
In this prompt the located heads carry a smaller share on moral than on normative pairs (ratio 0.86, [0.81, 0.91]; on $S$ 0.89, [0.82, 0.98]), in the direction of frame B (\HTwoRatio{}) and further from 1.}

\paragraph{\rev{Another cue word.}}
\rev{In ``\stem{} is socially'' the word ``socially'' stands where frame A0 has ``morally''.
The located heads were found with ``morally'' at the end of the prompt, and with ``socially'' they carry nearly as large a share: 5.25, 4.75 and 4.24 times the random share, against 5.60, 5.02 and 4.32 in frame A0 (on $S$ and the 10 models with enough included pairs of each kind: 5.01, 4.68 and 4.20).
Removing them takes away 37\,\% of the gap on moral and 35\,\% on normative pairs, more than every random set in 14 and in 12 of the 14 models.
The located heads therefore do not belong to the word ``morally''.
This prompt also bears on the small lean to moral pairs that the located heads show in frame A0 (ratio \HZeroRatio{}).
With ``socially'' the ratio of the share on moral pairs to the share on normative pairs is 1.08 ([1.02, 1.14], on $S$), about two thirds as far from 1, and the etiquette pairs, which the word ``socially'' suits, do not take the place of the moral pairs (etiquette against the other normative pairs: 1.08, [0.97, 1.18]; moral against etiquette: 1.03, [0.94, 1.14]).
Without any cue the ratio is 1.05 ([0.99, 1.11], above), and with the priming sentences and the moral cue it is \HOneRatio{}.
A small lean to moral pairs thus appears whenever a cue word stands alone, about one and a half times as large with the moral cue as with ``socially''.}
 \grn{The synonym ``ethically'' was run in the same way, with the priming sentences and without them (``\stem{} is ethically''; score $S$). The located heads carry as large a share as with ``morally'': 5.50 and 5.25 times the random share on moral and on normative pairs with the priming sentences (5.44 and 5.47 with ``morally''), and 6.16 and 5.30 without them (5.44 and 4.89; the 13 and 11 models with enough included pairs of each kind). The ratio of the share on moral pairs to the share on normative pairs stays inside the margin: 1.06 ([1.00, 1.11]) with the priming sentences and 1.14 ([1.08, 1.21]) without them, against \HOneRatio{} and \HZeroRatio{} with ``morally''. In one model, Llama-3.2-1B-Instruct, the interval lies above the margin in the frame without the priming sentences (1.38, [1.26, 1.51]). The small lean to moral pairs under a moral cue is therefore not tied to the word ``morally''.}

\paragraph{\rev{Scenarios that we did not write.}}
\rev{The ETHICS benchmark \citep{hendrycks2021aligning} holds short scenarios in the first person, each labelled acceptable or unacceptable (``I trained my enormous dog to shake hands with people''; ``I smothered the crying baby in my arms'').
From the test split of its commonsense part we drew 400 scenarios, 200 of each label, among those that are one sentence of 6 to 20 words and contain no answer word of the score.
A scenario has no second version, so head patching is not possible, and we measure the removal.
Each scenario is shown in three prompts: the priming sentences, the scenario and ``This is morally''; the same without ``morally''; and the question prompt described above.
The separation of a model in a prompt is the mean score $S$ of the 200 acceptable scenarios minus the mean score of the 200 unacceptable ones.
Every model separates the two classes in every prompt (the lowest \ac{AUC} is 0.73; the medians over the models are 0.90, 0.89 and 0.93 in the three prompts).
Removing the located heads takes away a median of 39\,\%, 39\,\% and 37\,\% of the separation, against 1\,\%, 2\,\% and 3\,\% for the random sets, and more than every one of the 20 random sets in 13, 10 and 10 of the 14 models.
The order of the scenarios changes less than the size of the separation: the median \ac{AUC} falls from 0.90 to 0.86, from 0.89 to 0.84 and from 0.93 to 0.87.
In three cases the removal does not reduce the separation: Gemma-2B-\ac{IT} and Llama-3.2-3B-\ac{IT} in the question prompt, and Qwen2.5-14B-\ac{IT} in the prompt without ``morally''.
The models thus need the located heads to judge sentences that were not written for this study.
All scenarios are moral, so this analysis says nothing about specificity.}

\paragraph{\rev{Descriptive pairs about a person.}}
\rev{The located heads carry about a fifth more of the effect on normative than on descriptive pairs (H3, \Cref{sec:heads-heldout}).
But the 40 descriptive pairs differ from the normative pairs in two ways at once: nobody's conduct is judged in them, and they are noun phrases about a thing, with no person and in one sentence template.
To separate the two we wrote 66 further descriptive pairs in which a person is present and nobody's conduct is judged: something good or bad happens to the person (``Winning the raffle at the village fair'' against ``Losing the raffle at the village fair''; ``Getting through the winter without a cold'' against ``... with a cold'').
These pairs use the four sentence templates of the moral and normative pairs and passed the same mechanical checks.
The three language models of \Cref{sec:followup} answered the annotators' form for them, and a pair was kept if at least two of the three models chose the intended better version, at least two saw nobody's conduct judged, the median moral rating was at most 1 and fewer than half of the answers were ``unclear''; 58 pairs were kept.
Head patching was run in frames A1 and B, and this analysis uses $S$ and the inclusion rule of the main analysis.
Without the cue (frame B), the located heads carry the same share on the 58 pairs about a person as on the normative pairs (ratio of normative to person 1.00, 95\,\% \ac{CI} [0.92, 1.09]) and 1.22 times their share on the 40 descriptive pairs about a thing ([1.11, 1.32]); as multiples of the random share, moral 4.36, normative 4.86, person 4.88, thing 4.01.
With the cue (frame A1), the pairs about a person lie halfway: normative against person 1.11 ([1.04, 1.19]), person against thing 1.08 ([1.01, 1.15]); multiples 5.44, 5.47, 4.97 and 4.59.
With all 66 pairs the two ratios are 1.02 and 1.19 without the cue, and 1.13 and 1.06 with it.
Without the cue, then, the difference of a fifth lies between phrases about a person and phrases about a thing, and it does not require that conduct is judged; with the cue about half of it remains between the normative pairs and the pairs about a person.
The labels of these 58 pairs were checked by language models, which pass a test against our annotators on the rated pairs but are more lenient than the annotators (\Cref{sec:followup}), and not by the annotators themselves.}

\paragraph{\rev{A choice between the two versions, without evaluative words.}}
\rev{In all prompts above the judgment is read from words such as ``good'' and ``wrong''.
In the last prompt neither the prompt nor the answer holds an evaluative word.
The two versions of a pair are shown as two options and the model is asked to choose (``Option A: \emph{one version}.
Option B: \emph{the other version}.
I choose Option''), and the answer is read from the logits of the labels ``A'' and ``B''.
Each pair gives two prompts, one with the better version as option A and one with the better version as option B.
The gap of the pair is the logit of A minus the logit of B in the first prompt, minus the same difference in the second prompt, so that a habit of answering with one label cancels.
The models choose the better version: the gap is positive for a median of 90\,\% of the 227 pairs (74 to 99\,\% per model).
Removing the located heads takes away a median of 24\,\% of the gap on moral pairs and 21\,\% on normative pairs, three to four times what the random sets take away (5\,\% and 7\,\%).
But the models differ.
The located heads remove more than every random set in 6 of the 14 models on moral pairs (Gemma-2B, Qwen2.5-3B, the two Llama-3.1-8B models and the two Qwen2.5-14B models) and in 7 on normative pairs (the same six and Llama-3.2-3B), and in Llama-3.2-1B-\ac{IT} and Qwen2.5-7B-\ac{IT} the gap grows when the located heads are removed.
A control tells a need that is tied to evaluation from a need that any choice between two listed options would show.
The control has the same options and the same labels, and its question requires no evaluation: which option holds a given word (the prompt ends in ``The word \emph{w} is in Option'', where \emph{w}, in quotation marks, is the first word that only the worse version contains, so that the right answer is the option that the model does not prefer).
The models give the right answer in both orders for a median of 99\,\% of the pairs.
In the control, the located heads take away about what the random sets take away (7\,\% against 5\,\% of the gap on moral pairs, 6\,\% against 6\,\% on normative pairs), and they remove more than every random set in three models, Gemma-2B, Qwen2.5-3B and Qwen2.5-14B.
In Gemma-2B and Qwen2.5-3B the removal of the located heads takes away the whole gap in the choice and in the control alike, so their result in the choice does not show a need that is tied to evaluation; in Qwen2.5-14B it takes away 24\,\% of the gap on moral pairs in the choice and 11\,\% in the control.
In ten of the other twelve models (all except Llama-3.2-1B-\ac{IT} and Qwen2.5-7B-\ac{IT}) the located heads take away more of the gap on moral pairs in the choice than in the control (for example 44\,\% against 15\,\% in Llama-3.1-8B); this comparison is descriptive.
By the criterion of this test, the located heads would count as needed for the choice if they removed more than every random set in more than half of the models and their median share of the gap was at least three times that of the random sets, and as not needed if they did so in fewer than a quarter of the models or their median share was below twice that of the random sets.
The result lies between the two: in a choice between two options, removing the located heads costs the models three to four times as much as removing random heads (medians), but the located heads exceed every random set of ten heads in 6 and 7 of the 14 models, not in more than half.
The results of this paper about the located heads therefore hold for judgments that are read from evaluative words, and hold in part for a choice.}

\paragraph{\rev{What these tests add.}}
\rev{In the 14 models, the located heads carry about the same multiple of the random share when the prompt asks nothing, when it asks a plain question and when the cue is another word, and the models need the located heads in these three prompts and on scenarios that we did not write.
Without the cue, the difference between normative and descriptive pairs is a difference between phrases about a person and phrases about a thing; with the cue, half of it is.
One limit remains: when the judgment is read from a choice between two options and not from evaluative words, the located heads are needed in some models only.}

\section{The features on the new pairs, in detail}
\label{app:features}

\subsection{The located features, tested on new pairs}
\label{sec:features-heldout}
This section uses the new pairs retained by the annotators, at the final token; A1 is the primary frame.
The feature sets are those located on the original pairs: for each dictionary, the moral-versus-neutral set \rev{(the located features)} and the smaller moral-versus-norm set (\Cref{sec:features-dev}).

\paragraph{Capture check.}
The dictionaries see little of the pair difference.
Every one of the 14 \rev{dictionaries} is flagged in both frames (\Cref{tab:features-heldout}).
In frame A1 the mean $R^2$ is below zero in 13 \rev{dictionaries} (from $-1.54$ for the \ac{MLP}-0 dictionary of Llama-3.2-1B to $-0.02$) and 0.38 in the layer-19 dictionary of Llama-3.2-3B-\ac{IT}, with cosines between $d$ and $\hat d$ of 0.2 to 0.7; in frame B it is 0.10 to 0.49.
The reconstruction of each \rev{hidden} state is fair (relative error 0.04 to 0.47), but the difference between the two versions of a pair is a small part of the \rev{hidden} state, and the dictionaries reproduce it poorly.
On the original pairs all 14 \rev{dictionaries} are below 0.5 in the original frame, and in a rehearsal on 18 of those pairs in the new frames all were flagged in A1 while four reached 0.5 in B, so the flags were expected.
The feature results below therefore describe what the dictionaries capture, not what the models lack.

\paragraph{H5: do the located features follow the frame?}
The test is the swap of frames of \Cref{sec:features-dev}, now on new pairs of every category.
If the \rev{located features} respond to the word ``morally'', they should fire on descriptive pairs almost as much as on moral pairs when the word is present, and fall silent for every category when it is absent; if they respond to moral content, they should fire on moral pairs in both frames and little on descriptive pairs.
For each dictionary's moral-versus-neutral set, H5 therefore predicts that the mean activation on descriptive pairs in frame A1 is more than half of that on moral pairs in frame A1, and near zero in frame B for every category.
The endpoint is the ratio $A(\mathrm{A1},\text{descr})/A(\mathrm{A1},\text{moral})$ per \rev{dictionary}: the set follows the frame if its interval lies above 0.5, and the content if it lies below.
The ratio is above 0.5 in all \HFiveFrameDriven{} \rev{dictionaries}, from 0.70 [0.68, 0.72] in Llama-3.2-1B to 0.99 in its \ac{MLP}-0 dictionary and 0.98 in the two Qwen2.5-7B dictionaries (\Cref{tab:features-heldout}), and the same sets are nearly silent in frame B: on moral pairs their mean activation without the cue is 1 to 33\,\% of its value with the cue.
These features light up for ``The harvest after a season of steady rain is morally'' almost as much as for a moral pair, and go dark for a moral pair without the word.

The smaller moral-versus-norm sets (18 to 124 features per \rev{dictionary}) behave differently: their ratio of descriptive to moral activation in A1 is 0.09 to 0.40 in 11 \rev{dictionaries} (they follow the content), 0.71 and 0.72 in the \ac{MLP}-0 and layer-19 dictionaries, and 0.37 [0.27, 0.50] in the layer-3 dictionary (inconclusive), and their ratio of norm to moral activation is 0.2 to 0.55 in the same 12 \rev{dictionaries}: on new pairs they respond about twice as much to moral as to normative content, which is what they were selected for, within dictionaries that capture little of the difference.
So the \rev{located features} (moral against neutral) switch on whenever the prompt ends in ``morally'', whatever the pair is about, and switch off when it does not; only the smaller sets chosen against norm pairs prefer moral content (other wordings of the cue are in \Cref{sec:features-cues}).

\begin{table}[t]
\centering
\caption{\textbf{Feature results on the new pairs, per \rev{dictionary}}: capture of the pair difference at the final token (mean $R^2$ over categories; * = below 0.5), the H5 ratio with its 95\,\% interval (f = follows the frame, c = follows the content, i = inconclusive), and the H6 counts of features that separate the groups, moral versus norm before the bar and the three splits among norms after it, all on groups equalised to \HSixGroupSize{} pairs; bold where moral versus norm exceeds every split.}
\label{tab:features-heldout}
\scriptsize
\setlength{\tabcolsep}{3pt}
\begin{tabular}{llccccc}
\toprule
& & \multicolumn{2}{c}{Capture $R^2$} & H5: descr./moral & \multicolumn{2}{c}{H6 counts: moral vs norm $\mid$ the three splits among norms} \\
\cmidrule(lr){3-4}\cmidrule(lr){6-7}
Model & Site / $k$ & A1 & B & A1 [95\,\% \ac{CI}] & A1 & B \\
\midrule
Llama-3.2-1B & R7 / 64 & -0.23* & 0.11* & 0.70 [0.68, 0.72] f & \textbf{14.3} $\mid$ 0.3, 4.0, 6.0 & 6.2 $\mid$ 8.8, 2.3, 9.7 \\
Llama-3.2-1B & R7 / 128 & -0.18* & 0.21* & 0.73 [0.70, 0.75] f & \textbf{23.4} $\mid$ 3.0, 11.6, 6.0 & 16.4 $\mid$ 8.2, 13.0, 18.6 \\
Llama-3.2-1B-\ac{IT} & R7 / 64 & -0.25* & 0.14* & 0.79 [0.77, 0.81] f & \textbf{15.0} $\mid$ 0.0, 0.1, 0.7 & \textbf{14.3} $\mid$ 6.8, 3.3, 13.3 \\
Llama-3.2-1B-\ac{IT} & R7 / 128 & -0.21* & 0.24* & 0.78 [0.76, 0.80] f & \textbf{30.9} $\mid$ 1.5, 6.8, 5.8 & 16.9 $\mid$ 8.7, 15.6, 20.0 \\
Llama-3.2-1B & \ac{MLP}0 / 64 & -1.54* & 0.37* & 0.99 [0.99, 1.00] f & 0.2 $\mid$ 0.0, 0.9, 0.5 & 0.1 $\mid$ 0.7, 0.1, 0.8 \\
Llama-3.2-3B & R11 / 64 & -0.11* & 0.21* & 0.75 [0.73, 0.78] f & \textbf{21.9} $\mid$ 4.7, 5.5, 9.2 & 17.6 $\mid$ 2.6, 3.9, 17.7 \\
Llama-3.2-3B & R11 / 128 & -0.02* & 0.31* & 0.73 [0.71, 0.75] f & \textbf{52.4} $\mid$ 10.8, 20.6, 19.4 & 24.6 $\mid$ 11.6, 15.3, 34.0 \\
Llama-3.2-3B-\ac{IT} & R11 / 128 & -0.04* & 0.33* & 0.81 [0.79, 0.83] f & \textbf{59.5} $\mid$ 4.0, 4.0, 19.4 & 32.0 $\mid$ 15.0, 1.9, 49.1 \\
Llama-3.2-3B-\ac{IT} & R3 / 128 & -0.04* & 0.23* & 0.86 [0.84, 0.87] f & 0.9 $\mid$ 3.4, 0.0, 2.4 & \textbf{4.4} $\mid$ 3.4, 1.8, 0.2 \\
Llama-3.2-3B-\ac{IT} & R7 / 128 & -0.17* & 0.27* & 0.77 [0.75, 0.79] f & \textbf{15.1} $\mid$ 0.3, 3.5, 13.3 & 19.8 $\mid$ 4.9, 15.8, 27.4 \\
Llama-3.2-3B-\ac{IT} & R19 / 128 & 0.38* & 0.49* & 0.92 [0.90, 0.94] f & \textbf{54.9} $\mid$ 10.3, 14.2, 39.2 & 77.4 $\mid$ 44.4, 53.4, 81.8 \\
Llama-3.1-8B & R12 / JumpReLU & -0.09* & 0.10* & 0.86 [0.84, 0.88] f & \textbf{7.3} $\mid$ 2.5, 1.0, 6.5 & 10.7 $\mid$ 3.0, 9.1, 18.1 \\
Qwen2.5-7B & R0 / 128 & -0.02* & 0.38* & 0.98 [0.97, 0.99] f & \textbf{0.2} $\mid$ 0.1, 0.1, 0.0 & 0.1 $\mid$ 0.0, 0.0, 0.3 \\
Qwen2.5-7B-\ac{IT} & R0 / 128 & -0.06* & 0.39* & 0.98 [0.98, 0.99] f & \textbf{0.3} $\mid$ 0.1, 0.0, 0.0 & \textbf{0.2} $\mid$ 0.1, 0.1, 0.0 \\
\bottomrule
\end{tabular}

\end{table}

\subsection{Features found on the new pairs}
\label{sec:features-discovery}
The new pairs also allow a search for features that separate moral from normative content.
Such a search needs a yardstick: any two groups of pairs differ in something, so some features will separate them.
The yardstick here is the norm pairs themselves.
If moral pairs are just one more kind of normative pair, then telling moral from normative pairs should need about as many features as telling one pair of norm kinds from another, say prudential and etiquette from aesthetic and epistemic.
If moral content stands apart, the moral-versus-norm comparison should yield clearly more.

\paragraph{H6: no more moral features than features for kinds of norm.}
For each feature active on at least five prompts of the moral and norm pairs, we test the moral-versus-norm difference in mean change $a_{ij}$ with a two-sided permutation test, Benjamini-Hochberg at 0.05 over features, using 20{,}000 permutations or 20 per tested feature, whichever is larger, so that the correction can detect even a single feature.
Calibration: the same test for the three splits of the four kinds of norm into two against two (for example prudential and etiquette against aesthetic and epistemic), with moral and norm items subsampled to the same group sizes (20 subsamples).
The endpoint is the count for moral versus norm next to the counts for norm versus norm.
If no feature is detected in any comparison, no directional reading is given.
The logic is that of a control experiment: the splits among norms show how many features any two groups of this size and form differ in, so that a count for moral versus norm means something only when it exceeds them.

In frame A1, moral versus norm yields more features than any of the three splits among norms in \HSixMoreThanNorm{} of the \HSixConditions{} \rev{dictionaries} (\Cref{tab:features-heldout}).
In the ten dictionaries that detect features at all, this means 7 to 60 features at equalised group sizes of 46 pairs against at most 1 to 39 for the best split among norms (29 to 165 features at full size); the two Qwen2.5-7B dictionaries, read at the first layer, detect almost nothing in either frame (at most 4 features at full size) and count only nominally.
The two exceptions are also dictionaries with almost no detected features, the \ac{MLP}-0 dictionary of Llama-3.2-1B (0.2 against 0.9) and the layer-3 dictionary of Llama-3.2-3B-\ac{IT} (0.9 against 3.4).

In frame B it is the other way round: moral versus norm yields more features than every split among norms in only \HSixMoreThanNormB{} \rev{dictionaries}, and in the other 11 one of the splits, most often prudential and epistemic against etiquette and aesthetic, yields as many or more (for example 10.7 against 18.1 in Llama-3.1-8B, 77.4 against 81.8 in the layer-19 dictionary).
So the features at the final token distinguish moral from normative pairs more sharply than they distinguish kinds of norm only when the prompt \rev{carries the moral cue}; without the cue, the distinction between moral and normative pairs is one normative distinction among others.

\paragraph{Ablation (secondary).}
Removing \rev{the two feature sets} tests whether they matter for the model's preference, and for which pairs.
Each \rev{set} is removed at the final token with the intervention that keeps the reconstruction error, against 20 random sets of the same size and similar activity.
The endpoint is the change of the behavioural gap among included items, set minus random, by category and frame, with the norm of the removed vector reported for the set and for the random sets.

The moral-versus-neutral sets change the behavioural gap when removed in frame A1, and they change it for every category: in Llama-3.1-8B the gap falls by 0.27 more than under the random sets on moral pairs, by 0.19 on norm pairs and by 0.10 on descriptive pairs; in the Llama-3.2-3B dictionaries by 0.15 to 0.32 on moral and 0.03 to 0.18 on norm pairs; in the \ac{MLP}-0, layer-3, layer-7 and layer-19 dictionaries and in both Qwen2.5-7B dictionaries the gap grows instead (by 0.12 to 0.47), mostly for moral and norm pairs alike.
The norm of the removed vector is 1.5 to 2.2 times that of the random sets in these \rev{dictionaries}, so the random sets remove less and the sizes are not to be read closely.
In frame B the effects of these sets are within 0.05 of zero in 13 \rev{dictionaries}; the layer-19 dictionary of Llama-3.2-3B-\ac{IT}, whose set is large and active in both frames, loses 0.18 to 0.21 of the gap for every category.
The moral-versus-norm sets have no measurable effect in either frame (all within 0.05, except $-0.11$ to $-0.22$ in frame B for the same layer-19 dictionary).
The \rev{located features} are, causally as well as descriptively, features of the moral frame.

\paragraph{What the new features are, and what removing them does (exploratory).}
These analyses ask whether the features of H6 are moral features in any causal sense.
First we looked them up in ordinary text (WikiText-103) and in the original pairs.
They are a mixture: features of moral vocabulary in the stems (stealing, torture, abuse, fraud, a stranger, a child, profit, rescue), a few features for abstract moral notions (justifying and excusing, deeds and evil, injustice and corruption, and features that fire on the word ``morally'' in text yet were selected on moral content in the frame without the word), and many generic features of function words and punctuation that happen to differ between the two groups of pairs.
Removing the full H6 lists at the final token, keeping the reconstruction error and comparing with random sets of similar activity, reduces the moral gap more than the normative gap in several of the small Llama dictionaries, mostly without the cue (for example by 0.10 against 0.01 in Llama-3.2-1B-\ac{IT}); but these lists were selected on the same pairs on which the effect is measured, so part of it is built in.

We therefore repeated the test with the pairs split in two halves: features selected on one half, removed on the other, in both directions, so that every pair is read with features chosen without it.
The held-out effect is about half as large and still there.
Without the cue, in 4 of 10 dictionaries the features cut the moral gap by 0.01 to 0.06, that is by 5 to 10\,\% of it, and the normative gap by at most 0.02: Llama-3.2-1B-\ac{IT} $-0.054$ [$-0.073$, $-0.036$] against $-0.000$, its $k=128$ dictionary $-0.062$ against $-0.012$, Llama-3.2-3B base $-0.032$ against $-0.017$, Llama-3.2-1B base $k=128$ $-0.013$ against $-0.001$, with descriptive pairs unaffected throughout.
With the cue, the same holds in 2 of 10.
The Llama-3.1-8B dictionary and the layer-3, 7 and 19 dictionaries of Llama-3.2-3B-\ac{IT} show no effect specific to moral pairs; with the cue, the layer-19 list removes a fifth of the gap for every category alike.
So in the small models the \rev{hidden} state at the final token carries a component specific to moral pairs.
Its causal weight is small, it shows on pairs on which it was not selected, and it does not need the word ``morally''.
Whether that component is more than the valence of moral vocabulary, the list above leaves open.

\subsection{Other wordings and positions of the cue (exploratory)}
\label{sec:features-cues}
The moral cue does not suit every pair equally.
With ``is morally'' the models put more probability on the answer words, and separate the two versions of a pair more, for moral than for normative pairs, and least for descriptive pairs (medians over the 14 models: probability of the answer words 0.75, 0.58 and 0.43; behavioural gap 1.00, 0.69 and 0.37); without the cue the three categories differ much less (0.59, 0.56 and 0.48; 0.71, 0.68 and 0.64).
With the cue every model still favours the better version in at least 89\,\% of the pairs of each category, but ``with the cue'' also means ``with a \rev{word} that suits the moral pairs best''.
To see what the features of \Cref{sec:features-heldout,sec:features-discovery} respond to, we ran further frames on the same pairs and \rev{dictionaries}: a synonym (``\stem{} is ethically''), the moral cue placed before the stem (``Morally speaking, \stem{} is'', so that the prompt ends in ``is'', as without the cue), and two cues that are not moral, ``\stem{} is logically'', meant to suit the epistemic pairs, and ``\stem{} is socially'', meant to suit the etiquette pairs.
In the Llama and Qwen tokenizers ``ethically'' is two tokens, so its features are read at the second one, ``ically'' (\Cref{app:feature-method}).

\begin{table}[!htb]
\centering
\caption{\textbf{What the features follow (exploratory).}
For each frame, the activation of the \rev{located features} on moral pairs relative to their activation with ``is morally'' (median and range over the ten \rev{dictionaries} at layer 7 or deeper), and the number of the 14 \rev{dictionaries} in which more features separate moral from normative pairs than separate kinds of norm (H6).
In the answers, the gap of the moral pairs relative to that of the normative pairs is larger than without a cue with ``is morally'' (in all 7 models of these \rev{dictionaries}), with ``Morally speaking'' (in 5 of 7) and with ``is socially'' (in 6 of 7); the gap of the etiquette pairs relative to that of the prudential, aesthetic and epistemic pairs is larger with ``is socially'' in all 7, and the gap of the epistemic pairs relative to that of the other normative pairs is larger with ``is logically'' in none of the 7.}
\label{tab:cues}
\small
\setlength{\tabcolsep}{5pt}
\begin{tabular}{llcc}
\toprule
Frame & End of the prompt & \rev{Located features} & Moral pairs stand apart (H6) \\
\midrule
Moral cue (A1) & \stem{} is morally & 1 & 12 of 14 \\
Synonym & \stem{} is ethically & 0.79 (0.64 to 0.90) & 7 of 14 \\
Moral cue before the stem & Morally speaking, \stem{} is & 0.18 (0.10 to 0.49) & 9 of 14 \\
Social cue & \stem{} is socially & 0.50 (0.33 to 0.54) & 2 of 14 \\
Logical cue & \stem{} is logically & 0.31 (0.18 to 0.47) & 0 of 14 \\
No cue (B) & \stem{} is & 0.16 (0.03 to 0.33) & 3 of 14 \\
\bottomrule
\end{tabular}
\end{table}

The \rev{located features} respond to ``ethically'' at about four fifths of their level with ``morally'' (\Cref{tab:cues} and \Cref{fig:cue-frames}): on moral pairs they reach 0.64 to 0.90 of their activation with ``morally'' in the ten dictionaries at layer 7 or deeper, and 0.77 at layer 3 of Llama-3.2-3B-\ac{IT}.
With ``ethically'' they again fire on descriptive pairs as well, at 0.69 to 0.98 of their level on moral pairs.
Only the three dictionaries at the first layer, where the \rev{hidden} state at the final position mostly reflects the token itself, stay close to their level without a cue (0.08 to 0.11, against 0.01 to 0.06).
With ``socially'' the same sets reach about half of their activation with ``morally'' (median 0.50, 0.33 to 0.54), with ``logically'' about a third (median 0.31), and with ``Morally speaking'' at the start of the sentence hardly more than without a cue (median 0.18, against 0.16).
So these features respond to the word at the position where they are read, most of all to a moral word in either wording, not to moral content, and hardly to a moral \rev{cue} placed earlier in the prompt.
The smaller sets chosen against norm pairs keep their preference for moral content with ``ethically'': their activation on descriptive pairs is 0.16 to 0.35 of that on moral pairs in 8 of the 10 dictionaries at layer 7 or deeper.

The distinction between moral and normative pairs found on the new pairs (H6) follows the \rev{moral cue, wherever it stands,} instead.
More features separate moral from normative pairs than separate kinds of norm in 9 of the 14 dictionaries when ``Morally speaking'' opens the sentence, in 7 with ``ethically'', in 2 with ``socially'' and in none with ``logically'', against 12 with ``morally'' and 3 without a cue.
So the \rev{dictionaries} mark moral pairs as a class of their own when the prompt \rev{carries a moral cue, whether the cue} ends the prompt or opens the sentence (somewhat less often with the synonym), and hardly under a cue of another kind.
Whether a moral \rev{cue} does this because it suits the moral pairs or because it brings out something specific to moral content, the first three frames cannot decide.
The planned test needed a non-moral cue that suits another kind of norm, and ``logically'' did not: in none of the 7 models did it make the answers separate the epistemic pairs more, relative to the other normative pairs, than without a cue, so that comparison was not read.

We therefore added ``\stem{} is socially'', which differs from the prompt with ``morally'' only in its last word and goes well with the answer words for the etiquette pairs (``socially acceptable'', ``socially inappropriate'').
In the answers it suits them: relative to that of the prudential, aesthetic and epistemic pairs, their gap is larger than without a cue in all 7 models (median ratio 1.64, against 0.77 without a cue).
It also suits the moral pairs about as well as ``is morally'' does: relative to that of the normative pairs, their gap is larger than without a cue in 6 of the 7 models (median ratio 1.44, against 1.33 with ``morally'' and 1.07 without a cue).
The planned comparison set the moral pairs and the etiquette pairs, each against the prudential, aesthetic and epistemic pairs, and was to be read only if the moral pairs stood apart in at least four more dictionaries with ``morally'' than without a cue.
They stood apart in 3 against none, mostly because among the calibration comparisons the epistemic pairs separate from the other kinds in many features, so by the rule the comparison is not informative.

What the counts show can still be described.
With ``socially'' the etiquette pairs stand apart in 5 of the 14 dictionaries (four of them in Llama-3.2-3B; the fifth, the \ac{MLP}-0 dictionary, detects almost nothing), and in none with ``morally'', with ``logically'' or without a cue.
Compared with the frame without a cue, a secondary measure, the logarithm of a group's count over the largest calibration count (each plus one), rises more for the etiquette pairs with ``socially'' than for the moral pairs with ``morally'' (medians 0.75 and 0.40).
A \rev{cue} that is not moral but suits one kind of norm can therefore set that kind apart in the features as well, as the fit of the \rev{cue} would predict.
The H6 count under ``socially'' (2 of 14) does not bear on this, because ``socially'' also suits the etiquette pairs, which are part of the normative pairs.
These observations were not a planned test and rest on small counts, but they suggest that part of the separation of the moral pairs under a moral \rev{cue} may come from how well the \rev{cue} suits them.

\section{\rev{Where moral pairs differ from the other pairs, in detail}}
\label{sec:exploratory}
The located heads treat every standard alike.
The analyses of this section ask whether that is because the models cannot tell the standards apart, and where, if anywhere, moral content leaves a mark of its own.
They are exploratory: a frame that asks the models which standard a pair is about, an analysis of single neurons, a reading of the final position through a Jacobian lens, the addition of directions to the \rev{hidden state}, and three sets of new pairs that fix the edited word or the scenario, annotated by language models that were first checked against our annotators.
Four further cue frames for the features are reported with the feature study (\Cref{sec:features-cues}), and an observation on a late-layer dip of the emerging judgment, made on the original pairs and examined again on the new ones, is in \Cref{app:dip}.

\subsection{Can the models tell which standard applies?}
\label{sec:category-frame}
The frames of the study ask ``good or bad?'', a question on which morality and manners agree, and the located heads answer it the same way for every category.
A different question is whether the models can tell \emph{which} standard a pair is about at all.
We asked it with a frame of its own, ``\stem{} is a matter of'', reading the next-token probabilities of fixed category words (morality or ethics; prudence or safety; manners or politeness; taste or beauty; evidence or logic; and fact or chance for the descriptive pairs), once with and once without five fixed example sentences that give the model the label vocabulary (one per standard, none a study pair).
The frame was chosen on 36 original pairs and then run on both versions of every accepted pair; the annotators' standard is the reference.
All 14 models can (\Cref{tab:category-frame}).

With the examples, the probability of the moral words separates moral from normative pairs with an \ac{AUC} of 0.94 to 0.99, and the most probable of the five standards is the annotators' standard for 81 to 92\,\% of the prompts (median 0.88; chance 0.20), from Llama-3.2-1B to the 14B models; without the examples the accuracy is 0.54 to 0.76 and the \ac{AUC} 0.88 to 0.96.
Pooled over models, moral pairs are called moral 88\,\% of the time, prudential pairs prudential 88\,\%, etiquette 84\,\%, aesthetic 92\,\% and epistemic 81\,\%, and the confusions are the natural ones: etiquette with aesthetic, epistemic with moral, moral with prudential.
So the information that a pair is a moral rather than a prudential or an aesthetic matter is present in every model, and one word of the prompt is enough to bring it out, while the components that produce the good-or-bad judgment (the located heads of \Cref{sec:heads-heldout}) treat all standards alike.
Whether the information is there when nobody asks for it, and whether it reaches the judgment at all, is the subject of \Cref{sec:jlens}.

\begin{table}[t]
\centering
\caption{\textbf{Which standard is this pair about?}
For each model, the frame ``\stem{} is a matter of'' read on fixed category words, on both versions of the accepted pairs whose standard the annotators confirmed (159 pairs, 318 prompts): \ac{AUC} of the moral words for moral against normative pairs; accuracy of the most probable of the five standards against the annotators' standard (chance 0.20); and the share of the next-token distribution that the category words cover.
With five fixed example sentences and without.}
\label{tab:category-frame}
\small
\setlength{\tabcolsep}{5pt}
\begin{tabular}{lcccccc}
\toprule
& \multicolumn{3}{c}{With five example sentences} & \multicolumn{3}{c}{Without examples} \\
\cmidrule(lr){2-4}\cmidrule(lr){5-7}
Model & \ac{AUC} & accuracy & covered & \ac{AUC} & accuracy & covered \\
\midrule
Llama-3.2-1B & 0.97 & 0.84 & 9\,\% & 0.93 & 0.54 & 6\,\% \\
Llama-3.2-1B-\ac{IT} & 0.95 & 0.85 & 8\,\% & 0.91 & 0.66 & 3\,\% \\
Llama-3.2-3B & 0.98 & 0.90 & 9\,\% & 0.91 & 0.60 & 5\,\% \\
Llama-3.2-3B-\ac{IT} & 0.97 & 0.87 & 9\,\% & 0.91 & 0.71 & 3\,\% \\
Gemma-2B & 0.96 & 0.85 & 9\,\% & 0.90 & 0.61 & 5\,\% \\
Gemma-2B-\ac{IT} & 0.94 & 0.81 & 8\,\% & 0.89 & 0.76 & 4\,\% \\
Qwen2.5-3B & 0.98 & 0.92 & 11\,\% & 0.88 & 0.63 & 3\,\% \\
Qwen2.5-3B-\ac{IT} & 0.98 & 0.88 & 13\,\% & 0.94 & 0.74 & 5\,\% \\
Qwen2.5-7B & 0.96 & 0.86 & 9\,\% & 0.94 & 0.61 & 4\,\% \\
Qwen2.5-7B-\ac{IT} & 0.96 & 0.87 & 9\,\% & 0.95 & 0.65 & 3\,\% \\
Llama-3.1-8B & 0.98 & 0.89 & 11\,\% & 0.91 & 0.69 & 5\,\% \\
Llama-3.1-8B-\ac{IT} & 0.98 & 0.89 & 11\,\% & 0.90 & 0.72 & 3\,\% \\
Qwen2.5-14B & 0.98 & 0.89 & 9\,\% & 0.95 & 0.66 & 4\,\% \\
Qwen2.5-14B-\ac{IT} & 0.98 & 0.89 & 8\,\% & 0.96 & 0.71 & 3\,\% \\
\midrule
Median & 0.98 & 0.88 & & 0.91 & 0.66 & \\
\bottomrule
\end{tabular}

\end{table}

\subsection{Neurons: a small moral residue and \rev{neurons shared by every judgment}}
\label{sec:neurons}
The components above are attention heads and dictionary features.
We also looked at the smallest units a transformer offers without a dictionary, the individual \ac{MLP} neurons (the inputs of each layer's down-projection at the final token, 8\,000 to 19\,000 per layer), with the question of H6 asked of each layer: how many neurons respond to the edit differently for moral than for normative pairs, against the three splits of the four kinds of norm into two against two.
Two corrections were needed, and neither made the counts informative; the reason is worth stating, because it applies to any count of this kind.
Moral pairs move the models more than normative pairs do (with the cue by 13 to 91\,\%), so on raw responses every neuron that carries the judgment at all looks moral; we therefore used each neuron's share of its layer's total response, and separately compared moral pairs with normative pairs whose behavioural gap has the same size.
On both measures tens of thousands of neurons per model still separate moral from normative pairs, and they exceed every split among norms in most layers of most models.

Two comparisons within the moral pairs, at one group size (32 pairs a side, counts summed over layers), show what such counts can and cannot say.
Two random halves of the moral pairs differ in almost no neurons (at most 2 in 25 of the 28 combinations of model and frame; 900 to 1\,100 in three Qwen models without the cue), and the moral pairs about care differ from the other moral pairs in 13 to 1\,900.
Moral and normative pairs differ in 180 to 47\,000, more than care and the other moral pairs in every model and frame (median ratio 38 with the cue and 18 without).
But the splits between kinds of norm, whose two sides differ in standard as well as in topic, do not consistently exceed the difference of topic within morality (in 9 of the 28 cases one of them falls below it), so counts at this resolution do not tell a difference of standard from a difference of topic.
We therefore read them as a limit of detection rather than as a finding; what counts is the causal test on held-out pairs.

For that test we kept, per layer, the 25 neurons with the strongest moral-versus-normative difference in share, chosen on half of the pairs (200 to 900 neurons per model, 0.1 to 0.2\,\% of its \ac{MLP} neurons), replaced their activity by its mean at the final token on the other half of the pairs, against ten random sets with similar activity, and read the change of the gap relative to each pair's own gap (\Cref{tab:neurons}).
Without the cue, the moral gap falls and the normative and descriptive gaps do not in four models, all of the Llama family: Llama-3.2-1B (moral $-9$\,\%, normative $-1$\,\%, descriptive $+2$\,\%), Llama-3.2-3B ($-8$, $-2$, $+3$), Llama-3.1-8B ($-5$, $0$, $+1$) and Llama-3.1-8B-Instruct ($-9$, $+1$, $+1$).
In Llama-3.2-1B-Instruct and Qwen2.5-3B the fall is general, and in the Qwen 7B and 14B models and in Gemma-2B there is none, or the moral gap grows.
With the cue only Llama-3.1-8B-Instruct shows a specific fall ($-3$\,\% against $0$).

This is the pattern of the features (\Cref{sec:features-discovery}) at the level of neurons: a small causal contribution specific to moral pairs, strongest without the word ``morally'', in one family, carried by a few hundred neurons.
New pairs that hold the scenario fixed show that it follows the subject matter of the moral pairs more than the moral standard (\Cref{sec:followup}).
Two further lists gave no clear result.
Neurons selected for responding more to a judged edit than to a descriptive one lowered both normative gaps while sparing the descriptive pairs in five of the 28 combinations of model and frame, and the neurons that tell a moral matter from the others in the categorisation frame, removed while the model judges, did not lower the moral gap specifically once the change is read relative to the gap (3 of 14 models with the cue, 1 without).

The same recordings answer a different question: which neurons shift in a consistent direction whenever the acceptable version of a pair becomes the unacceptable one, and whether the same neurons do so for moral and for normative pairs.
They largely do.
Across the neurons of a layer, the mean signed shift on moral pairs correlates with the mean signed shift on normative pairs at 0.70 to 0.94 in the late layers of every model, and with the shift on descriptive pairs at 0.45 to 0.88.
Removing the 25 most consistent neurons per layer, selected on normative pairs on one half and tested on the other, lowers the gap for moral and for normative pairs alike in 10 of 14 models with the cue, in eight of them for descriptive pairs too, by 3 to 20\,\%; without the cue the fall spares the descriptive pairs in five models (Llama-3.2-1B, Llama-3.2-3B-Instruct, Qwen2.5-14B-Instruct, Qwen2.5-3B, Gemma-2B), while in the 8B and the Qwen2.5-7B models removing these neurons raises the gap instead.
A code for stance shared between moral and normative judgment, and in part with descriptive evaluation, is the counterpart, at the level of neurons, of the result for the located heads.
All of this is exploratory, and the pairs on which a list was chosen never served to evaluate it.

\begin{table}[t]
\centering
\caption{\textbf{Held-out ablation of neuron lists (exploratory).}
For each model and frame, the change of the behavioural gap when a list of neurons is replaced by its mean at the final token, minus the mean change under ten random lists with similar activity, as a percentage of each pair's own gap, for moral (M), normative (N) and descriptive (D) pairs; lists of 25 neurons per layer selected on half of the pairs and evaluated on the other half, both ways.
Left: neurons selected for the moral-versus-normative difference in share; $\dagger$ marks a fall specific to moral pairs (the moral interval below zero, and the interval of the moral minus the normative change below zero, in absolute and in relative terms).
Right: neurons selected for a consistent direction of shift on normative pairs; $*$ marks a fall on moral and normative pairs with the descriptive pairs spared, $**$ a fall on all three.}
\label{tab:neurons}
\scriptsize
\setlength{\tabcolsep}{3.5pt}
\begin{tabular}{@{}l rrr rrr rrr rrr@{}}
\toprule
 & \multicolumn{6}{c}{Moral-versus-normative lists} & \multicolumn{6}{c}{Stance lists} \\
\cmidrule(lr){2-7}\cmidrule(lr){8-13}
 & \multicolumn{3}{c}{with cue} & \multicolumn{3}{c}{without cue} & \multicolumn{3}{c}{with cue} & \multicolumn{3}{c}{without cue} \\
\cmidrule(lr){2-4}\cmidrule(lr){5-7}\cmidrule(lr){8-10}\cmidrule(lr){11-13}
Model & M & N & D & M & N & D & M & N & D & M & N & D \\
\midrule
Llama-3.2-1B & $+1$ & $-2$ & $-4$ & $-9$$^\dagger$ & $-1$ & $+1$ & $-15$$^{*}$ & $-11$ & $0$ & $-13$$^{*}$ & $-8$ & $-1$ \\
Llama-3.2-1B-Instruct & $-2$ & $-3$ & $-4$ & $-4$ & $-4$ & $+4$ & $-12$$^{**}$ & $-19$ & $-21$ & $-6$ & $-8$ & $+2$ \\
Llama-3.2-3B & $0$ & $-3$ & $0$ & $-8$$^\dagger$ & $-2$ & $+3$ & $-9$$^{**}$ & $-14$ & $-11$ & $-1$ & $-15$ & $-10$ \\
Llama-3.2-3B-Instruct & $+1$ & $+1$ & $-2$ & $+7$ & $0$ & $-3$ & $-11$$^{**}$ & $-20$ & $-29$ & $-16$$^{*}$ & $-8$ & $-2$ \\
Llama-3.1-8B & $-2$ & $-4$ & $-3$ & $-5$$^\dagger$ & $0$ & $+1$ & $-9$$^{**}$ & $-11$ & $-8$ & $+12$ & $0$ & $+1$ \\
Llama-3.1-8B-Instruct & $-3$$^\dagger$ & $0$ & $+2$ & $-9$$^\dagger$ & $+1$ & $+1$ & $-3$$^{**}$ & $-3$ & $-6$ & $+15$ & $+7$ & $0$ \\
Gemma-2B & $0$ & $-3$ & $-2$ & $+13$ & $+1$ & $+4$ & $-13$$^{**}$ & $-21$ & $-24$ & $-21$$^{*}$ & $-15$ & $-3$ \\
Gemma-2B-it & $0$ & $0$ & $+2$ & $-5$ & $+3$ & $-2$ & $+2$ & $-1$ & $-4$ & $-11$$^{**}$ & $-20$ & $-15$ \\
Qwen2.5-3B & $-4$ & $-6$ & $-6$ & $-7$ & $-8$ & $-3$ & $-8$$^{**}$ & $-7$ & $-2$ & $-10$$^{*}$ & $-14$ & $0$ \\
Qwen2.5-3B-Instruct & $-1$ & $-2$ & $-4$ & $-2$ & $+1$ & $+2$ & $-2$$^{**}$ & $-5$ & $-3$ & $+5$ & $-9$ & $-6$ \\
Qwen2.5-7B & $0$ & $-2$ & $-2$ & $+1$ & $+1$ & $-1$ & $+1$ & $-1$ & $+1$ & $+8$ & $+16$ & $+14$ \\
Qwen2.5-7B-Instruct & $-1$ & $-6$ & $0$ & $+12$ & $+3$ & $+1$ & $-3$ & $-17$ & $-9$ & $+31$ & $+9$ & $+7$ \\
Qwen2.5-14B & $-1$ & $+1$ & $+1$ & $+2$ & $-2$ & $-4$ & $0$ & $-2$ & $-17$ & $-2$ & $-5$ & $0$ \\
Qwen2.5-14B-Instruct & $-1$ & $-1$ & $-2$ & $+7$ & $+1$ & $-2$ & $-1$ & $-8$ & $-8$ & $-6$$^{*}$ & $-7$ & $-1$ \\
\bottomrule
\end{tabular}

\end{table}

\subsection{The final position through a Jacobian lens}
\label{sec:jlens}
The logit lens reads an intermediate \rev{hidden} state with the output embedding, as if no layer were left to run.
A Jacobian lens \citep{gurnee2026workspace} first carries the \rev{hidden} state of layer $\ell$ to the last layer through $J_\ell$, the average Jacobian of the last layer's \rev{hidden} state with respect to that of layer $\ell$, and then applies the final normalisation and the output embedding.
It thus \rev{estimates, at each layer, the model's next-word probabilities}, taking into account the layers that are still to come \rev{(which is loosely interpretable as showing what the model is about to say at that layer)}.
We fitted it with the reference implementation for each of the eight models up to 3B parameters, on 100 paragraphs of WikiText-103 drawn once with a fixed seed (a separate draw of 50 for the two Llama-3.2-3B models) and truncated to 128 tokens, and read the final position of every prompt of frames A1 and B at every layer.
The lens makes two questions answerable.
The first is whether the \rev{hidden state holds the standard at stake} when the prompt does not ask for it, which we read from the lens probability of the category words of \Cref{sec:category-frame} at the final position of a frame that asks for a good-or-bad judgment.
The second is whether that \rev{information} reaches the judgment.
For this we use directions: a set of words has a direction at layer $\ell$, the one in which the \rev{hidden} state must move to raise the words' score through the lens ($J_\ell^{\top}$ applied to their output embedding, scaled by the final normalisation), so the category or the answer can be removed (projected out), reversed or exchanged at the final position of layer $\ell$, and the effect read on the output as before.

\begin{figure}[t]
\centering
\includegraphics[width=\linewidth]{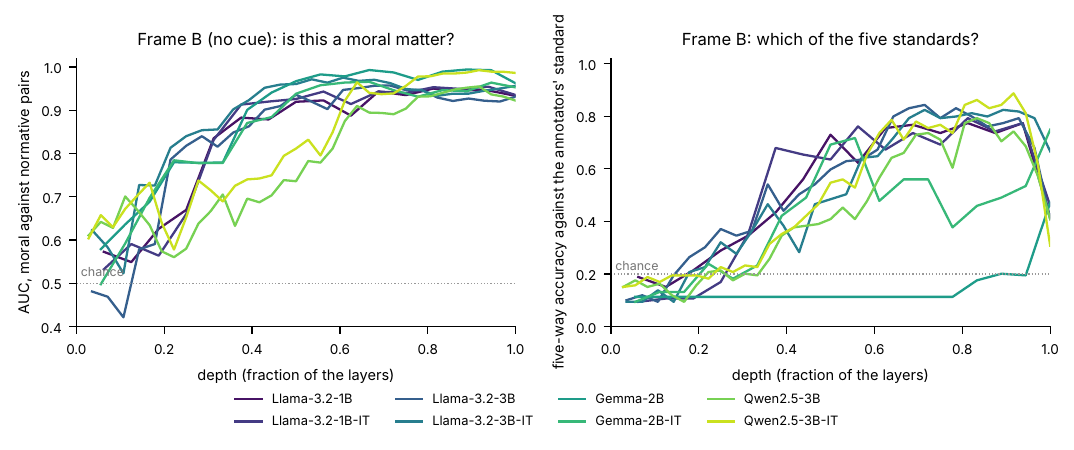}
\caption{\textbf{The hidden state carries the standard when nobody asks for it.}
Frame B (no cue), read through the Jacobian lens at the final position of every layer, in the eight models up to 3B parameters.
Left: how well the lens probability of the moral words separates moral from normative pairs (\ac{AUC}; 0.5 is chance).
Right: how often the most probable of the five standards' words is the annotators' standard, on the pairs whose standard they confirmed (chance 0.2).
The \ac{AUC} passes 0.75 between 0.2 and 0.5 of the depth in every model; the five-way reading rises with it in the Llama and Qwen models and peaks before the output, while in the Gemma models it appears only at the output.}
\label{fig:lens}
\end{figure}

Without the cue, the standard is there (\Cref{fig:lens}).
In frame B, which ends ``\stem{} is'' and names no standard, the lens probability of the moral words separates moral from normative pairs with an \ac{AUC} of 0.95 to 0.99 at the best layer, which lies at 0.6 to 0.9 of the depth (the \ac{AUC} passes 0.75 from 0.2 to 0.5 of the depth on).
In the Llama and Qwen models the most probable of the five standards' words is the annotators' standard for 77 to 89\,\% of the 159 pairs whose standard they confirmed, at the best layer (chance 20\,\%), more than at the output itself (31 to 67\,\%), where the next word is about to be an answer; in the Gemma models it is 47 and 75\,\%, at the output.
With the cue the word ``morally'' dominates this reading: the moral words win the choice among the five standards for nearly every pair, although they still rank moral pairs above normative ones (\ac{AUC} 0.84 to 0.96).
The judgment itself, the gap $S(\text{perm}) - S(\text{rest})$ read through the lens, reaches half of its final size at 0.6 to 0.9 of the depth in the Llama and Qwen models and late in the Gemma models (in Gemma-2B only at the output), with no consistent order between moral, normative and descriptive pairs (they differ by at most 0.18 of the depth).

The interventions act at the final position of three layers, at a quarter, a half and three quarters of the depth, against ten controls built in the same way from random pairs of vocabulary tokens, and the change of each pair's gap is read relative to that gap, as for the neurons and by the same rules.
The answer direction, the lens direction of the positive minus the negative answer words, behaves as the judgment itself from about three quarters of the depth: reversing it there reverses most or all of the gap for moral, normative and descriptive pairs alike in the Llama models and Gemma-2B-\ac{IT} (in Llama-3.2-1B with the cue by 112, 104 and 108\,\% of the gap), and a smaller part of it in the other three models (in the two Qwen models less for moral than for normative pairs); by that rule it is a general fall in seven of the eight models in each frame.

The category direction, the lens direction of the moral words minus that of the other standards' words, carries a small part of the judgment that is specific to moral pairs, and only without the cue.
Removed, reversed or exchanged for the direction of the safety words at one depth in frame B, it lowers the moral gap by 2 to 8\,\% (in Gemma-2B-\ac{IT} by 9 to 11\,\%, and by 19\,\% when reversed), while the descriptive gap barely moves (a fall of at most 2\,\%) and the normative gap falls by about half as much in the two Qwen models and Llama-3.2-3B and not at all in the others: a fall specific to moral pairs in seven of the eight models, all but Llama-3.2-1B.
With the cue the fall is specific to moral pairs in one model only.
This is the size of the feature and neuron residues (\Cref{sec:features-discovery,sec:neurons}), now found with a direction that was not selected on any pair: the hidden states carry, without being asked, which standard a pair is about, and that \rev{information} feeds the good-or-bad judgment, by a few per cent.

\paragraph{\vio{Do the located heads write the category?}}\label{sec:category-heads} \vio{To test whether the category of a pair reaches the located heads, we recorded, at the final position of both versions of every pair in frames A1 and B and in all 14 models, what the located heads write into the hidden state (the sum of their ten outputs), what each of the 20 random sets of ten heads of the removal test writes, and the hidden state after every layer.
A linear classifier (ridge regression on 32 principal components, after a fixed random projection to 256 dimensions) was trained on four fifths of the pairs and tested on the rest, with both versions of a pair always in the same part, five times with new splits; the score is the balanced accuracy.
\grn{The principal components and the classifier are always fitted on the training part only, and fitted again for every permutation of the labels; the standardisation and the projection, which use no labels, are fitted once on all prompts \grn{(they see the prompts of the test part, not their labels)}, which gives the same accuracy as fitting them on the training part (within 0.001 in the two models that we checked).
In every model and frame, the accuracy lies \grn{above every one of 600 permutations of the labels (for the five standards, lowest accuracy 0.72 against a highest permutation accuracy of 0.29), so that every test passes a Holm correction over the 14 models and the two frames ($p = 1/601$ in each test)}.}
The five standards can be read from what the located heads write in every model and frame: without the cue the median balanced accuracy over the models is 0.83 (guessing would be right in one case out of five, 0.20; random sets 0.83; hidden state after the deepest located head 0.91), and with the cue 0.86 (random sets 0.85; hidden state 0.89).
For moral against the other normative pairs it is 0.91 without the cue and 0.94 with it (random sets 0.90 and 0.93).
The located heads are above every random set in at most 3 of the 14 models and below every random set in none.
The category of a pair thus reaches the located heads as it reaches other heads, and the located heads still carry the same share of the effect for every category (\Cref{sec:heads-heldout}).
\grn{\grn{On the twins, which keep the scenario and the edit and change only who bears the consequence, this information follows the standard (below).}}}

\paragraph{\grn{The category on the twins.}}\label{sec:twins-standard} \grn{The main pairs differ in topic and wording as well as in standard.
The twins of \Cref{sec:followup} hold the scenario, the verb and the edit fixed and change only who bears the consequence, and with it the standard (moral against prudential in 30 twins, moral against epistemic in five).
Asked what each phrase is a matter of, with the five example sentences of \Cref{sec:category-frame}, the models give more probability to the moral words for the moral member than for its twin in a median of 84\,\% of the 35 twins (in 13 of the 14 models more often than chance allows, by a one-sided sign test at $p<0.05$; 80\,\% and 12 models on the 23 twins whose two members both passed the check of the language models).
The classifier of moral against the other normative pairs, trained on the main pairs and applied to the twins without retraining, gives the moral member the higher moral score, in the hidden state after the deepest located head, in 90\,\% of the twins with the cue and 91\,\% without it (all 14 models); on what the located heads write, in 89\,\% and 79\,\% of the twins (13 and 12 of the 14 models), and on what ten random heads write, in 89\,\% and 83\,\%.
The information about the category therefore follows the standard, and it reaches the located heads, as it reaches other heads, also when \grn{the scenario and the edit are} held fixed.}

\subsection{Adding directions at the final position}
\label{sec:steering}
Everything above removed or replaced components.
Here a direction is added.
From the \rev{hidden state} at the final position, after the layers at a quarter, a half and three quarters of the depth, we took two differences of means \rev{(as in \citealp{zou2023representation,rimsky2024steering})} on one half of the pairs (the halves of \Cref{sec:features-discovery}): the stance direction, the mean over moral and normative pairs of the acceptable minus the unacceptable version, and the category direction, the mean of the moral stems minus the mean of the normative stems.
Each was added at the final position of the pairs of the other half, and of every descriptive pair, at $\pm0.5$, $\pm1$ and $\pm2$ times its own length, and compared with five random directions of the same length.
A direction \rev{counts as} steerable if one step up and one step down move the score in opposite directions by at least a tenth of the pair's gap per step, and shared if it is steerable for moral, normative and descriptive pairs with the same sign.

The stance direction is shared in every model.
At three quarters of the depth one step raises the evaluation of both versions of a pair by 40 to 240\,\% of the pair's gap, typically 70 to 150\,\%, for moral, normative and descriptive pairs alike, with and without the cue, in all 14 models.
The stance directions computed on moral pairs alone and on normative pairs alone point the same way at that depth (cosine 0.72 to 0.94), and the one computed on descriptive pairs nearly so (0.62 to 0.88).
This is the counterpart, by addition, of what removal showed: the code for stance that the neurons share and the answer direction of the lens are not only needed for the judgment but enough to move it, for every kind of pair.

The category direction does what its name says in the frame that asks for the standard: added at three quarters of the depth in ``\stem{} is a matter of'', one step raises the probability of the moral words by 6 to 39 points (14 to 34 for moral pairs), for moral, normative and descriptive pairs alike, in all 14 models.
Added in the frames that ask for a judgment, it does not single out the moral pairs.
Without the cue it lowers the evaluation of every kind of pair in every model, by 5 to 51\,\% of the gap per step and not consistently more for moral pairs (shared by the three kinds by the rule above in 8 of 14 models; with the cue, where its effect is less consistent, in 6), and it points partly against the stance direction (cosine $-0.01$ to $-0.47$).
Read together with the \rev{interventions on} the lens's category direction, which lowered the moral gap specifically by a few per cent, the \rev{information about} the standard carries a little of the moral judgment of its own and, when pushed, mostly the negative tone that moral content has in these pairs.

\subsection{New pairs that fix the edited word or the scenario}
\label{sec:followup}
The neuron residue of \Cref{sec:neurons} leaves two explanations open that the new pairs cannot separate, because their moral and normative pairs differ both in the edited words and in what they are about: the neurons could respond to the edited words, or to the subject matter of moral pairs rather than to the moral standard.
We wrote three further sets of pairs.
In 80 function-word pairs (40 moral, 40 normative) the two versions differ only in a word with no valence of its own (\emph{with} and \emph{without}, \emph{before} and \emph{after}, a quantity, a number), and each kind of edit occurs in moral and in normative pairs.
In 35 twins, two pairs share the scenario, the verb and the edit and differ only in who bears the consequence: other people in the moral member, the agent alone in its prudential twin (``Driving off without fastening your child's seatbelt'' against ``... your own seatbelt''), or, in five twins, a person who is told something against a conclusion that is drawn (moral against epistemic).
A third set of 24 pairs describes someone's conduct with neither version better (``Painting the garden fence on Monday'' against ``... on Tuesday'').
It was written for a test of whether the neurons that separate normative from descriptive pairs respond to agency rather than to evaluation; that test needs a contrast we have not run, and these pairs are released but not analysed here.

\paragraph{Annotation by language models, checked against our annotators.}
These pairs were not rated by our annotators.
Three language models (Claude Sonnet~5.5, \ac{GPT}-5.5 and Gemini~3.1~Pro) answered the annotators' form for every pair, each in its own order, with its own placement of the two versions and without the intended category, and the retention rule of \Cref{sec:validation} was applied to their answers with two of three as the majority (for the conduct pairs: two of three see no better option, and two of three see someone's conduct).
They kept 73 of the function-word pairs (37 moral, 36 normative), the 23 twins whose two members both passed, and 21 of the conduct pairs.
Since labels checked by language models are not by themselves a validation, we first checked the three models against our annotators.
With the same settings they answered the form for the 242 pairs our annotators had rated, and we applied the alternative annotator test of \citet{calderon2025alttest} to the class of standard, the better option and each rater's own verdict on the pair (whether that rater's answers meet the retention rule).
Each model is compared with each annotator in turn, on how often it agrees with the other four at least as well as that annotator does; it passes against that annotator if it is not worse by more than ten points ($\varepsilon = 0.1$, the smallest margin the authors recommend; one-sided tests with the Benjamini-Yekutieli correction \rev{\citep{benjamini2001control}} at $q = 0.05$), and it passes a question if it does so against at least three of the five.
The new pairs would be used as annotated by the models \rev{only} if every model passed all three questions.
Every model passed all three against all five annotators; one of our annotators, tested in the same way against the other four, would not have passed on the class of standard.
On the pairs where at least four of our annotators agreed, the models gave the annotators' class of standard for 224 to 228 of 228 pairs and their better option for 238 to 242 of 242.
Two limits remain.
The models agree with one another more than our annotators do (Krippendorff's $\alpha$\rev{, \citealp{hayes2007answering},} for the class of standard 0.95 against 0.74), and as a group they are lenient: they would have kept 14 of the 15 pairs our annotators dropped, where three of our annotators under the same rule would have kept between 3 and 14 of them.
A few of the new pairs may therefore be pairs that people would drop.

\paragraph{Results.}
On the function-word pairs, the neuron lists of \Cref{sec:neurons}, chosen on the new pairs, lower the moral gap more than the normative one without the cue in 4 of the 14 models on the absolute change and in 2 relative to each pair's gap, both of the Llama family: Llama-3.2-1B (moral $-6$\,\%, normative $+1$\,\%) and Llama-3.1-8B-Instruct ($-3$, $+2$); with the cue, in 2 and in none.
Where the edit carries no valence of its own, the residue survives in part: it is not only a response to the edited words.

The twins show where it comes from.
With the scenario and the edit held fixed, the same lists lower the moral member and its twin alike, by 5.9 and 5.4\,\% in Llama-3.2-1B and by 4.5 and 3.8\,\% in Llama-3.2-3B without the cue; compared twin by twin, the moral member falls more than its twin only in the two 8B Llama models without the cue, by one to two points (3 of 56 combinations of model, frame and list).
The residue follows what the moral pairs are about more than the moral standard.
The rest is shared as well: the stance lists change the moral member and its twin alike (1 of 56 combinations differs), and in the eight models up to 3B parameters the located heads carry the same share of the effect for both (ratio 0.91 to 1.16; higher for the moral member only in Gemma-2B).

The models' behaviour, by contrast, does tell the two apart: the gap between the two versions is larger for the moral member than for its twin in 11 of the 28 combinations of model and frame, among them both frames of both Qwen2.5-14B models, and smaller in none.
The models often weigh a harm to others more than the same harm to oneself; of the components we can name, only the neuron lists of the two 8B Llama models treat the moral member differently, by a point or two.

\section{Human validation: instrument, rules and outcome}
\label{app:validation}

\paragraph{Instrument.}
Each annotator received one workbook: a sheet of instructions, a sheet of nine worked examples (three of them pairs in which more than one standard is rated high), and three parts of 75 pairs each, with the answers in yellow cells and a mark for rows that are unclear or odd.
The two options of a pair appear side by side; the order of the pairs and the left/right placement were drawn independently for each annotator.
The questions, in the order of the columns: which option is better (1, 2, or 0 for no clear difference); how much worse the worse option is, on a scale from 1 to 3 meant to be comparable across rows of different kinds; whether anyone's action, choice or work is being judged (Y/N; N leaves the remaining questions empty); and, for each of the five standards (moral, prudential, etiquette or social convention, aesthetic, epistemic), how much the difference between the two options is a matter of that standard, from 0 (not at all) to 3 (very much), rated separately.
The instructions state that these five ratings are about what kind of difference it is, not how big (that is the second question), that several can be high for the same row, that a zero may be left blank once at least one of the five is filled, and give five rules for hesitant cases (whose interest is at stake decides between moral and prudential; manners that also cost the doer something are etiquette first; a judged piece of work is aesthetic, a thing in a state is N; a person drawing a conclusion is epistemic, not N; when torn between two standards, rate both 2 rather than 3 and 0).
An earlier version of the form asked for one standard per row instead of the five ratings; it was replaced before any annotator had returned a workbook, and the annotators restarted with the graded form (\Cref{app:prereg}).
Annotators were told only that the task is a judgement about short phrases, that it takes about two hours, and that the instructions are in the file.
At the end they stated whether English is their first language, whether they had been told the aims of the study, whether they had read any earlier write-up of this project, whether they had seen model results, and whether they had rated these pairs before with an earlier form.
Language models were not used as raters or as helpers.

\paragraph{Acceptance rule.}
Each annotator's answers to a row are first reduced to the quantities the rule uses: the class of standard is the standard with the unique highest rating; N when the annotator answered that nobody is judged or rated every standard 0; a tie among non-moral standards counts as the non-moral class; a tie between the moral and a non-moral standard confirms no class.
The moral rating is the rating of the moral standard (0 when nobody is judged).
With $R$ annotators and ``two thirds'' meaning at least $\lceil 2R/3\rceil$, an item is accepted if (1) two thirds pick the intended better option; (2) two thirds give the intended class (moral for moral items, any of the four non-moral standards for norm items, N for descriptive items); (3) the median moral rating is at least 2 for moral items and at most 1 for norm and descriptive items; and (4) fewer than half of the annotators flag the item as unclear, every annotator rated it, and none left a deciding question blank.
Analyses of one kind of norm use only norm items for which two thirds rated the intended kind highest and at least 2.
The strict subset contains the accepted items on which the annotators were unanimous about the better option and the class of standard; the clean-contrast subset contains the accepted items whose median ratings are at least 2 for the intended standard and at most 1 for every other standard (descriptive items: N from two thirds).

\paragraph{Which workbooks count.}
Every workbook returned before the retained set was fixed is scored, except one in which more than 5\,\% of the rows lack an answer to a deciding question; such a workbook is left out and named.
Annotators could be added until then; the rule adapts through $R$.
Ratings that arrive later never change the retained set: they are scored as a separate panel and analysed as a sensitivity analysis next to the primary one.
No choice between panels is made after model results have been seen.
All ten workbooks were complete and were scored as returned.

\paragraph{Outcome.}
\Cref{tab:acceptance} gives the rated and retained pairs per category, with those whose subtype was confirmed, and the criteria failed by the rejected pairs: every rejection was on the class of standard, and one normative pair also failed the moral rating.
Agreement on the derived class is $\alpha=\AlphaQthree{}$ over the five standards and \AlphaQthreeClass{} at the level of the three classes the rule uses, over all \NratedTotal{} rated pairs.
The second round added \NratedRoundTwo{} spare prudential pairs after the first round had left the prudential subtype with 8 confirmed pairs where the plan required 10; 14 of them were retained and 7 confirmed, and every minimum was then met.
The five annotators are non-native speakers of English with 6 to 10 years of regular use; none had been told what the study expects, seen any earlier draft of this work or seen any model result.
No sensitivity scoring without an annotator was required.

\begin{table}[t]
\centering
\caption{\textbf{Outcome of the human validation.}
Rated and retained pairs per category, with those whose subtype was confirmed, the \rev{required} minimum, and the criteria failed by the rejected pairs (a pair can fail several).
\NratedRoundTwo{} of the rated pairs are the spare prudential pairs of the second round.}
\label{tab:acceptance}
\small
\setlength{\tabcolsep}{4pt}
\begin{tabular}{lrrrrl}
\toprule
Category & Rated & Retained & Subtype confirmed & Minimum & Rejections by criterion \\
\midrule
Moral & 89 & 80 & - & 60 & class 8 \\
Prudential & 41 & 37 & 15 & 10 confirmed & class 4 \\
Etiquette & 24 & 22 & 18 & 10 confirmed & class 2; moral rating 1 \\
Aesthetic & 24 & 24 & 22 & 10 confirmed & - \\
Epistemic & 24 & 24 & 24 & 10 confirmed & - \\
All normative & 113 & 107 & - & 60 & \\
Descriptive & 40 & 40 & - & 24 & - \\
\midrule
Total & 242 & 227 & & & \\
\bottomrule
\end{tabular}

\end{table}

\section{\rev{The analysis plan}}
\label{app:prereg}

\paragraph{What was fixed before any model was run on the new pairs.}
The hypotheses H1 to H6 and their endpoints; the three frames and the 28 answer words; the score $S$ and the inclusion threshold of 0.02; the head sets taken from the search on the original pairs (the located heads, \rev{the first five of them, the top head alone, the ten heads with the largest share and the 1\,\% of heads with the largest share}) and the feature sets taken from the same search (the moral-versus-neutral and moral-versus-norm sets of each \rev{dictionary}); the interpretation bands of \Cref{tab:bands} and the reading to be given under each; the full list of sensitivity analyses of \Cref{sec:heads-heldout}; the head selection on norm pairs that mirrors the one on moral pairs, and the calibrated feature search; the retention rule for the rated pairs, the minimum number of retained pairs per category and the use of the reserve pairs if a minimum was missed; the rating instrument itself; and the analysis code.
The plan, the candidate pairs and the code were \grn{complete} before the first model run on a new pair, and the list of retained pairs, the individual ratings and the rating tables were \grn{complete} before any model output was examined.

\paragraph{What was added afterwards, before any result was seen.}
(a) The exploratory analysis of the late-layer dip with the logit lens in frames A1 and B (\Cref{app:dip}).
(b) The conditional synonym frame ``is ethically'', run only if H1 and H2 fall in different reading bands (\grn{run for the features after the main results, Section~\ref{sec:features-cues}, and later for the heads, although H1 and H2 fall in the same band, Section~\ref{sec:further}}).
(c) Agreement on the standard reported also at the level of the three classes the retention rule uses and among the four non-moral kinds, as descriptive statistics.
(d) The reading of frame A0 next to A1 for the located heads: if the two fall in different bands, we say that the priming changed the circuit before interpreting the comparison of A1 with B.
(e) The analyses of the circuit in \Cref{sec:circuits}: mean-ablation of the head sets against random sets of the same size, attention targets and direct logit attribution of every head, layer-by-layer activation patching, and edge attribution patching into the located heads, together with the corpus contexts of the \ac{SAE} features.
Their endpoints were written down before any of them was run.
(f) Exact patching of the strongest estimated edges in four models whose estimate had been rough in the validation on the original pairs (\Cref{app:software}).
(g) The rating form: the form \grn{of the plan} asked each annotator for one standard per pair; it was replaced, before any annotator had returned a workbook, by the graded form described in \Cref{sec:validation}, which rates every standard separately, and the retention rule is applied to the class of standard derived from those ratings.
The annotators restarted from the beginning with the graded form.
(h) The second rating round on the 17 spare prudential pairs, which the plan prescribed for a subtype below its minimum, and the runs of the two 14B models on the spare pairs; made after the ratings of the first round and before any model result on a new pair had been seen.

\rev{The further tests of the located heads in \Cref{sec:further} (other prompts, descriptive pairs about a person, scenarios of the ETHICS benchmark, a choice between two options) were designed after the main results were known, like the other analyses that the text marks in this way.
For each of them, what would be measured and how each possible outcome would be read was written down before the models were run; they were run on all 14 models.}

None of these additions changes a primary endpoint.

\section{Software validation}
\label{app:software}

All checks below use the original pairs or synthetic data only.

\paragraph{Self-tests.}
On the author's machine (Apple silicon, \ac{MPS}, float32) 22 of 22 self-tests passed.
The head driver reproduces the stored per-pair head effects of the original pairs at $r=0.978$ to $0.997$ (threshold 0.97), and on \ac{CPU} at 0.9964.
The feature driver reproduces the stored features of the original pairs at $r=1.00000$ for the 12 Llama \rev{dictionaries}; for the two Qwen \rev{dictionaries}, $r=0.9925$ and $0.9951$ with the leading token and $1.00000$ without it; the difference comes from the tokenisation described in \Cref{sec:models}.

\paragraph{Planted effects.}
On synthetic data, the head analysis recovers a planted moral/norm ratio of 2.0 as 2.12 and 1.99 and a planted null as 0.96 and 0.99; the feature test finds 1, 3 and 8 planted features and none under the null.
Simulated workbooks were taken through every path of the rating pipeline (first panel, unfinished workbook, reserve round, fixing the retained set, later panels).

\paragraph{Rehearsal of the frames.}
Before the new pairs were run, the head driver was run with frames A1 and B on 36 original pairs (14 moral, 14 norm, 8 descriptive) on every model that could be loaded (13 of 14; Qwen2.5-14B base exceeded the memory of the machine and was not rehearsed); no head result was computed, since the located heads were selected on these pairs.
No frame failed for any model: the mean probability on the answer words was 0.39 to 0.78 in frame A1 and 0.42 to 0.74 in frame B, and the share of pairs passing the inclusion rule was 0.71 to 1.00 (A1) and 0.86 to 1.00 (B) for moral items, 0.71 to 1.00 and 0.79 to 1.00 for norm items, and 0.50 to 1.00 and 0.62 to 1.00 for descriptive items (4 to 14 items per cell).
Descriptive items drew less probability onto the answer words in frame A1 (0.21 to 0.56 per model, against 0.53 to 0.92 for moral items).
Measured time per item and frame ranged from 0.6\,s (Gemma-2B) to 18\,s (Llama-3.1-8B) and about 40\,s (Qwen2.5-14B-Instruct).
The feature driver and analysis ran for all 14 \rev{dictionaries} on 18 original pairs; of their numbers only the capture check is used (\Cref{sec:features-heldout}).

\paragraph{Edge attribution patching.}
Before the circuit analyses were run on the new pairs, the first-order edge estimate of \Cref{sec:circuits} was compared with exact edge patching on 12 original pairs per model (the strongest ten and ten random upstream edges for every query, key and value input of each located head and for the direct path to the logits).
Over all edges and pairs, the correlation between estimate and exact effect was 0.997 and 0.995 in Llama-3.1-8B and its instruct version, 0.985 and 0.987 in Llama-3.2-3B and its instruct version, 0.966 and 0.935 in Llama-3.2-1B and its instruct version, 0.993 in Qwen2.5-14B-\ac{IT}, 0.953 in Qwen2.5-7B, 0.838 in Qwen2.5-3B-\ac{IT}, 0.781 in Qwen2.5-7B-\ac{IT}, 0.766 in Gemma-2B, 0.705 in Qwen2.5-14B, 0.568 in Qwen2.5-3B and 0.554 in Gemma-2B-\ac{IT}, with sign agreement 0.90 to 0.98 everywhere except 0.81 in Qwen2.5-7B-\ac{IT}.
On the new pairs, the strongest estimated edges were then patched exactly in four of the six models below 0.85: Gemma-2B-\ac{IT}, Qwen2.5-3B, Qwen2.5-14B and Qwen2.5-7B-\ac{IT} (the last also has the lowest sign agreement).
Gemma-2B and Qwen2.5-3B-\ac{IT} were not included.

\section{Reproducibility and use of \ac{AI} assistance}
\label{app:records}

\paragraph{Release.}
\grn{The release \OSFref{} holds: the candidate pairs with their intended categories, the follow-up pairs of \Cref{sec:followup}, the original pairs and the ETHICS scenarios we used; the individual ratings (without names) and every answer of the three language models to the annotators' form; the located heads and features of every model; all code, with one script per part of the paper that runs every command; and the result files behind every table and figure.
The 13 dictionaries we trained are released with the paper.}

\paragraph{Compute.}
One Apple computer with an M2 Max chip and 96\,\ac{GB} of unified memory (\ac{MPS} backend, float32) for every model.
On the rated pairs, the head runs took 6 minutes to 2.4 hours per model up to 8B and about 5 hours per 14B model; the feature runs, under an hour per \rev{dictionary}; each stage of the circuit analysis, between a few per cent and half of a head run per model; and fitting the Jacobian lens with the reference implementation of \citet{gurnee2026workspace}, 2 to 5.5 hours per model.
The analyses themselves run on a \ac{CPU} in about two hours.
\grn{Each dictionary was trained on one NVIDIA \ac{GPU} (H100, L40S, RTX 5090 or RTX PRO 6000) in about 6 to 12 \ac{GPU} hours.}

\paragraph{Use of \ac{AI} assistance.}
We used Claude 5.2 Fable (Anthropic) to implement and refactor the codebase and improve the quality of the presentation (wordings and images); A Claude Fable also drafted the candidate new pairs, which were then checked mechanically and judged by the human annotators, who used no language model.
It also drafted the follow-up pairs of \Cref{sec:followup}, which were checked mechanically and judged by three language models through their providers' interfaces (Claude Sonnet~5.5, \ac{GPT}-5.5 and Gemini 3.1 Pro), after those models had been checked against the human annotators; these calls cost about 8.50 \ac{US} dollars.
\rev{The descriptive pairs about a person of \Cref{sec:further} were drafted, checked mechanically and judged by the three language models in the same way.}
We also used Claude (Fable 5.2) and \ac{GPT}-6 (Astra) with maximum reasoning effort to attack our paper's claims and check for internal weaknesses in our contributions and we used these feedback to adjust some sentences and make our claims clearer.
The authors are responsible for the design of the study, the code, the results and the text.

\end{document}